%% file: main.tex
\documentclass[final,3p,times]{elsarticle}
\usepackage{amssymb}
\usepackage[ansinew]{inputenc}  % Input-Encodung: ansinew fuer Windows
\usepackage{multicol}
\usepackage{amsmath}
\usepackage{graphicx}
\usepackage{booktabs}
\usepackage{xcolor}
\usepackage{svg}
\usepackage{nth}
\usepackage{calc}
\usepackage{cancel}
\usepackage{mathtools}
\usepackage{placeins}
\usepackage{listings}
\usepackage{xspace}
\usepackage{footnote}
\usepackage[bottom]{footmisc}
\usepackage{subfig}
\usepackage{layouts}
\usepackage{physics}
\usepackage{romannum}
\usepackage{bm}
\usepackage{scalerel}
\usepackage{stackengine}
\usepackage{mleftright}
\usepackage{courier}
\usepackage{multirow}
\usepackage{xfrac}
\usepackage{tablefootnote}
\usepackage{bbding}
\usepackage{stmaryrd}
\usepackage{wasysym}
\usepackage{nicefrac}

\usepackage{algorithm} 
\usepackage{algpseudocode} 
\usepackage{transparent}

\algnewcommand{\algorithmicor}{\textbf{ or }}
\algnewcommand{\OR}{\algorithmicor}
\algnewcommand{\AND}{\textbf{ and }}

\makeatletter
\def\ps@pprintTitle{%
	\let\@oddhead\@empty
	\let\@evenhead\@empty
	\def\@oddfoot{\centerline{\thepage}}%
	\let\@evenfoot\@oddfoot}
\makeatother

\DeclareMathOperator*{\argmax}{arg\,max}
\DeclareMathOperator*{\argmin}{arg\,min}
\DeclareMathOperator{\err}{err}
\newcommand{\Model}{\mathcal{M}}

\definecolor{TUMDarkGreen}{RGB}{000,124,048}
\newcommand{\linedashdotteddarkgreen}{{\color{TUMDarkGreen}-- $\cdot$ --}}

\newcommand{\linedashedred}{{\color{TUMRed}-- --}}

\newcommand{\linedottedgreen}{{\color{TUMGruen}$\cdots$}}
\newcommand{\BlockBlau}{{\color{TUMBlau}\ding{122}}}
\newcommand{\BlockHellBlau}{{\color{TUMBlau!20}\ding{122}}}
\newcommand{\BlockHellGreen}{{\color{TUMGruen!20}\ding{122}}}
\newcommand{\CircBlau}{{\color{TUMBlau}$\varocircle$}}
 \newlength\fheight  % definiert die HÃ¶he   fÃ¼r die Ausgabe der Tikz -plots
 \newlength\fwidth   % definiert die Breite fÃ¼r die Ausgabe der Tikz -plots

\svgsetup{inkscapearea=page}

\newcommand{\diag}{\mathop{\mathrm{diag}}}
\newcommand{\conj}[1]{\overline{#1}}

\definecolor{TUMBlau}{RGB}{0,101,189}
\definecolor{TUMBlauDunkel}{RGB}{0,82,147}
\definecolor{TUMBlauHell}{RGB}{152,198,234}
\definecolor{TUMBlauMittel}{RGB}{100,160,200}
\definecolor{TUMElfenbein}{RGB}{218,215,203}
\definecolor{TUMGruen}{RGB}{162,173,0}
\definecolor{TUMOrange}{RGB}{227,114,34}
\definecolor{TUMDunkelGrau}{gray}{0.8}
\definecolor{TUMGrau}{gray}{0.5}
\definecolor{TUMHellGrau}{gray}{0.2}
\definecolor{TUMLila}{RGB}{105,008,090}
\definecolor{TUMRed}{RGB}{156,013,022}

\usepackage{tikz}		
\usetikzlibrary{external}

\definecolor{TUMBlau}{RGB}{0,101,189}
\definecolor{TUMBlauDunkel}{RGB}{0,82,147}
\definecolor{TUMBlauHell}{RGB}{152,198,234}
\definecolor{TUMBlauMittel}{RGB}{100,160,200}
\definecolor{TUMElfenbein}{RGB}{218,215,203}
\definecolor{TUMGruen}{RGB}{162,173,0}
\definecolor{TUMOrange}{RGB}{227,114,34}
\definecolor{TUMGrau}{gray}{0.6}
\definecolor{TUMLila}{RGB}{105,008,090}
\definecolor{TUMRed}{RGB}{156,013,022}

\newcommand{\linefullblue}{{\color{TUMBlau}---}}
\newcommand{\linefullorange}{{\color{TUMOrange}---}}
\newcommand{\linefullgreen}{{\color{TUMGruen}---}}

\newcommand{\linedashedgreen}{{\color{TUMGruen}- - -}}

\newcommand{\Grad}{\ifmmode \mGrad \else  $\hspace{-0.3em}^\circ$\,  \fi}
\newcommand{\mGrad}{\ensuremath{^\circ}} % geht noch besser

\newcommand{\lived}[2]{($\ast$#1\ifthenelse{\equal{#2}{}}{}{, $\dagger$#2})}

\newcommand{\compaug}[1]{\underline{#1}}
\newcommand{\pdvtwo}[2]{\frac{}{}}

\renewcommand{\Re}{\operatorname{Re}}
\renewcommand{\Im}{\operatorname{Im}}

\newcommand{\sign}{\operatorname{sgn}}

\newcommand{\RNum}[1]{\uppercase\expandafter{\romannumeral #1\relax}}

\newcommand{\E}{\operatorname{\mathbb{E}}}

\newcommand{\Var}{\operatorname{Var}}

\makeatletter
\newcommand{\specialcell}[1]{\ifmeasuring@#1\else\omit$\displaystyle#1$\ignorespaces\fi}
\makeatother

\usepackage{bbm}

\begin{document}

\begin{frontmatter}

\pagenumbering{arabic} 



\title{Sparse Bayesian Learning for Complex-Valued Rational Approximations}


%\author[label1]{Felix Schneider \and Iason Papaioannou \and Max Ehre \and Daniel Straub \and Christoph Winter}
\author[label2]{Felix Schneider}
\author[label1]{Iason Papaioannou}
\author[label2]{Gerhard M\"{u}ller}

\address[label1]{Engineering Risk Analysis Group, Technical University of Munich, Arcisstr. 21, 80333 Munich, Germany}
\address[label2]{Chair of Structural Mechanics, Technical University of Munich, Arcisstr. 21, 80333 Munich, Germany}

\begin{abstract}
Surrogate models are used to alleviate the computational burden in engineering tasks, which require the repeated evaluation of computationally demanding models of physical systems, such as the efficient propagation of uncertainties.
For models that show a strongly non-linear dependence on their input parameters, standard surrogate techniques, such as polynomial chaos expansion, are not sufficient to obtain an accurate representation of the original model response. 
It has been shown that for models with discontinuities or rational dependencies, e.g., frequency response functions of dynamic systems, the use of a rational (Padé) approximation can significantly improve the approximation accuracy. 
In order to avoid overfitting issues in previously proposed standard least squares approaches, we introduce a sparse Bayesian learning approach to estimate the coefficients of the rational approximation. 
Therein the linearity in the numerator polynomial coefficients is exploited and the denominator polynomial coefficients as well as the problem hyperparameters are determined through type-II-maximum likelihood estimation. 
We apply a quasi-Newton gradient-descent algorithm to find the optimal denominator coefficients and derive the required gradients through application of $\mathbb{CR}$-calculus. 
The method is applied to the frequency response functions of an algebraic frame structure model as well as that of an orthotropic plate finite element model.
\end{abstract}

\begin{keyword}
Sparse Bayesian Learning \sep Sparse Models \sep Rational Approximation \sep Structural Dynamics \sep Surrogate Model \sep Frequency Response Function

%% or \MSC[2008] code \sep code (2000 is the default)g
\end{keyword}

\end{frontmatter}

%\linenumbers

\section{Introduction}
\label{sec:intro}
In many engineering fields, mathematical models are used to describe the behavior of an engineering system.
Typically, the model is defined by a set of differential equations, whose parameters define the characteristics of the system. 
% In structural dynamics, these models often describe the equation of motion of a system in terms of its displacement field. 
Quantities of engineering interest, such as displacements or stresses, can be obtained through solving these governing differential equations.
Commonly, the parameters in these models are assumed to be known and deterministic.
However, often these parameters are not known with certainty. 
To account for this uncertainty, the model can be defined in a probabilistic setting, which leads to differential equations, whose coefficients are random variables.
In many applications the aim of the subsequent analysis is to identify the probabilistic description of the system response or a function thereof in order to, e.g., assess the safety or serviceability of structural design. 
When a model includes a spatial dependency, typically, the spatial domain is discretized by a numerical method, often the finite element method.
In this case, the problem reduces to a discrete finite element system with random inputs. 
For instance, the response quantity of interest could be the frequency response  of a linear dynamic model of a structure with random structural parameters. 
Thereby,  the input-output relationship between system response and input forces for each outcome of the random structural parameters can be described through frequency transfer functions.

A number of methods have been developed to solve the problem of quantifying the uncertainty in the model response. 
A straightforward approach is the Monte Carlo (MC) method \cite{rubinstein2016simulation}. 
This method has the advantage that it only requires evaluations of the deterministic finite element system for a set of realizations of the random inputs and therefore can be coupled with \textit{black-box} finite element solvers. 
Moreover, its efficiency does not depend on the number of random inputs. 
However, it suffers from slow convergence rates. 
Advanced sampling methods, such as quasi Monte Carlo methods are able to accelerate the convergence of crude Monte Carlo, but they still require a considerable number of model evaluations for convergence \cite{owen2003quasi}. 
The moments of the response can be approximated with perturbation approaches, which give accurate solutions at low uncertainty levels, e.g. \cite{Lutes.2004,elishakoff2003finite}. 
The full probabilistic structure of the response can be determined through application of the law of preservation of probability content, leading to the probability density evolution method \cite{li2004probability,li2016probability}.

Rather than solving the problem directly, one may construct surrogate models that approximate the original, often computationally intensive model through a simple mathematical form, which is then used for uncertainty propagation. 
Examples of surrogate models are polynomial chaos expansions (PCE) \cite{ghanem1991stochastic,xiu2002wiener}, Neumann series expansions \cite{yamazaki1988neumann} and machine learning techniques such as neural networks \cite{papadrakakis1996structural} and Gaussian process regression \cite{girard2002gaussian}.
Surrogate models based on PCE have been extensively applied in uncertainty quantification.
They are based on projecting the model output onto the space spanned by a basis of multivariate polynomials that are orthogonal with respect to the input probability measure.
For practical purposes, the theoretically infinite set of basis polynomials is truncated, based on a chosen scheme. 
Popular choices include the total degree truncation, hyperbolic truncation or interaction order truncation as discussed in \cite{luethen.2021}. 
The projection can be computed via stochastic Galerkin schemes \cite{ghanem1991stochastic, matthies2005galerkin}, which require modification of existing deterministic solvers and are thus intrusive, or collocation-type methods \cite{xiu2007efficient,babuvska2007stochastic,berveiller2006stochastic}, which are non-intrusive as they require only discrete evaluations of the model. They can thus be coupled with black-box deterministic solvers.
Collocation-type methods estimate the coefficients of the expansion by numerical quadrature \cite{xiu2007efficient}, interpolation \cite{babuvska2007stochastic} or regression methods\cite{berveiller2006stochastic}.

Despite their successful application, the above methods suffer from a factorial growth of the number of coefficients in the PCE expansion in terms of the input dimensions and polynomial degrees.
In particular, for interpolation and regression methods the number of required model evaluations is approximately proportional to the number of unknown coefficients.
Hence, the computational cost of these  methods then becomes prohibitive in problems with high dimensional inputs and strong nonlinearities.
Several approaches to reduce the number of model evaluations for obtaining an accurate representation have been proposed in the context of polynomial chaos expansions. 
A popular approach is to find a subset of significant basis functions, i.e., a sparse PCE representation, that are sufficient to describe the model output.
We refer the reader to \cite{luethen.2021} for a comprehensive overview over sparse PCEs.
Adaptive strategies aim at finding the relevant terms in the set of basis functions, and thus a sparse PCE representation, through iteratively adding and deleting basis terms in the expansion.
In \cite{Blatman.2010} a stepwise regression technique was proposed that retains only a small number of significant basis terms.
A similar stepwise scheme was proposed in \cite{choi2004polynomial}.
The stepwise regression procedure is further improved in \cite{Blatman.2011}, wherein least angle regression is utilized for finding the significant terms in the basis.
In \cite{PENG201492}, a weighted $l_1$-minimization approach is proposed in conjunction with non-adapted random sampling.

Another approach is to cast the regression problem in a Bayesian setting \cite{Tipping.2001,ji2008bayesian,tsilifis2019compressive,sargsyan2014dimensionality,tsilifis2020sparse}.
The sparsity is imposed through a special sparsity-inducing prior structure. 
Therein, one assigns a prior distribution to the PCE coefficients and hyperpriors to the parameters of the prior distribution.
For linear models with Gaussian priors, the posterior distribution of the coefficients can be obtained analytically conditional on the distribution hyperparameters. 
Usually it is not possible to obtain the full joint distribution of the PCE coefficients and the hyperparameters and thus, one resorts to choosing the hyperparameters which maximize the model evidence, i.e., through solving a type-II-maximum likelihood estimation problem. 
The problem is thereby transformed to an optimization problem.
Various ways to perform the optimization of the marginal likelihood have been proposed, e.g., in \cite{Tipping.2001,tipping2003fast}. 
In \cite{Tipping.2001}, a sequential pruning approach is presented, wherein one starts from a full basis set and iteratively prunes basis terms based on their coefficient's precision.
In \cite{tipping2003fast}, a fast marginal likelihood maximisation that is based on subsequent addition and deletion of basis functions is proposed.
The posterior distribution of the hyperparameters is finally approximated by a Dirac at the optimal point.
In \cite{zhou2019} a hybrid sparse Bayesian Learning approach is presented, combining PCE with kernel and kriging methods.
In \cite{tsilifis2020sparse} the joint posterior distribution of the PCE coefficients and hyperparameters is approximated through application of variational inference.
Sparse Bayesian learning approaches have also been proposed for nonlinear models, e.g.,  \cite{zhou2019sparse,pan2015sparse}. 
In \cite{zhou2019sparse} a Bayesian learning approach in the context of finding a sparse set of parameters for a deep neural network is presented. 
A sparse Bayesian approach to the identification of nonlinear state space systems is presented in \cite{pan2015sparse}.

Instead of reducing the number of basis terms in the PCE representation, another approach aims at finding a suitable lower dimensional input space on which one can construct the PCE representation \cite{tipireddy2014basis}.
Through identifying a set of important input directions in the original input space, standard regression approaches can be applied, since the number of basis terms can be significantly reduced prior to computing the PCE coefficients. 
Several methods have been proposed to find a suitable lower-dimensional input space, e.g., in \cite{tsilifis2019compressive,papaioannou2019pls}.

For models that show discontinuities or a rational dependency in terms of the model parameters, the convergence of the PCE expansion becomes slow \cite{Chantrasmi.2009, Jacquelin.2015}. 
In this case, it can be beneficial to resort to more suitable surrogate models, e.g., rational or Pad\'{e}-type approximations, as proposed in \cite{Chantrasmi.2009,JacquelinE.2016,Schneider.2019,Lee.2021}.
A rational approximation is built from two polynomials, e.g., PCEs, that are divided by each other. 
Through the rational dependency, a very accurate representation of the original model response can be achieved for models which depend on the input parameters in a rational manner.
In \cite{Chantrasmi.2009}, the rational approximation was used to quantify the uncertainty in the response of complex fluid dynamic models with discontinuities.
In the specific context of approximating frequency response functions (FRF) for dynamic models, it was shown in \cite{Jacquelin.2015,Jacquelin.2015b} that the accuracy of standard PCE is poor and that spurious eigenfrequencies are introduced into the approximation.
In order to circumvent the slow convergence of standard PCE in the context of surrogate modeling for FRFs, the use of rational approximations is proposed in \cite{JacquelinE.2016,Schneider.2019,Lee.2021}. 
The polynomial coefficients are either found through stochastic Galerkin \cite{JacquelinE.2016,Lee.2021} or regression \cite{Schneider.2019} methods.
Other surrogate models for FRFs have also been proposed in the literature.
In \cite{yaghoubi2017sparse}, a stochastic frequency transformation was introduced, based on which a sparse PCE representation of the FRFs can be found.
In \cite{LU20191}, the authors present a multi-output Gaussian process model for uncertainty quantification of FRF models.

In the present contribution we propose a novel sparse Bayesian learning approach for rational approximations of complex-valued functions with real-valued random inputs.
The considered rational approximation is built from two polynomial chaos expansions with complex coefficients.
We make use of the fact that the model is linear in the numerator coefficients and find the posterior distribution of the numerator polynomial coefficients conditional on the denominator coefficients as well as the hyperparameters.
Subsequently, we find the maximum a-posteriori (MAP) esimate of the denominator coefficients conditional on the hyperparameters.
Since the denominator polynomial coefficients are complex-valued, we resort to the generalized $\mathbb{CR}$-calculus to derive the gradient of the objective function that appears in the MAP estimation.
Finally, an optimal set of hyperparameters is found through maximizing the model evidence, i.e., through a type-II-maximum likelihood estimation. 
We solve the problem in an iterative manner discarding (or pruning) all irrelevant terms until a convergent solution has been found.
We test the proposed method in linear structural dynamics problems, where the system response is described in terms of the frequency response function.

The outline of the paper is as follows. 
First, a description of the rational approximation surrogate model is given and a recently introduced regression-based method for estimating its coefficients is reviewed in Section~\ref{sec:ra_intro}.
In Section~\ref{sec:ra}, the novel sparse Bayesian approach to learning the coefficients in the rational approximation is introduced. 
Section~\ref{sec:res} presents a detailed numerical study on two models that investigates the performance of the proposed method. 
The first example investigates the approximation of the frequency domain response of a single degree of freedom frame structure, whereas the second example considers the response of the finite element model of a cross-laminated timber plate with orthotropic material behavior.
The paper closes with the conclusions in Section~\ref{sec:concl}.
\section{The Rational Approximation Surrogate Model}
\label{sec:ra_intro}
\subsection{Model description}
Consider a numerical model $\mathcal{M}$ that maps from the $d$-dimensional real space to the space of complex numbers, i.e., $\mathcal{M}: \, \mathbb{R}^d \to \mathbb{C}$.
The model $\mathcal{M}$ can for example return the uncertain dynamic frequency domain response of a mechanical structure.
$\mathbf{X}$ is a random vector with outcome space $\mathbb{R}^d$ and given joint probability density function, and models the uncertain input parameters of the numerical model. 
Then, $Y = \mathcal{M} (\mathbf{X})$ is a random variable with outcome space $\mathbb{C}$. 
Without loss of generality, we assume that the random vector $\mathbf{X}$ follows the independent standard Gaussian distribution. 
If $\mathbf{X}$ follows a non-Gaussian distribution, it is possible to express $Y$ as a function of an underlying independent standard Gaussian vector through an isoprobabilistic transformation \cite{rosenblatt1952remarks}. 
Let $P \left( \mathbf{X} \right)$ and $Q \left( \mathbf{X} \right)$ be truncated polynomial chaos representations, such that% with maximum orders $m_p$ and $m_q$
\begin{align}
	P \left( \mathbf{X}; \mathbf{p} \right) = \sum_{i=0}^{n_{\scriptscriptstyle p}-1} p_i \Psi_i \left( \mathbf{X} \right) , \\ Q \left( \mathbf{X}; \mathbf{q} \right) = \sum_{i=0}^{n_{\scriptscriptstyle q}-1} q_i \Psi_i \left( \mathbf{X} \right) .
	\label{eq:Hermite_pade}
\end{align}
Here $\{p_i \in \mathbb{C},i=0,\ldots,n_p-1\}$ and $\{q_i \in \mathbb{C},i=0,\ldots,n_q-1\}$ are complex coefficients and $\Psi_i$ are the multivariate orthonormal (probabilist) Hermite polynomials. 
The set $\{\Psi_i, i = 0,\ldots,n \}$ are constructed through the $d$-fold tensorization of the univariate normalized Hermite polynomials, i.e.,
\begin{equation}
	\Psi_{\bm{\alpha}} = \prod_{i=1}^{d} \psi_{\alpha_i} (X_i) \, ,
\end{equation}
In here, $\bm{\alpha} \in \mathbb{N}^d$ denotes the index set of the corresponding multivariate polynomial.
Two different truncation schemes are employed in this paper, the total degree and the hyperbolic truncation scheme \cite{Blatman.2011}.
In the total degree truncation, we retain all polynomials with a total polynomial degree less than or equal to $m$, i.e., 
\begin{equation}
	\sum_{i=1}^{d} \alpha_{i} \leq m \, ,
	\label{eq:total_degree_trunc}
\end{equation}
whereas in the hyperbolic truncation, we retain all polynomials whose index set obeys 
\begin{equation}
	\left( \sum_{i=1}^{d} \alpha_{i}^q \right)^{\frac{1}{q}} \leq m \, .
\end{equation}
For $q=1$ the hyperbolic truncation results in the total degree truncation scheme.
The resulting number of polynomial terms in the total degree truncation is $n=\binom{d+m}{m}$.
The truncated set of multivariate polynomials is finally sorted in the lexicographic order \cite{cox2007ideals}.
The truncation rules are separately applied to both, numerator and denominator polynomial, with maximum polynomial degrees $m_p$ and $m_q$ and truncation degrees $q_p$ and $q_q$.

We define the rational approximation (RA) $\mathcal{R} (\mathbf{X})$ obtained by taking the ratio of the two PCE representations of Eq.~\eqref{eq:Hermite_pade}:
\begin{equation}
	\mathcal{R} (\mathbf{X}; \mathbf{p}, \mathbf{q}) = \frac{P \left( \mathbf{X}; \mathbf{p} \right)}{Q \left( \mathbf{X}; \mathbf{q} \right)} = \frac{\sum_{i=0}^{n_{\scriptscriptstyle p}-1} p_i \Psi_i \left( \mathbf{X} \right)}{ \sum_{i=0}^{n_{\scriptscriptstyle q}-1} q_i \Psi_i \left( \mathbf{X} \right)} .
	\label{eq:ratio_rep}
\end{equation}

Stochastic collocation \cite{Chantrasmi.2009,Schneider.2019} and Galerkin \cite{JacquelinE.2016,Lee.2021} methods to determine the coefficients $\mathbf{p}$ and $\mathbf{q}$ in the expansions in Eq.~\eqref{eq:ratio_rep} have been presented in the literature.
In the following we shortly present the least-squares approach as presented in \cite{Schneider.2019} as it provides a natural choice for the initial point in the later presented sparse Bayesian algorithm.

\subsection{Least-Squares Approach for the Rational Approximation}
\label{sec:lsq}
In order to determine the unknown coefficients in Eq.~\eqref{eq:ratio_rep}, a regression method is developed in \cite{Schneider.2019}.
In this approach the coefficients are found by minimization of the modified mean-square error $\widetilde{\err}$, defined as
\begin{equation}
	\widetilde{\err} = \E \left[ \left| \Model \left( \mathbf{X} \right) Q \left( \mathbf{X} \right) - P \left( \mathbf{X} \right) \right|^2 \right] .
	\label{eq:res_l}
\end{equation}
$\widetilde{\err}$ is the mean-square of the truncation error $ \Model \left( \mathbf{X} \right) -  R \left( \mathbf{X} \right)$ multiplied by the denominator $Q \left( \mathbf{X} \right)$ of the rational approximation.
Using a set of samples $\{\mathbf{x}_k,k=1,\ldots,N\}$ of $\mathbf{X}$ and corresponding model evaluations $\{\mathcal{M}(\mathbf{x}_k),k=1,\ldots,N\}$, 
we estimate the coefficients $\{p_i\}$ and $\{q_i\}$ through minimizing a sample estimate of $\widetilde{\err}$. 
% Throughout this paper, we use Latin hypercube sampling (LHS) in order to generate samples of $\mathbf{X}$.
Substituting the expressions of Eq.~\eqref{eq:Hermite_pade} in Eq.~\eqref{eq:res_l} and performing the sampling approximation, we define the following minimization problem
\begin{equation}
	\left\{ \mathbf{p}, \mathbf{q} \right\} = \argmin_{\left\{ \tilde{\mathbf{p}}, \tilde{\mathbf{q}} \right\} \in \mathbb{C}^{n_p + n_q}} \frac{1}{N} \sum_{k=1}^{N} \left| \Model \left( \mathbf{x}_k \right) \sum_{i=0}^{n_q - 1} \tilde{q}_i \Psi_i \left( \mathbf{x}_k \right) - \sum_{i=0}^{n_p - 1} \tilde{p}_i \Psi_i  \left( \mathbf{x}_k \right) \right|^2 .
	\label{eq:argmin_pade}
\end{equation}
The minimizer is the solution of the following homogeneous linear system of equations of dimensions $(n_p+n_q) \times (n_p+n_q)$
\begin{equation}
	\mathbf{A} \bm{\rho} = \mathbf{0}.
	\label{eq:pad_eq_sys_final2}
\end{equation}
Here $\bm{\rho} = [\mathbf{p};\mathbf{q}]\in \mathbb{C}^{(n_p+n_q)}$ is the vector of unknown coefficients and $\mathbf{A} \in \mathbb{C}^{(n_p+n_q) \times (n_p+n_q)}$ is defined as follows 
\begin{equation}
	\mathbf{A} =
	\begin{bmatrix}
		\mathbf{\Psi}_P^T \mathbf{\Psi}_P & - \mathbf{\Psi}_P^T \diag \left( \mathbf{y} \right) \mathbf{\Psi}_Q \\ - \mathbf{\Psi}_Q^T \diag \left( \conj{\mathbf{y}} \right) \mathbf{\Psi}_P & \mathbf{\Psi}_Q^T \diag \left( \mathbf{y} \circ \conj{\mathbf{y}} \right) \mathbf{\Psi}_Q
	\end{bmatrix},
	\label{eq:pad_eq_sys_A}
\end{equation}
where $\diag \left( \cdot \right)$ denotes the diagonal matrix whose diagonal entries are the elements of $\left( \cdot \right)$, $\circ$ denotes the Hadamard product and $\conj{\cdot}$ denotes complex conjugation.
Matrices $\mathbf{\Psi}_P \in \mathbb{R}^{N\times n_p}$ and $\mathbf{\Psi}_Q \in \mathbb{R}^{N\times n_q}$ have as $(i,j)$-element $\Psi_j(\mathbf{x}_i)$ and vector $\mathbf{y} \in \mathbb{C}^N$ has as $i$-element the model evaluation $\Model \left( \mathbf{x}_i \right)$.
A non-trivial solution $\bm{\rho}\neq\mathbf{0}$ to the homogeneous system of Eq.~\eqref{eq:pad_eq_sys_final2} can be found through the minimum-norm least-squares solution.
\begin{equation}
	\bm{\rho} = \underset{\widehat{\bm{\rho}} \in \mathbb{C}^{(n_p+n_q)}}{\argmin} \norm{\mathbf{A} \widehat{\bm{\rho}}}_2 \enspace \text{subject to} \enspace \norm{\widehat{\bm{\rho}}}_2 = 1 \, .
\end{equation}
A solution to this problem can be found through applying singular value decomposition \cite{klema1980singular}.
\section{Sparse Bayesian Rational Approximation}
\label{sec:ra}
The number of unknown coefficients in the rational model of Eq.~\eqref{eq:ratio_rep} increases fast with increasing input dimensionality and polynomial orders.
For the case of the total degree truncation scheme of Eq.~\eqref{eq:total_degree_trunc}, the number of terms in the polynomial expansions of the numerator and denominator polynomials ($n_p$ and $n_q$, respectively) increase factorially with both the dimension and total polynomial order.
If the size of the experimental design is small, the least-squares approach presented in Section \ref{sec:lsq} is prone to overfitting. 
This implies significant computational demands, especially in problems where the underlying numerical model $\mathcal{M}$ is computationally intensive, as is often the case with finite element models.
To circumvent this problem, we propose a Bayesian probabilistic approach for determining the coefficients in the rational approximation model in Eq.~\eqref{eq:ratio_rep}, which we term sparse Bayesian rational approximation (SBRA).
The goal is to enable identifying those numerator and denominator coefficients that have the highest contribution to the predictability of the rational approximation. 
The proposed approach is based on the formalism of Tipping \cite{Tipping.2001}, which is generalized to enable the treatment of rational polynomial models with complex-valued coefficients.  

Our aim is to learn the coefficients of the rational approximation using a set of $N$ observation pairs on input samples $\{\mathbf{x}_k,k=1,\ldots,N\}$ of $\mathbf{X}$ and corresponding model evaluations $\{y_k = \mathcal{M}(\mathbf{x}_k),k=1,\ldots,N\}$. 
In the following we exploit a Bayesian perspective and pose the problem in a probabilistic setting.
We treat the coefficients in the RA as random variables and apply Bayes' theorem:
\begin{equation}
	f (\mathbf{p} , \mathbf{q} | \mathbf{y} ) = c_E^{-1} L (\mathbf{p} , \mathbf{q} | \mathbf{y}) f (\mathbf{p} , \mathbf{q}) \, ,
\end{equation}
where $f (\mathbf{p} , \mathbf{q} | \mathbf{y} )$ denotes the posterior distribution of the coefficients, $L (\mathbf{p} , \mathbf{q} | \mathbf{y})$ denotes the likelihood function and $f (\mathbf{p} , \mathbf{q})$ the prior distribution of the coefficients. The value $c_E$ is the normalization factor and is known as the model evidence.
In order to derive the likelihood function $L (\mathbf{p} , \mathbf{q} | \mathbf{y}) \propto f (\mathbf{y} | \mathbf{p}, \mathbf{q})$, we adopt the following additive error model
\begin{equation}
	y_k = R (\mathbf{x}_k; \mathbf{p} , \mathbf{q}) + \varepsilon_k \, ,
	\label{eq:model_error}
\end{equation}
where $\varepsilon_k$ denotes the additive error for the $k$-th observation. 
We model the errors as random variables following a multivariate proper zero-mean complex Gaussian distribution, i.e., $f (\bm{\varepsilon}) = \mathcal{CN} (\mathbf{0}, \beta^{-1} \mathbf{I}_N, \mathbf{0})$.
The complex normal distribution is defined in appendix \ref{app:compl_normal}. 
Through this assumption, the complementary covariance matrix $\widetilde{\bm{\Sigma}}_{\varepsilon\varepsilon}$, as defined in Eq.~\eqref{eq:comp_cov_matrix_def}, is zero and thus the real and imaginary parts of $\bm{\varepsilon}$ are uncorrelated and share the same covariance matrix.
$\beta$ defines the error precision, i.e., $\beta = \Var \left[ \epsilon_k \right]^{-1}$, which is common to all errors $\epsilon_k, k = 1,\ldots, N$.
An illustration of the error model can be found in Fig.~\ref{fig:error_model}.

%
% \begin{equation}
	%     f_{\bm{E}} (\bm{\varepsilon}) = \mathcal{CN} (\mathbf{0}, \beta^{-1} \mathbf{I}_N, \mathbf{0})
	% \end{equation}
%
Under the above assumptions, the likelihood reads
\begin{equation}
	L (\mathbf{p} , \mathbf{q} | \mathbf{y}) = \beta^N \exp{- \beta \left( \mathbf{y} - \mathbf{r} (\mathbf{x}; \mathbf{p}, \mathbf{q}) \right)^H \left( \mathbf{y} - \mathbf{r} (\mathbf{x}; \mathbf{p}, \mathbf{q}) \right)} \, ,
\end{equation}
where $\mathbf{r} (\mathbf{x}; \mathbf{p}, \mathbf{q}) = \diag(\bm{\Psi}_Q \mathbf{q})^{-1} \bm{\Psi}_P \mathbf{p} \in \mathbb{C}^N$ has as $i$-element the surrogate model evaluation $\mathcal{R} (\mathbf{x}_i; \mathbf{p}, \mathbf{q})$ and $\bm{\Psi}_P$ and $\bm{\Psi}_Q$ are defined in \ref{sec:lsq}.
\begin{center}
%	\includesvg{figures/error_model}
	\begingroup%
	\makeatletter%
	\providecommand\color[2][]{%
		\errmessage{(Inkscape) Color is used for the text in Inkscape, but the package 'color.sty' is not loaded}%
		\renewcommand\color[2][]{}%
	}%
	\providecommand\transparent[1]{%
		\errmessage{(Inkscape) Transparency is used (non-zero) for the text in Inkscape, but the package 'transparent.sty' is not loaded}%
		\renewcommand\transparent[1]{}%
	}%
	\providecommand\rotatebox[2]{#2}%
	\newcommand*\fsize{\dimexpr\f@size pt\relax}%
	\newcommand*\lineheight[1]{\fontsize{\fsize}{#1\fsize}\selectfont}%
	\ifx\svgwidth\undefined%
	\setlength{\unitlength}{202.5bp}%
	\ifx\svgscale\undefined%
	\relax%
	\else%
	\setlength{\unitlength}{\unitlength * \real{\svgscale}}%
	\fi%
	\else%
	\setlength{\unitlength}{\svgwidth}%
	\fi%
	\global\let\svgwidth\undefined%
	\global\let\svgscale\undefined%
	\makeatother%
	\begin{picture}(1,0.96296296)%
		\lineheight{1}%
		\setlength\tabcolsep{0pt}%
		\put(0,0){\includegraphics[width=\unitlength,page=1]{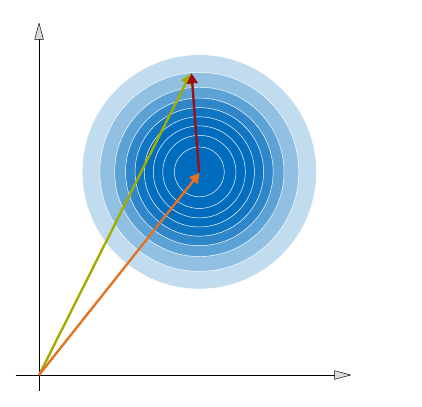}}%
		\put(0.20541004,0.14716733){\color[rgb]{0.89019608,0.44705882,0.13333333}\makebox(0,0)[lt]{\lineheight{1.25000012}\smash{\begin{tabular}[t]{l}$\mathcal{R} (\mathbf{x}_k;\mathbf{p},\mathbf{q})$\end{tabular}}}}%
		\put(0.60416063,0.82732615){\color[rgb]{0.61176471,0.05098039,0.08627451}\makebox(0,0)[lt]{\lineheight{1.25000012}\smash{\begin{tabular}[t]{l}$\varepsilon_k$\end{tabular}}}}%
		\put(0,0){\includegraphics[width=\unitlength,page=2]{error_model_svg-tex.pdf}}%
		\put(0.807022,0.10546478){\color[rgb]{0,0,0}\makebox(0,0)[lt]{\lineheight{1.25000012}\smash{\begin{tabular}[t]{l}$\Re$\end{tabular}}}}%
		\put(0.11246431,0.90464226){\color[rgb]{0,0,0}\makebox(0,0)[lt]{\lineheight{1.25000012}\smash{\begin{tabular}[t]{l}$\Im$\end{tabular}}}}%
		\put(0,0){\includegraphics[width=\unitlength,page=3]{error_model_svg-tex.pdf}}%
		\put(0.40931329,0.215818){\color[rgb]{0,0.42352941,0.74117647}\makebox(0,0)[lt]{\lineheight{1.25000012}\smash{\begin{tabular}[t]{l}$\mathcal{CN} (y_k;\mathcal{R} (\mathbf{x}_k;\mathbf{p},\mathbf{q}), \beta^{-1})$\end{tabular}}}}%
		\put(0.09957288,0.37005101){\color[rgb]{0.63529412,0.67843137,0}\makebox(0,0)[lt]{\lineheight{1.25000012}\smash{\begin{tabular}[t]{l}$\mathcal{M} (\mathbf{x}_k)$\end{tabular}}}}%
	\end{picture}%
	\endgroup%
	\captionof{figure}{
		Illustration of the model error in the complex plane that is defined to derive the likelihood function. 
		We assume an additive error $\varepsilon_k$ between the original model response $\mathcal{M} (\mathbf{x}_k)$ and the rational approximation $\mathcal{R} (\mathbf{x}_k)$. The error is assumed to be complex normally distributed.
		Under the stated assumptions, this will render the distribution of the data point $y_k$ given the parameters of the rational model, $\mathbf{p}$ and $\mathbf{q}$, to be rotationally symmetric in the complex plane around the surrogate model evaluation $\mathcal{R} (\mathbf{x}_k; \mathbf{p}, \mathbf{q})$.
	}
	\label{fig:error_model}
\end{center}

The prior distributions for both sets of coefficients are modelled as zero mean complex proper Gaussian distributions, i.e.,
\begin{gather}
	f (\mathbf{p} | \bm{\alpha}_p) = \mathcal{CN} ( p_i | \mathbf{0} , \bm{\Lambda}_{pp}^{-1}, \mathbf{0}) \label{eq:prior_p} \, , \\
	f (\mathbf{q} | \bm{\alpha}_q) = \mathcal{CN} ( q_i | \mathbf{0} , \bm{\Lambda}_{qq}^{-1}, \mathbf{0}) \, . \label{eq:prior_q}
\end{gather}
where $\bm{\Lambda}_{pp} = \diag{ \bm{\alpha}_p }$ and $\bm{\Lambda}_{qq} = \diag{ \bm{\alpha}_q }$ constitute the precision matrices and $\bm{\alpha}_p= \left[ \alpha_{p,1}; \ldots ; \alpha_{p,n_p} \right]$ and $\bm{\alpha}_q= \left[ \alpha_{q,1}; \ldots ; \alpha_{q,n_q} \right]$ are vectors containing the $n_p$ and $n_q$ hyperparameters (precisions) for each of the marginal prior distributions of the $n_p$ and $n_q$ coefficients in the expansion of the rational model.
We assume independence between the individual hyperparameters.
Following \cite{Tipping.2001,Berger.1985}, we specify hyperpriors over $\bm{\alpha}_p$ and $\bm{\alpha}_q$ as well as over the error precision $\beta$.
A suitable choice for these hyperpriors are Gamma distributions, i.e.,
\begin{gather}
	f ( \bm{\alpha}_p | a,b ) = \prod_{i=1}^{n_p} \mathcal{GA} ( \alpha_{p,i} | a, b ) \, , \\
	f ( \bm{\alpha}_q | a,b ) = \prod_{i=1}^{n_q} \mathcal{GA} ( \alpha_{q,i} | a, b ) \, , \\
	f ( \beta | c,d ) = \mathcal{GA} ( \beta | c,d ) \, . 
\end{gather}
The definition of the Gamma distribution is given in appendix \ref{app:gamma}. In the remainder of this work, we set $a=b=c=d=0$, which renders the hyperparameters to be improperly uniformly distributed over the log-space (cf. \cite{Tipping.2001}).
Following \cite{Tipping.2001}, this hierarchical prior is expected is to induce sparsity in the numerator and denominator coefficients.
The hierarchical Bayesian structure is depicted in Fig.~\ref{fig:hier_bayes}.
\begin{center}
%	\includesvg{figures/bayes_model_wbg_full}
	\begingroup%
	\makeatletter%
	\providecommand\color[2][]{%
		\errmessage{(Inkscape) Color is used for the text in Inkscape, but the package 'color.sty' is not loaded}%
		\renewcommand\color[2][]{}%
	}%
	\providecommand\transparent[1]{%
		\errmessage{(Inkscape) Transparency is used (non-zero) for the text in Inkscape, but the package 'transparent.sty' is not loaded}%
		\renewcommand\transparent[1]{}%
	}%
	\providecommand\rotatebox[2]{#2}%
	\newcommand*\fsize{\dimexpr\f@size pt\relax}%
	\newcommand*\lineheight[1]{\fontsize{\fsize}{#1\fsize}\selectfont}%
	\ifx\svgwidth\undefined%
	\setlength{\unitlength}{396.8503937bp}%
	\ifx\svgscale\undefined%
	\relax%
	\else%
	\setlength{\unitlength}{\unitlength * \real{\svgscale}}%
	\fi%
	\else%
	\setlength{\unitlength}{\svgwidth}%
	\fi%
	\global\let\svgwidth\undefined%
	\global\let\svgscale\undefined%
	\makeatother%
	\begin{picture}(1,0.39285714)%
		\lineheight{1}%
		\setlength\tabcolsep{0pt}%
		\put(0,0){\includegraphics[width=\unitlength,page=1]{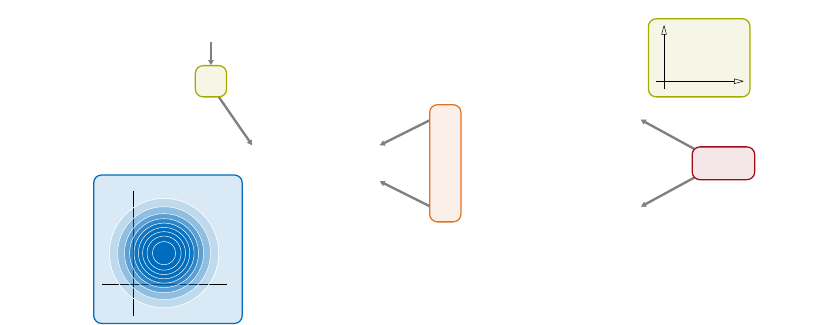}}%
		\put(0.53129329,0.24073528){\color[rgb]{0.89019608,0.44705882,0.13333333}\makebox(0,0)[lt]{\lineheight{1.25000012}\smash{\begin{tabular}[t]{l}$\mathbf{p}$\end{tabular}}}}%
		\put(0.53129329,0.13679182){\color[rgb]{0.89019608,0.44705882,0.13333333}\makebox(0,0)[lt]{\lineheight{1.25000012}\smash{\begin{tabular}[t]{l}$\mathbf{q}$\end{tabular}}}}%
		\put(0,0){\includegraphics[width=\unitlength,page=2]{bayes_model_wbg_full_svg-tex.pdf}}%
		\put(0.85626917,0.18687366){\color[rgb]{0.61176471,0.05098039,0.08627451}\makebox(0,0)[lt]{\lineheight{1.25000012}\smash{\begin{tabular}[t]{l}$a,b$\end{tabular}}}}%
		\put(0.24781114,0.28420254){\color[rgb]{0.63529412,0.67843137,0}\makebox(0,0)[lt]{\lineheight{1.25000012}\smash{\begin{tabular}[t]{l}$\beta$\end{tabular}}}}%
		\put(0,0){\includegraphics[width=\unitlength,page=3]{bayes_model_wbg_full_svg-tex.pdf}}%
		\put(0.23638778,0.35300848){\color[rgb]{0.61176471,0.05098039,0.08627451}\makebox(0,0)[lt]{\lineheight{1.25000012}\smash{\begin{tabular}[t]{l}$c,d$\end{tabular}}}}%
		\put(0,0){\includegraphics[width=\unitlength,page=4]{bayes_model_wbg_full_svg-tex.pdf}}%
		\put(0.48239329,0.36266088){\color[rgb]{0.89019608,0.44705882,0.13333333}\makebox(0,0)[lt]{\lineheight{1.25}\smash{\begin{tabular}[t]{l}$\scriptstyle\mathbb{C}$\end{tabular}}}}%
		\put(0,0){\includegraphics[width=\unitlength,page=5]{bayes_model_wbg_full_svg-tex.pdf}}%
		\put(0.6468129,0.0130329){\color[rgb]{0.89019608,0.44705882,0.13333333}\makebox(0,0)[lt]{\lineheight{1.25}\smash{\begin{tabular}[t]{l}$\scriptstyle\mathbb{C}$\end{tabular}}}}%
		\put(0.25560756,0.02248234){\color[rgb]{0,0.42352941,0.74117647}\makebox(0,0)[lt]{\lineheight{1.25}\smash{\begin{tabular}[t]{l}$\scriptstyle\mathbb{C}$\end{tabular}}}}%
		\put(0,0){\includegraphics[width=\unitlength,page=6]{bayes_model_wbg_full_svg-tex.pdf}}%
		\put(0.33633942,0.18876354){\color[rgb]{0,0.42352941,0.74117647}\makebox(0,0)[lt]{\lineheight{1.25000012}\smash{\begin{tabular}[t]{l}$\mathbf{y} = \mathcal{M} (\mathbf{X})$\end{tabular}}}}%
		\put(0.73783366,0.24073528){\color[rgb]{0.63529412,0.67843137,0}\makebox(0,0)[lt]{\lineheight{1.25000012}\smash{\begin{tabular}[t]{l}$\bm{\alpha}_{p}$\end{tabular}}}}%
		\put(0.73783366,0.13679182){\color[rgb]{0.63529412,0.67843137,0}\makebox(0,0)[lt]{\lineheight{1.25000012}\smash{\begin{tabular}[t]{l}$\bm{\alpha}_{q}$\end{tabular}}}}%
		\put(0,0){\includegraphics[width=\unitlength,page=7]{bayes_model_wbg_full_svg-tex.pdf}}%
	\end{picture}%
	\endgroup%
	
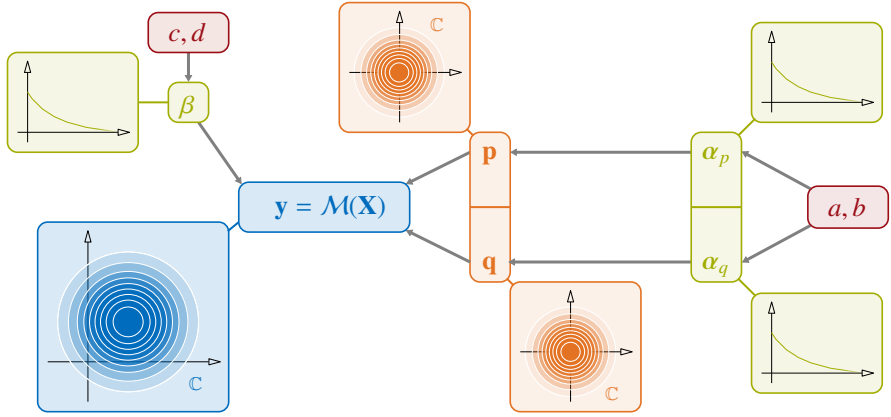
\captionof{figure}{
		Illustration of the hierarchical Bayesian model. 
		Based on the error formulation in Eq.~\eqref{eq:model_error} and the assumption of complex-normally distributed errors, the data conditional on the coefficients $\mathbf{p}$ and $\mathbf{q}$, i.e., the likelihood, will follow a complex normal distribution.
		The coefficients $\mathbf{p}$ and $\mathbf{q}$ are also complex-valued and assigned complex nomal distributions, which are again conditional on a set of hyperparameters.
		The hyperparameters in the likelihood and the prior distributions, the precisions $\beta$, $\bm{\alpha}_p$ and $\bm{\alpha}_q$, are real-valued and modeled through Gamma distributions. 
		Finally, we set $a=b=c=d=0$, which renders the distributions of the hyperparameters to be uniformly distributed over the log-scale.
	}
	\label{fig:hier_bayes}
\end{center}

In contrast to linear models, for the rational model, no closed-form solution for the joint posterior distribution of the coefficients $\mathbf{p}$ and $\mathbf{q}$ is available.
We therefore resort to the following iterative approach, in which we make use of the linearity with respect to the numerator coefficients $\mathbf{p}$. 
First, the posterior distribution of the numerator coefficients $\mathbf{p}$ conditional on the denominator coefficients as well as the hyperparameters $\bm{\alpha}_p$ and $\beta$ is computed analytically; due to the self-conjugacy of the complex normal distribution, the posterior distribution of $\mathbf{p}$ is complex normal, i.e., it takes the following form \cite{pedersen2015sparse}:
\begin{equation}
	f(\mathbf{p}|\mathbf{y}, \mathbf{q}, \bm{\alpha}_p, \beta) = \frac{1}{\pi^{n_p} \det \bm{\Sigma}} \exp{ \left( \mathbf{p} - \bm{\mu} \right)^H \bm{\Sigma}^{-1} \left( \mathbf{p} - \bm{\mu} \right)} \, ,
\end{equation}
with the posterior covariance matrix
\begin{equation}
	\bm{\Sigma} = \left( \bm{\Lambda}_{pp} + \beta \bm{\Psi}^H \bm{\Psi} \right)^{-1} \, ,
	\label{eq:post_cov_p}
\end{equation}
and the posterior mean
\begin{equation}
	\bm{\mu} = \beta \bm{\Sigma} \bm{\Psi}^H \mathbf{y} \, .
	\label{eq:post_mean_p}
\end{equation}
In here, $\bm{\Psi} = \diag(\bm{\Psi}_Q \mathbf{q})^{-1} \bm{\Psi}_P$ and $\left( \cdot \right)^H$ denotes the Hermitian transpose.
We note that the posterior distribution of $\mathbf{p}$ is also a proper complex Normal distribution, i.e., $\widetilde{\bm{\Sigma}}_{pp} = \mathbf{0}$.
The marginal evidence, $f( \mathbf{y} | \mathbf{q}, \bm{\alpha}_p, \beta)$, is obtained in closed-form and reads \cite{pedersen2015sparse}:
\begin{equation}
	f(\mathbf{y} | \mathbf{q}, \bm{\alpha}_p, \beta) = \frac{1}{\pi^N \det(\beta^{-1} \mathbf{I} + \bm{\Psi} \bm{\Lambda}_{pp}^{-1} \bm{\Psi}^H)} \exp{- \mathbf{y}^H \left( \beta^{-1} \mathbf{I} + \bm{\Psi} \bm{\Lambda}_{pp}^{-1} \bm{\Psi}^H \right)^{-1} \mathbf{y}} \, .\label{eq:marg_likelihood_y_cond_q}
\end{equation}
Based on Eq.~\eqref{eq:marg_likelihood_y_cond_q}, we find the maximum a-posteriori (MAP) estimate for the denominator coefficients, $\mathbf{q}^\ast$, conditional on the hyperparameters, through solving the following optimization problem
\begin{equation}
	\mathbf{q}^\ast = \argmax_{\mathbf{q} \in \mathbb{C}^{n_q}} f( \mathbf{y} | \mathbf{q}, \bm{\alpha}_p, \beta) f(\mathbf{q} | \bm{\alpha}_q ) \, .
	\label{eq:q_MAP}
\end{equation}
Based on $\mathbf{q}^\ast$, we can compute the mean vector  $\bm{\mu}$ and covariance matrix $\bm{\Sigma}$ of the numerator coefficients, which fully define the Gaussian distribution. 
We employ a Dirac approximation of the posterior distribution of $\mathbf{q}$ at the MAP estimate, i.e., $f(\mathbf{q}|\mathbf{y},\bm{\alpha}_q,\beta) \approx \delta(\mathbf{q}-\mathbf{q}^\ast)$, which gives the following approximation for the evidence conditional on the hyperparameters, 
\begin{equation}
	f( \mathbf{y} | \bm{\alpha}_p, \bm{\alpha}_q, \beta) \approx f ( \mathbf{y} | \mathbf{q}^\ast, \bm{\alpha}_p, \beta) f(\mathbf{q}^\ast | \bm{\alpha}_q ) \, .
\end{equation}
Subsequently, we maximize the model evidence over the remaining hyperparameters, which is also known as type-II-maximum likelihood, 
\begin{equation}
	[ \bm{\alpha}_p^\ast, \bm{\alpha}_q^\ast, \beta^\ast ] = \argmax_{\underset{\in \mathbb{R}^{n_p \times n_q \times 1}}{[ \bm{\alpha}_p, \bm{\alpha}_q, \beta ]}} f ( \mathbf{y} | \mathbf{q}^\ast, \bm{\alpha}_p, \beta) f(\mathbf{q}^\ast | \bm{\alpha}_q ) \, ,
	\label{eq:hyper_MAP}
\end{equation}
in order to find an optimal set of hyperparameters.
In the following, we explicitly write out the resulting expressions in the above steps.
% The posterior distribution of the numerator coefficients, conditional on all other parameters, reads
% %
% \begin{equation}
	% 	f(\mathbf{p}|\mathbf{y}, \mathbf{q}, \bm{\alpha}_p, \beta) = \frac{1}{\pi^{n_p} \det \bm{\Sigma}} \exp{ \left( \mathbf{p} - \bm{\mu} \right)^H \bm{\Sigma}^{-1} \left( \mathbf{p} - \bm{\mu} \right)} \, ,
	% \end{equation}
% %
% with the posterior covariance matrix
% %
% \begin{equation}
	% 	\bm{\Sigma} = \left( \bm{\Lambda}_{pp} + \beta \bm{\Psi}^H \bm{\Psi} \right)^{-1} \, ,
	% 	\label{eq:post_cov_p}
	% \end{equation}
% %
% and the posterior mean
% %
% \begin{equation}
	% 	\bm{\mu} = \beta \bm{\Sigma} \bm{\Psi}^H \mathbf{y} \, .
	% 	\label{eq:post_mean_p}
	% \end{equation}
% %
% In here, $\bm{\Psi} = \diag(\bm{\Psi}_Q \mathbf{q})^{-1} \bm{\Psi}_P$ and $\left( \cdot \right)^H$ denotes the Hermitian transpose.
% We note that the posterior distribution of $\mathbf{p}$ is also a proper complex Normal distribution, i.e., $\widetilde{\bm{\Sigma}}_{pp} = \mathbf{0}$.
% The marginal evidence is given by
% %
% \begin{equation}
	% 	f(\mathbf{y} | \mathbf{q}, \bm{\alpha}_p, \beta) = \frac{1}{\pi^N \det(\beta^{-1} \mathbf{I} + \bm{\Psi} \bm{\Lambda}_{pp}^{-1} \bm{\Psi}^H)} \exp{- \mathbf{y}^H \left( \beta^{-1} \mathbf{I} + \bm{\Psi} \bm{\Lambda}_{pp}^{-1} \bm{\Psi}^H \right)^{-1} \mathbf{y}} \, .\label{eq:marg_likelihood_y_cond_q}
	% \end{equation}
%

In order to find the MAP estimate of $\mathbf{q}$, we maximize the log of the objective function in Eq.~\eqref{eq:q_MAP}, which inserting Eqs.~\eqref{eq:marg_likelihood_y_cond_q} and \eqref{eq:prior_q} into Eq.~\eqref{eq:q_MAP} and taking the logarithm results in
\begin{equation}
	\mathbf{q}^\ast = \argmax_{\mathbf{q} \in \mathbb{C}^{n_q}} \left[ 
	\ln \det \bm{\Sigma} 
	% 	- \ln \det \left( \bm{\Lambda}_{pp} + \beta \bm{\Psi}^H \bm{\Psi} \right) 
	- \beta \mathbf{y}^H \left(  \mathbf{y} - \bm{\Psi} \bm{\mu} \right) 
	- \mathbf{q}^H \bm{\Lambda}_{qq} \mathbf{q} \right] \, .
	\label{eq:q_MAP2}
\end{equation}
Eq.~\eqref{eq:q_MAP2} is a nonlinear optimization problem in complex variables.
In the objective function, three terms appear.
The first term is $\ln \det \bm{\Sigma}$, i.e., the log-determinant of the posterior covariance matrix of the numerator coefficients, conditional on the denominator coefficients, wherein the denominator coefficients enter through the matrix $\bm{\Psi}$.
The second term $- \beta \mathbf{y}^H \left(  \mathbf{y} - \bm{\Psi} \bm{\mu} \right)$ includes the model prediction error and penalizes the misfit between the RA and the data points. 
The third term $- \mathbf{q}^H \bm{\Lambda}_{qq} \mathbf{q}$ can be interpreted as the sum of the square-magnitude of the denominator coefficients and thus penalizes the coefficients' magnitude.
In order to find a minimum, we resort to a gradient based maximization technique. 
Since $f( \mathbf{y} | \mathbf{q}, \bm{\alpha}_p, \beta) f(\mathbf{q} | \bm{\alpha}_q )$ is the product of two probability density functions (PDFs), it is real-valued and thus a necessary condition for the objective function in Eq.~\eqref{eq:q_MAP2} to take a maximum is given by
\begin{equation}
	\pdv{\conj{\mathbf{q}}} \left[ 
	\ln \det \bm{\Sigma} 
	% 	- \ln \det \left( \bm{\Lambda}_{pp} + \beta \bm{\Psi}^H \bm{\Psi} \right) 
	- \beta \mathbf{y}^H \left(  \mathbf{y} - \bm{\Psi} \bm{\mu} \right) 
	- \mathbf{q}^H \bm{\Lambda}_{qq} \mathbf{q} \right] = \mathbf{0} \, , 
	\label{eq:deriv_def_conj_q}
\end{equation}
where $\pdv{\conj{\mathbf{q}}}$ denotes the generalized (or Wirtinger) derivative with respect to the complex conjugate of the denominator coefficients $\mathbf{q}$, denoted as the conjugate cogradient.
The definition of the generalized derivatives can be found in \cite{Schreier.2010,kreutz2009complex}.
In Eq.~\eqref{eq:deriv_def_conj_q}, it is sufficient to consider only the conjugate cogradient, since the objective function is real-valued. 
In this case it holds $\pdv{f( \mathbf{q} ) }{\mathbf{q}} = \conj{\pdv{f( \mathbf{q} ) }{\conj{\mathbf{q}}}}$.
In order to solve Eq.~\eqref{eq:q_MAP2}, we employ a Quasi-Newton method and 
%
% use a symmetric rank-one conjugate gradient Steinhaug algorithm, provided by \cite{Sorber.2013}.
% Details about the algorithm can be found in \cite{Sorber.2012}. 
% The algorithm uses a rank-one update to update an approximation of the Hessian matrix of the problem in each iteration step.
%
use a limited memory Broyden-Fletcher-Goldfarb-Shanno (L-BFGS) algorithm, provided by \cite{Sorber.2013}.
Details about the algorithm can be found in \cite{Sorber.2012}.
The algorithm uses a quasi-Newton step to update an approximation of the Hessian matrix of the problem in each iteration.
We use the available line-search algorithm in the implementation by \cite{Sorber.2013}.
In the scope of this work, we also solved the optimization problem with other gradient-based methods, such as a symmetric rank one conjugate gradient and nonlinear conjugate gradient algorithm, also available in \cite{Sorber.2013}.
It was found, however, that for the investigated problems, the L-BFGS algorithm appears to be the most robust algorithm.

The quasi-Newton method requires the derivatives of the objective function with respect to the conjugate denominator coefficients.
The derivative of the log-objective with respect to the $i$-th conjugate denominator coefficient $\conj{q}_i$ can be found analytically and reads
\begin{equation}
	\pdv{\conj{q}_{i}} \left[ 
	\ln \det \bm{\Sigma} 
	% 	- \ln \det \left( \bm{\Lambda}_{pp} + \beta \bm{\Psi}^H \bm{\Psi} \right) 
	- \beta \mathbf{y}^H \left(  \mathbf{y} - \bm{\Psi} \bm{\mu} \right) 
	- \mathbf{q}^H \bm{\Lambda}_{qq} \mathbf{q} \right] = - \tr{\bm{\Gamma}_i \bm{\Psi}} - \beta \mathbf{y}^H \bm{\Psi} \left( \bm{\Gamma}_i \left( \bm{\Psi} \bm{\mu} -  \mathbf{y} \right) \right) - \alpha_{q,i} q_i \, .
\end{equation}
The matrix $\bm{\Gamma}_i$ is given by $\bm{\Gamma}_i = \beta \bm{\Sigma} \bm{\Psi}_p^T \bm{\Xi}_i$, where $\bm{\Xi}_i$ is defined in Eq.~\eqref{eq:derivQInvHerm}. 
The full derivation of the partial derivatives with respect to the conjugate denominator coefficients is given in appendix \ref{app:grad_q}.
Once $\mathbf{q}^\ast$ is known, we can write Eq.~\eqref{eq:hyper_MAP} as
\begin{equation}
	[ \bm{\alpha}_p^\ast, \bm{\alpha}_q^\ast, \beta^\ast ] = 
	\argmax_{\underset{\in \mathbb{R}^{n_p \times n_q \times 1}}{[ \bm{\alpha}_p, \bm{\alpha}_q, \beta ]}} \left[ 
	\ln \det \bm{\Sigma} 
	% - \ln \det \left( \bm{\Lambda}_{pp} + \beta \bm{\Psi}^H \bm{\Psi} \right)
	+ N \ln \beta 
	+ \ln \det \bm{\Lambda}_{pp} 
	+ \ln \det \bm{\Lambda}_{qq} 
	- \beta \mathbf{y}^H \left(  \mathbf{y} - \bm{\Psi} \bm{\mu} \right) 
	- \mathbf{q}^{\ast H} \bm{\Lambda}_{qq} \mathbf{q}^\ast 
	\right]  \, .
	\label{eq:hyper_MAP2}
\end{equation}
It should be noted that $\mathbf{q}^\ast$ also enters $\bm{\Sigma}$ and $\bm{\mu}$ via Eqs.~\eqref{eq:post_cov_p} and \eqref{eq:post_mean_p}.

Again, we investigate the terms appearing in Eq.~\eqref{eq:hyper_MAP2} in more detail.
The first four terms in Eq.~\eqref{eq:hyper_MAP2} enter positively in the objective function. 
While an interpretation of $\ln \det \bm{\Sigma}$ ist not straightforward, the other three terms simply add the logarithm of the hyperparameters $\beta$, $\bm{\alpha}_{p}$ and $\bm{\alpha}_{q}$. 
Thus the value of the objective function is increased with increasing hyperparameters.
On the other hand side, the last two terms enter negatively in the objective function, where again, $- \beta \mathbf{y}^H \left(  \mathbf{y} - \bm{\Psi} \bm{\mu} \right)$ penalizes the misfit between the data and the RA, and the last term penalizes the quadratic form in $\mathbf{q}$.
The derivatives of the log-objective function with respect to the hyperparameters can be found in appendices \ref{app:grad_ap}, \ref{app:grad_aq} and \ref{app:grad_beta}.
The update rules for all three hyperparmaters read
\begin{gather}
	\alpha_{p,i} = \frac{1}{\Sigma_{ii} + \abs{\mu_i}^2}  \, , \label{eq:upd_rule_alpha_p}\\
	\alpha_{q,i} = \frac{1}{\abs{q_i^\ast}^2} \, , \label{eq:upd_rule_alpha_q} \\
	\beta = \frac{N}{\norm{\mathbf{y} - \bm{\Psi} \bm{\mu}}^2 + \tr \left( \bm{\Sigma} \bm{\Psi}^H \bm{\Psi} \right)} \label{eq:upd_rule_beta} \, ,
\end{gather}
where $\Sigma_{ii}$ denotes the $i$-th diagonal entry of the posterior covariance matrix $\bm{\Sigma}$ and $\mu_i$ denotes the $i$-th entry of posterior mean vector $\bm{\mu}$.
The result in Eqs.~\eqref{eq:upd_rule_alpha_p} and \eqref{eq:upd_rule_beta} are in close resemblance with the results for real-valued, linear Gaussian models in \cite{Tipping.2001}.
We note that the update rule for the denominator precisions solely depends on the corresponding coefficient magnitude.
Furthermore, we observe that the quadratic form in Eq.~\eqref{eq:hyper_MAP2}, $\mathbf{q}^{\ast H} \bm{\Lambda}_{qq} \mathbf{q}^\ast$, returns the number of selected denominator coefficients under the derived update rule in Eq.~\eqref{eq:upd_rule_alpha_q}, since 
\begin{equation}
	\mathbf{q}^{\ast H} \bm{\Lambda}_{qq} \mathbf{q}^\ast = \sum_{i=1}^{n_q} \alpha_{q,i} \abs{q_i^\ast}^2 = \sum_{i=1}^{n_q} 1 = n_q \, .
\end{equation}
This term acts as a regularizer that penalizes the number of polynomial terms in the denominator polynomials and leads to a sparse solution.
Following \cite{Tipping.2001}, after each update step, we prune the numerator and denominator basis terms from the expansions if they exceed a certain pruning threshold $\alpha_{\max,p}$ or $\alpha_{\max,q}$. 
The procedure of subsequent estimation of the MAP-parameters and pruning is repeated until convergence or after the maximum number of iteration steps has been reached. 
To check convergence, we monitor both the maximum change in the logarithm of all coefficient precisions, i.e., $\max \left( \left[ \Delta \log \bm{ \alpha}_p; \Delta \log \bm{\alpha}_q \right] \right)$ and the change in the logarithm of the likelihood precision $\Delta \log \beta$. 
We terminate the algorithm once both values fall below a pre-defined threshold.
From our experience, the likelihood precision convergences faster than the coefficient precisions.
Finally, a sparse set of coefficients for the numerator and denominator polynomials is obtained.

It was observed in some of the numerical investigations that the magnitude of the coefficients of the rational approximation becomes very small and thus the precisions become very large.
The reason for this is the fact that the numerator and denominator coefficients can be arbitrarily scaled by the same complex number, without altering the rational model output.
Therefore, the coefficients in the expansions can be made arbitrarily small. 
Due to the prescribed preference for small coefficients that is encoded in the prior distribution, it can happen that the algorithm results in coefficients of very small magnitude.
Since our pruning rule is indirectly based on the denominator coefficient magnitude this can cause pruning of all terms. 
We normalize the denominator coefficients after each update in order to avoid this behavior through dividing them with the classical $k$-norm $\norm{\mathbf{q}}_k$. 
In our numerical investigations we investigated using $k=2$ and $k=\infty$, which results in the maximum absolute coefficient.
Both approaches lead to meaningful results.

The full algorithm is summarized in Alg.~\ref{alg:SBLRA}.
We use the least-squares solution in Section~\ref{sec:lsq} as the initial coefficients for the algorithm. 
Alternatively, the coefficients could be randomly sampled from the prior complex Gaussian distribution.
\begin{algorithm}
	\caption{Sparse Bayesian Rational Approximation} 
	\label{alg:SBLRA}
	\begin{algorithmic}
		\Require The maximum number of iterations, $i_{\max}$, the tolerances for the convergence criteria $\varepsilon_\alpha$ and $\varepsilon_\beta$, the pruning thresholds $\alpha_{p,\max}$ and $\alpha_{q,\max}$, the initial coefficients $ \mathbf{p}_{\mathrm{init}}$ and $\mathbf{q}_{\mathrm{init}}$, $k$ used in $k$-normalization of $\mathbf{q}$
		\While $i < i_{\max} \land (\max(\Delta \log \bm{\alpha}) > \varepsilon_\alpha \lor \Delta \log \beta > \varepsilon_\beta)$
		\State Find useful numerator weights $p_i$ with $\alpha_{p,i}$ lower than threshold $\alpha_{p,\max}$, prune all other basis functions.
		\State Update denominator coefficients $\mathbf{q}$ with normalized MAP-estimate, $\mathbf{q} \leftarrow \mathbf{q}^{\ast}\cdot \norm{\mathbf{q}^{\ast}}_k^{-1}$, by solving Eq.~\eqref{eq:q_MAP}.
		\State Find useful denominator coefficients $q_i$ with $\alpha_{q,i}$ lower than threshold $\alpha_{q,\max}$, prune all other basis functions.
		\State Update posterior mean and covariance for numerator coefficients, $\bm{p} \leftarrow \bm{\mu} (\bm{q}^{\ast})$,  $\bm{\Sigma} \leftarrow \bm{\Sigma} (\bm{q}^{\ast})$ using Eqs.~\eqref{eq:post_cov_p} and \eqref{eq:post_mean_p}.
		\State Update $\bm{\alpha}_p$, $\bm{\alpha}_q$ and $\beta$ using the update rules in Eqs.~\eqref{eq:upd_rule_alpha_p}, \eqref{eq:upd_rule_alpha_q} and \eqref{eq:upd_rule_beta}
		\State $\Delta \log \alpha_{\max,i} \leftarrow \Delta \log \alpha_{\max,i+1}$
		\EndWhile \\
		\Return Retained coefficients $\mathbf{p}$ and $\mathbf{q}$
		
	\end{algorithmic} 
\end{algorithm}
%
%
% \subsection{Real-Valued models} 
% \label{subsec:ra_real}
%
% 
%
\section{Numerical Examples}
\label{sec:res}
In this Section the performance of the proposed sparse Bayesian rational approximation is investigated on the basis of two numerical examples.
The first example is a simple algebraic frequency response function model of a single degree of freedom shear frame structure with seven input variables.
The second example is the frequency response function of a cross-laminated timber plate with eleven input variables, which is obtained from a finite element model.
In both cases the experimental design is generated with latin hypercube sampling (LHS).
We assess the approximation accuracy of the SBRA based on the relative empirical error as defined in appendix \ref{app:rel_emp_error}.
\subsection{Algebraic Model: Frequency Response of a Shear Frame Structure}
\begin{figure}[!hbt]
	\centering
%	\includesvg{figures/Frame_structure}
	\begingroup%
	\makeatletter%
	\providecommand\color[2][]{%
		\errmessage{(Inkscape) Color is used for the text in Inkscape, but the package 'color.sty' is not loaded}%
		\renewcommand\color[2][]{}%
	}%
	\providecommand\transparent[1]{%
		\errmessage{(Inkscape) Transparency is used (non-zero) for the text in Inkscape, but the package 'transparent.sty' is not loaded}%
		\renewcommand\transparent[1]{}%
	}%
	\providecommand\rotatebox[2]{#2}%
	\newcommand*\fsize{\dimexpr\f@size pt\relax}%
	\newcommand*\lineheight[1]{\fontsize{\fsize}{#1\fsize}\selectfont}%
	\ifx\svgwidth\undefined%
	\setlength{\unitlength}{212.5984252bp}%
	\ifx\svgscale\undefined%
	\relax%
	\else%
	\setlength{\unitlength}{\unitlength * \real{\svgscale}}%
	\fi%
	\else%
	\setlength{\unitlength}{\svgwidth}%
	\fi%
	\global\let\svgwidth\undefined%
	\global\let\svgscale\undefined%
	\makeatother%
	\begin{picture}(1,0.73333333)%
		\lineheight{1}%
		\setlength\tabcolsep{0pt}%
		\put(0,0){\includegraphics[width=\unitlength,page=1]{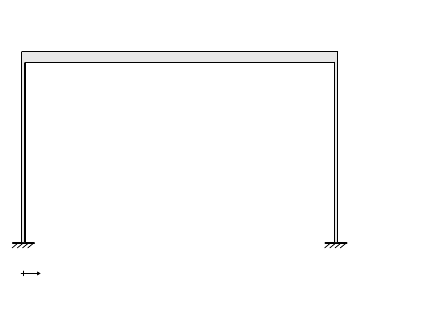}}%
		\put(0.09525,0.12302781){\color[rgb]{0,0,0}\makebox(0,0)[lt]{\lineheight{1.25}\smash{\begin{tabular}[t]{l}$u_b(t)$\end{tabular}}}}%
		\put(0,0){\includegraphics[width=\unitlength,page=2]{Frame_structure_svg-tex.pdf}}%
		\put(0.09525,0.66983329){\color[rgb]{0,0,0}\makebox(0,0)[lt]{\lineheight{1.25}\smash{\begin{tabular}[t]{l}$u(t)$\end{tabular}}}}%
		\put(0,0){\includegraphics[width=\unitlength,page=3]{Frame_structure_svg-tex.pdf}}%
		\put(0.38682685,0.06305573){\color[rgb]{0,0,0}\makebox(0,0)[lt]{\lineheight{1.25000012}\smash{\begin{tabular}[t]{l}$l$\end{tabular}}}}%
		\put(0,0){\includegraphics[width=\unitlength,page=4]{Frame_structure_svg-tex.pdf}}%
		\put(0.93363249,0.36660284){\color[rgb]{0,0,0}\makebox(0,0)[lt]{\lineheight{1.25000012}\smash{\begin{tabular}[t]{l}$h$\end{tabular}}}}%
		\put(0.07672222,0.37365838){\color[rgb]{0,0,0}\makebox(0,0)[lt]{\lineheight{1.25}\smash{\begin{tabular}[t]{l}$E_c, I_c$\end{tabular}}}}%
		\put(0.47398757,0.64049133){\color[rgb]{0,0,0}\makebox(0,0)[lt]{\lineheight{1.25}\smash{\begin{tabular}[t]{l}$A_g,\rho_g$\end{tabular}}}}%
		\put(0,0){\includegraphics[width=\unitlength,page=5]{Frame_structure_svg-tex.pdf}}%
		\put(0.8008055,0.12302781){\color[rgb]{0,0,0}\makebox(0,0)[lt]{\lineheight{1.25}\smash{\begin{tabular}[t]{l}$u_b(t)$\end{tabular}}}}%
		\put(0.61830499,0.37365838){\color[rgb]{0,0,0}\makebox(0,0)[lt]{\lineheight{1.25}\smash{\begin{tabular}[t]{l}$E_c, I_c$\end{tabular}}}}%
	\end{picture}%
	\endgroup%
	\caption{Sketch of the frame structure. It is parameterized by the column's Young's modulus $E_c$, the column's inertial moment $I_c$, the girder's mass density $\rho_g$, the girder's cross section $A_g$, the column's height $h$ and the girder's length $l$ as well as the overall system loss factor $\eta$.}
	\label{fig:fig_frame_1DOF}
\end{figure}

\begin{figure}[!hbt]
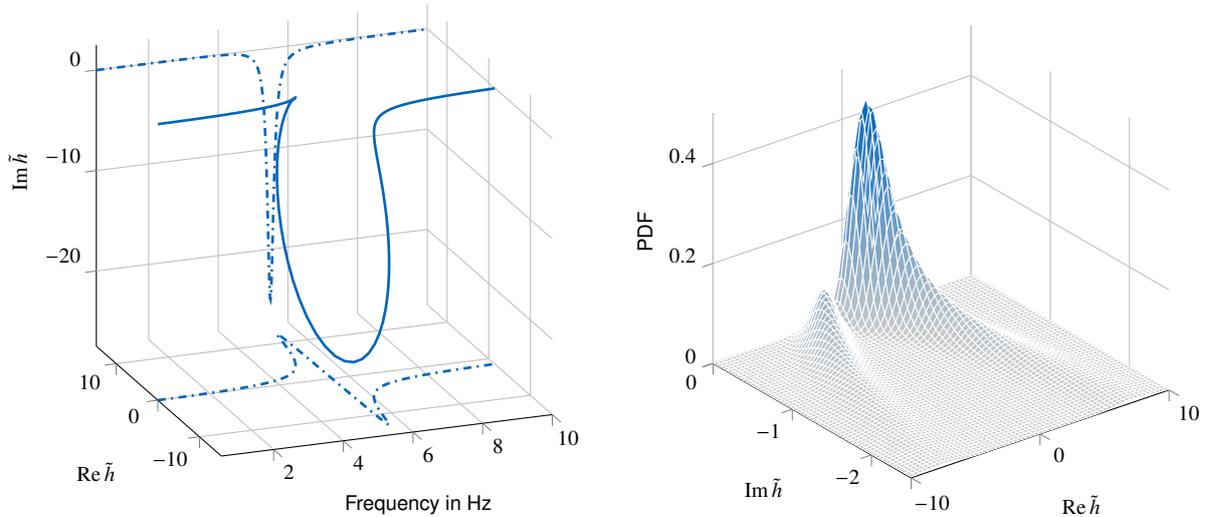
%		$$$$$$$$$$$$$$$ --> see "`MISC/PLOT_Exp_Designs.m"'
%	\TikzQuadrXS
	\setlength\fheight{6cm}
	\setlength\fwidth{6cm}
	% \centering
	\strut\hfill
	\subfloat[Real ($\Re{\Tilde{h}}$) and imaginary part ($\Im{\Tilde{h}}$) of the transfer function evaluated for the nominal input parameters. The dashed lines depict the projection in the real and imaginary axis, respectively. \label{fig:nominal_transfer_function_frame_re_im}]{
		\input{frame_SDOF_3d_re_im.tikz}
	}
	\hfill
	\subfloat[Estimate of the joint probability density function of the real and imaginary part of the frame model response obtained through kernel density estimation based on a set of $10^5$ samples. \label{fig:frame_resp_pdf_test_set}]{
		\input{2d_pdf_re_im.tikz}
	} 
	\strut\hfill
	% \subfloat[Absolute Value ($\abs{\Tilde{h}}$) and phase ($\angle{\Tilde{h}}$) of the transfer function. \label{fig:nominal_transfer_function_frame_abs_arg}]{\input{figures/frame/frame_SDOF_3d_abs_arg.tikz}} 
	\caption[]{Illustration of the share frame model model response.}
	\label{fig:nominal_transfer_function_frame}
\end{figure}

For the following section, the frequency response function of a shear frame, relating the girder displacement $u(t)$ to the base displacements $u_b (t)$, is considered. 
The system is illustrated in Fig.~\ref{fig:fig_frame_1DOF}.
We follow a standard modelling approach and consider that the stiffness of the system is solely contributed by the columns, while the columns' masses are neglected.
Furthermore the girder is assumed to be rigid, such that left and right column displacements are equal.
Then, the system can be considered as a single degree of freedom system and the stiffness of the structure is given by $k = 24 EI_c h^{-3}$, where $EI_c$ is the bending stiffness of one column, and $h$ is the storey height. 
The system mass is $m = \rho A_g l$, where $\rho$, $A_g$ and $l$ are the girder density, cross-sectional area and length, respectively.
Then, the frequency response function $\Tilde{h}$ is given by
\begin{align}
	\Tilde{h}  : \mathbb{R} \times \mathbb{R}^d & \longrightarrow \mathbb{C}  \nonumber \\
	\omega \times \mathbf{x} & \longmapsto \frac{\omega^2}{\omega_n^2 - \omega^2 + \mathrm{i} \eta \sign(\omega) \omega_n^2} \, .
	\label{eq:frame_standard_FRF_linhyst}
\end{align}
In there, $\omega_n = \sqrt{km^{-1}}$ is the natural frequency of the system, $\eta$ is the frequency-independent loss factor of the system and $\sign(\cdot)$ denotes the signum-function.
The random vector $\mathbf{X}$ collects all parameters of the model, i.e., 
\begin{equation}
	\mathbf{X} = \left[ E, I_c, h, \rho, A_g, l, \eta \right] \, .
\end{equation}
The parameters are assumed to be independent and lognormally distributed.
The individual mean values and coefficients of variation are given in Tab.~\ref{tab:table_frame_nom_para}.
\begin{table}[!hbt]
	\begin{center}
		{
			%        	\small
			\caption{Frame structure's distribution parameter values.}
			\centering
			\begin{tabular}{lccc}
				\toprule
				Parameter & & Mean value & Coefficient of variation \\ \midrule
				Columns' Young's modulus & $E$ & $3 \cdot 10^{10} \, \mathrm{\frac{N}{m^2}}$ & $0.1$ \\
				Columns' moment of inertia & $I_c$ & $\frac{\pi \left( 0.3 \mathrm{m} \right)^4}{64}$ & $0.1$ \\
				Columns' height & $h$ & $4 \, \mathrm{m}$ & $0.1$ \\ \midrule
				Girder's density & $\rho$ & $2.5 \cdot 10^{3} \, \mathrm{kg}\mathrm{m}^{-3}$ & $0.05$ \\
				Girder's cross-sectional area & $A_g$ & $0.3 \, \mathrm{m} \cdot 0.5 \, \mathrm{m}$ & $0.1$\\
				Girder's length & $l$ & $10 \, \mathrm{m}$ & $0.1$ \\ \midrule
				Loss Factor & $\eta$ & $0.04$ & 0.3 \\
				\bottomrule
			\end{tabular}
			\label{tab:table_frame_nom_para}
		}
	\end{center}
\end{table}
The nominal eigenfrequency, based on the mean values of the input parameters is $\omega_n = 34.5 \, \mathrm{rad} \mathrm{Hz}$, or equivalently $f_n = 5.5 \, \mathrm{Hz}$. 
The nominal transfer function, based on the mean parameter values, is depicted in Fig.~\ref{fig:nominal_transfer_function_frame}. 
One can clearly observe the rational dependency on the frequency. 
The imaginary part of the transfer function is significantly non-zero only in the immediate vicinity of the nominal eigenfrequency.
Despite the fact that the model output cannot be visualized over the input space, since all parameters enter in the denominator of the transfer function, a similar rational dependency can be expected.

\begin{figure}[!hbt]
	\centering
	\setlength\fheight{4.5cm}
	\setlength\fwidth{14cm}
	\centering
	\subfloat[Case 1: $m_p = m_q = 10$, and $q_p = q_q = 0.5$]{\input{frame_rel_emp_error_10_10.tikz}} \\
	\subfloat[Case 2: $m_p = m_q = 5$ and $q_p = q_q = 0.7$]{\input{frame_rel_emp_error_5_5.tikz}} \\
	\subfloat[Case 3: $m_p = m_q = 3$ and $q_p = q_q = 1$]{\input{frame_rel_emp_error_3_3.tikz}}
	\caption{Survey on the relative empirical error of the sparse rational approximation applied to the frame model for varying numbers of samples in the experimental design. The boxplot depicts the interquartile range (IQR), defined by the 25\%- and 75\%-quantiles $q_1$ and $q_3$ as blue boxes (\BlockBlau). together with the median as black marker ($\bullet$) as well as the full blue line (\linefullblue). Blue circles (\CircBlau) denote outliers in the errors. Small horizontal shifts in the outliers are only for better readability. The blue vertical lines show the range of all data points except the outliers. Outliers are defined as all samples that are larger than $q_3 + 1.5 IQR$ or lower than $q_1 - 1.5 IQR$. Furthermore, the median relative empirical error of the least-squares solution is given (\linedashedgreen). The analysis is based on 50 different experimental designs.}
	\label{fig:SBL_res_frame_box_MSE}
\end{figure}
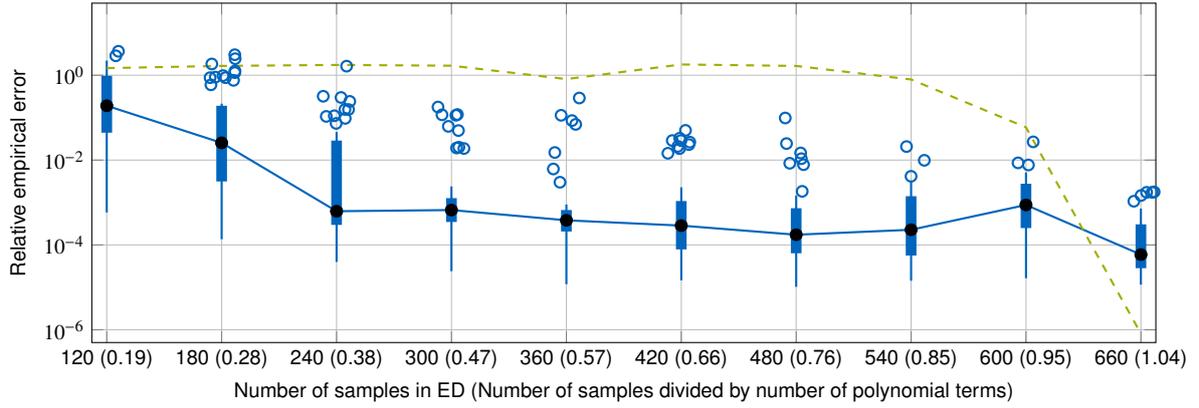
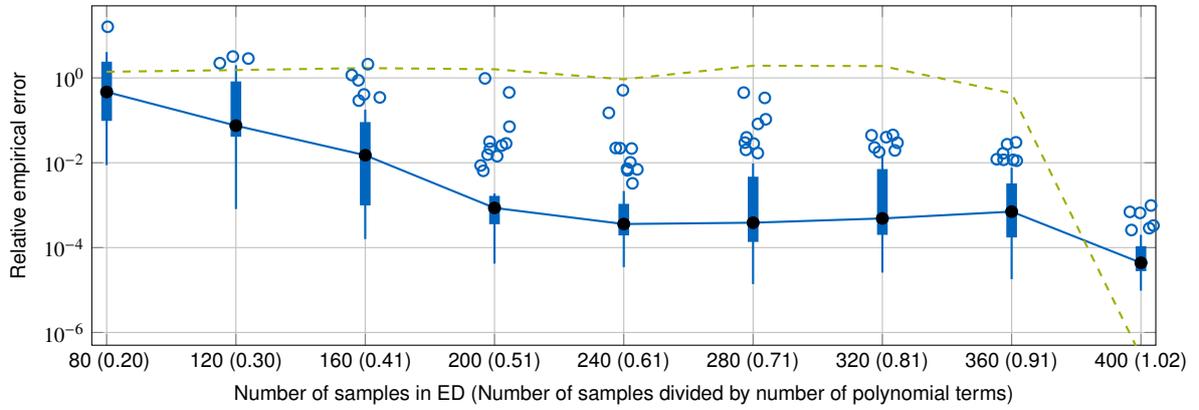
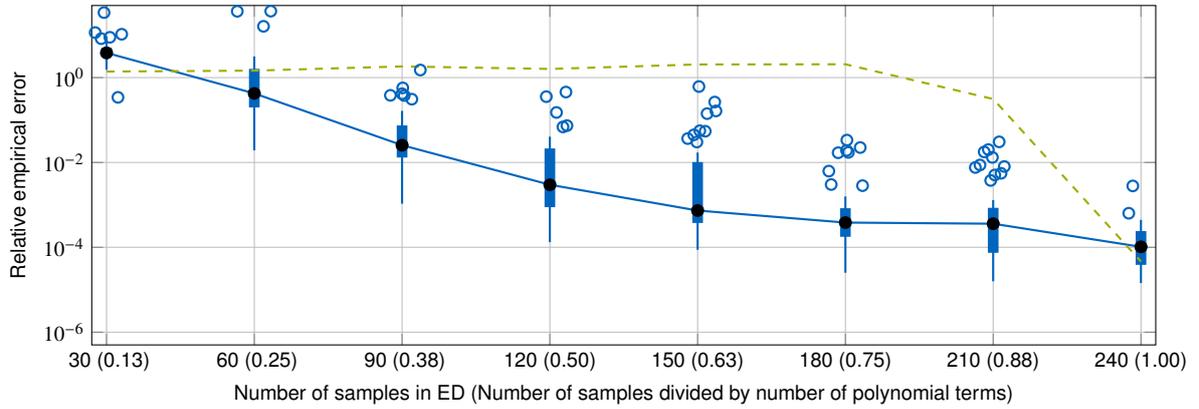

The model is investigated for the frequency $f = 5.1 \, \mathrm{Hz}$. 
Fig.~\ref{fig:frame_resp_pdf_test_set} depicts the joint PDF of the real and imaginary part of the model response. 
The PDF is obtained through kernel density estimation, based on $10^5$ samples from the input random variables.
One can observe that the PDF exhibits a clear bi-modality. 
%
% \begin{figure}[!hbt]%		$$$$$$$$$$$$$$$ --> see "`MISC/PLOT_Exp_Designs.m"'
	% \TikzQuadrXS
	% \setlength\fheight{6cm}
	% \setlength\fwidth{8cm}
	% \centering
	% \input{figures/frame/2d_pdf_re_im.tikz}
	% \caption[]{Joint probability density function of the real and imaginary part of the frame model response based on a set of $10^5$ samples.}
	% \label{fig:frame_resp_pdf_test_set}
	% \end{figure}
%
We use a pruning threshold of $\alpha_{\max} = 10^6$ in this section.
Different pruning thresholds have been investigated in an initial study, which showed that $\alpha_{\max} \in [ 10^6, 10^9]$ leads to similar performance for the investiaged problems.
Furthermore, three different basis sets are investigated:
\begin{itemize}
	\item Case 1: Maximum polynomial degrees $m_p = m_q = 10$, and truncation degrees $q_p = q_q = 0.5$
	\item Case 2: Maximum polynomial degrees $m_p = m_q = 5$ and truncation degrees $q_p = q_q = 0.7$
	\item Case 3: Maximum polynomial degrees $m_p = m_q = 3$ and truncation degrees $q_p = q_q = 1$
\end{itemize}
The total numbers of polynomial terms are $N_{\mathrm{pol}} = 632$, $N_{\mathrm{pol}} = 394$ and $N_{\mathrm{pol}} = 240$ for the first, second and third case, respectively. 
We investigate the performance of the method for different experimental design sizes and therefore choose $N$ from $80$ to $440$ in steps of $40$ for the first case, $N$ from $30$ to $270$ in steps of $30$ for the second case and $N$ from $30$ to $240$ in steps of $30$ for the third case.
Thus, for the largest number of samples in the experimental design, we approximately have as many samples as unknowns in the problem for the three cases.
The parameters used in the construction of the surrogate model are summarized in Tab.~\ref{tab:par_surr_frame_models}.
\begin{samepage}
	\begin{table}
		{
			\begin{center}
				%    	\footnotesize%of{table}
				\caption{Surrogate parameters in frame model}
				\begin{tabular}{lcccc}
					\toprule
					Parameter & & Case 1 & Case 2 & Case 3 \\ \midrule
					Maximum polynomial degrees & $m_p = m_q$ & $10$ & $5$ & $3$ \\[1mm]
					Hyperbolic truncation parameter & $q_p = q_q$ & $0.5$ & $0.7$ & $1$ \\[1mm]
					Number of polynomial terms & $N_p = N_q$ & $316$ & $197$ & $120$ \\[1mm]
					Number of samples & $N$ & $\{120 , 60, \ldots, 660 \}$ & $\{80 , 40, \ldots, 400 \}$ & $\{30 , 30, \ldots, 240 \}$ \\
					\bottomrule
				\end{tabular}
				\label{tab:par_surr_frame_models}
			\end{center}	
		}
	\end{table}
\end{samepage}
In Fig.~\ref{fig:SBL_res_frame_box_MSE} we depict the relative empirical error in terms of the number of samples in the experimental design for each of the three basis sets. 
The analysis is repeated 50 times and the results are summarized in form of a box plot.
Furthermore, the median relative empirical error of the least-squares solution, as presented in Section \ref{sec:lsq}, is plotted.
It can be observed that the sparse Bayesian approach always gives lower errors in median than the least-squares approach whenever the number of samples is smaller than the number of basis terms. 
For each of the basis sets, we observe a strong decrease in the error measure with increasing sample size.
The lowest median relative empirical errors occur for the largest number of samples in each case.
We note that the decrease is not monotonic for the cases 1 and 2. 
% For the first case, the maximum median relative empirical error occurs at $N=80$ with $\varepsilon_{\mathrm{emp},0.5} = 0.62$ and the minimum median relative empirical error occurs at $N=440$ with $\varepsilon_{\mathrm{emp},0.5} = 3.1 \cdot 10^{-5}$. 
While the difference between the least-squares and the SBRA error measure is rather small for a very low number of samples, the difference in the error measure is significant whenever the sample size is around 40\% of the number of polynomial terms.
A difference of around 3 orders of magnitude in median between the least-squares and the SBRA solution can be observed in this case.
The median relative empirical error tends towards around $10^{-4}$ for increasing sample size in all cases.
While a basis set with low-degree polynomials appears to be sufficient for the present model, the SBRA is nevertheless able to extract the relevant basis functions for basis sets with larger polynomial degree. 
Furthermore, in case 1, the error decreases fast for increasing $N$ and then ranges in the order of $10^{-4}$, while for the other two cases, the error decreases more steadily for increasing $N$.
When the number of data points is larger than the number of polynomial terms, the least-squares solution yields lower errors compared to the sparse Bayesian approach. 
Since we are mainly interested in the low-data case, we do not further compare both methods for larger $N$.

\begin{figure}[!hbt]
	\centering
	\setlength\fheight{5cm}
	\setlength\fwidth{7cm}
	\centering
	\input{comp_emp_error_median_lsq_randn.tikz}
	\caption{
		Comparison of the median relative empirical errors for the three basis sets and different initial points. 
		Blue lines (\linefullblue) represent case 1 ($m_p = m_q = 10$), orange lines (\linefullorange) represent case 2 ($m_p = m_q = 5$), and green lines (\linefullgreen) represent case 3 ($m_p = m_q = 3$). 
		Solid lines (---) represent the solutions based on the initial points stemming from the least squares solution, dash-dotted lines (--$\cdot$--) are based on random initial points sampled from a proper standard complex normal distribution. 
		It should be noted that the experimental designs for cases 1 to 3 are independent of each other, while the same experimental design is used for the two different initial points.
	}
	\label{fig:SBL_res_frame_box_MSE_basis_sets_lsq_rand_comp}
\end{figure}
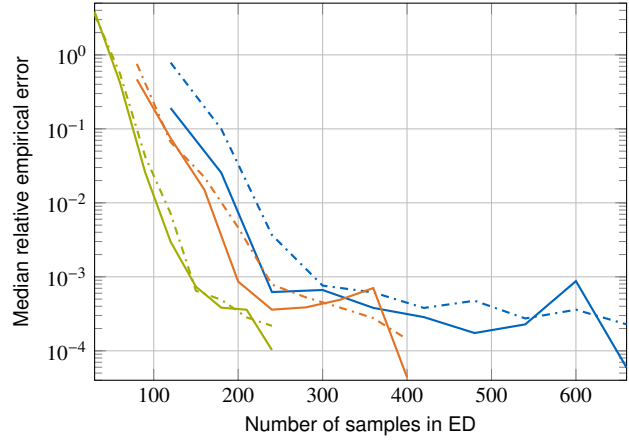

In Fig.~\ref{fig:SBL_res_frame_box_MSE_basis_sets_lsq_rand_comp}, we compare the median relative empirical errors for the three basis sets. 
Furthermore, we additionally evaluate the errors using random initial points for the RA coefficients.
We observe that when using a very small sample size, the solution based on the smallest basis set yields the lowest relative empirical error. 
However, for all three basis sets we achieve a significant decrease in the relative empirical error with increasing $N$.
The figure highlights the ability of the proposed SBRA to identify the important basis functions even among polynomials with high degree. 
We furthermore observe that the small increase in the relative empirical error for cases 1 and 2 around $N=600$ and $N=360$, respectively, is only present when we use the least squares solution as initial points.  
We conclude that for medium sample sizes it might be advantageous to use a different initial point. 
In contrast, for the low sample sizes, using the least squares solution as initial point, we obtain lower errors for cases 2 and 3.
In practice, we suggest to run the algorithm with different initial points to increase the robustness of the method and choose the model with the lowest error.
If no test set is available on can estimate the error based on cross-validation, see, e.g., \cite{pml1Book}.
%Furthermore, we observe that for a low fixed number of data points, the approximation with the lower maximum polynomial degree results in the lower error measures.
%Eventually the errors tend towards the same order of magnitude for all cases.

% \begin{figure}[hbt]
	%     \centering
	%     \setlength\fheight{5cm}
	% 	\setlength\fwidth{14cm}
	% 	\centering
	%     \input{frame_rel_emp_error_5_5.tikz}
	%     \caption{Survey on the relative empirical error of the sparse rational approximation applied to the frame model for varying numbers of samples in the experimental design.}
	%     \label{fig:SBL_res_frame_box_MSE}
	% \end{figure}

\begin{figure}[!hbt]
	\centering
	\setlength\fheight{5cm}
	\setlength\fwidth{6cm}
	\centering
	\strut\hfill
	\subfloat[Initial point from least-squares solution. \label{fig:SBL_res_frame_mean_DOS_lsq}]{\input{frame_DOS_mean_5_5.tikz}} \hfill
	\subfloat[Random initial point. \label{fig:SBL_res_frame_mean_DOS_rand}]{\input{frame_DOS_mean_5_5_rand.tikz}} 
	\strut\hfill
	\caption{Mean value of the degree of sparsity for case 2 ($m_p = m_q = 5$); total model (\linefullblue), the numerator polynomial (\linedashedred) and the denominator polynomial (\linedashdotteddarkgreen) and corresponding $90\%$-credible intervals for case 2, i.e., polynomial degrees $m_p = m_q = 5$, and truncation degrees $q_p = q_q =0.8$. The analysis is based on 50 different experimental designs.}
	\label{fig:SBL_res_frame_mean_DOS}
\end{figure}
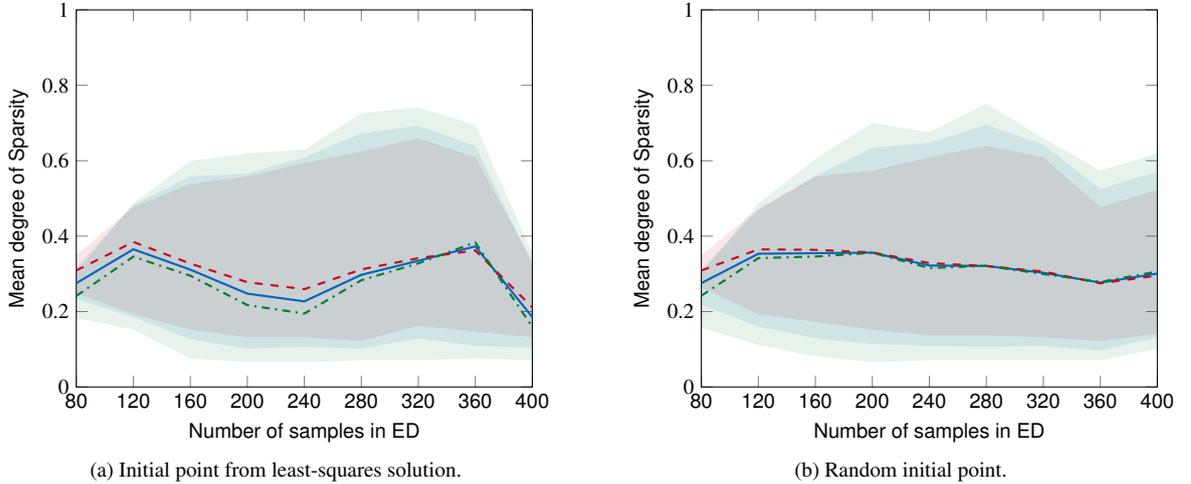

In Fig.~\ref{fig:SBL_res_frame_mean_DOS} we depict the degree of sparsity for case 2 in terms of the number of samples for two different initial points. 
The results in Fig.~\ref{fig:SBL_res_frame_mean_DOS_lsq} are based on the least-squares coefficients, while the results in Fig.~\ref{fig:SBL_res_frame_mean_DOS_rand} are based on a random sample.
The degree of sparsity is defined as the ratio of the number of retained basis terms and the total number of basis terms.
We evaluate the overall, total degree of sparsity, as well as the individual degrees of sparsity for the numerator and denominator separately.
In general, less than 40\% of terms are retained in the rational model on average. 
The degree of sparsity is slightly lower in the denominator polynomial for the least-squares initial point and almost the same for the random initial point.
In Fig.~\ref{fig:SBL_res_frame_mean_DOS_lsq} we observe a slight increase in the degree of sparsity for $N=360$ that we already observed in the relative empirical error.
For the random initial point, the degree of sparsity is almost constant for all considered sample sizes.

\begin{figure}[!hbt]
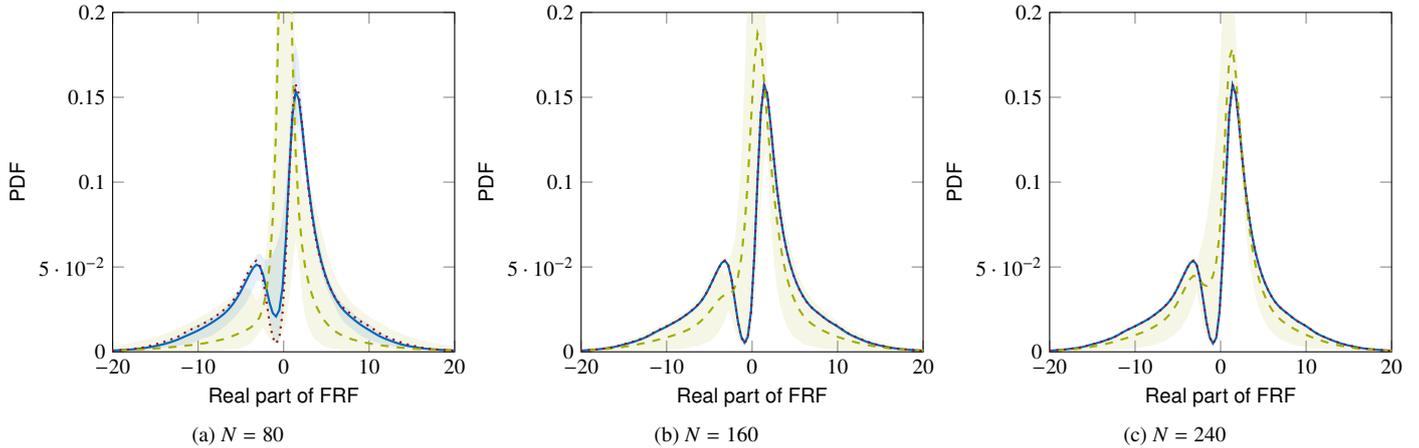

	\setlength\fheight{4.5cm}
	\setlength\fwidth{4.5cm}
	\centering
	\makebox[\textwidth][c]{
		\subfloat[$N = 80$]{\input{frame_pdf_y_re_N_080.tikz}} 
		\subfloat[$N = 160$]{\input{frame_pdf_y_re_N_160.tikz}} 
		\subfloat[$N = 240$]{\input{frame_pdf_y_re_N_240.tikz}} 
	}
	\caption{Probability density function of the real value of the frequency response function,  $\Re{\Tilde{h}}$ for case 2, i.e., polynomial degrees $m_p = m_q = 5$. Sparse approximation (\linefullblue) and corresponding 90 \% credible interval (\BlockHellBlau), least-squares approximation (\linedottedgreen) and corresponding 90 \% credible interval (\BlockHellGreen), Monte Carlo reference solution (\linedashedred).}
	\label{fig:SBL_res_frame_PDFs_real}
\end{figure}

Furthermore, we evaluate the marginal PDFs of the real part of the FRF for case 2 based on the test set and depict the results for the sample sizes $N=\{80, 160, 240\}$ in Fig.~\ref{fig:SBL_res_frame_PDFs_real}.
The depicted PDFs are averages based on the 50 repeated LHS experimental designs.
For each experimental design, the PDFs are based on $N_t = 10^5$ test set samples and obtained by kernel density estimation using the surrogate model. 
The reference solution is obtained by evaluating the model in Eq.~\eqref{eq:frame_standard_FRF_linhyst} for the test set samples. 
In addition to the average PDFs, the 90\%-credible intervals are depicted. 
From the results for the three different sample sizes, it can be observed that the SBRA is able to capture the bi-modality in the real part of the model response accurately already for a low number of samples.
Furthermore, the credible interval becomes very narrow from $N=160$ on, which shows that the model is able to reliably reproduce the density of the model response even for low sample sizes.
It can be observed that this is not the case for the approximation based on the least-squares solution, which converges slowly to the reference solution and exhibits a much higher spread in the resulting PDF.
%
% \begin{figure}[!hbt]
	% 	\setlength\fheight{4.5cm}
	% 	\setlength\fwidth{4.5cm}
	% 	\centering
	% 	\makebox[\textwidth][c]{
		% 			\subfloat[$N = 80$]{\input{frame_pdf_y_im_N_080.tikz}} 
		% 			\subfloat[$N = 160$]{\input{frame_pdf_y_im_N_160.tikz}} 
		% 			\subfloat[$N = 240$]{\input{frame_pdf_y_im_N_240.tikz}} 
		% 			}
	% 	\caption{Probability density function for absolute value of response $\abs{\Tilde{h}}$ for the first case, i.e., polynomial degrees $m_p = 10$, $m_q=3$ and truncation degrees $q_p = 0.5$, $q_q = 1$. Sparse approximation (\linefullblue) and corresponding 90 \% credible interval (\BlockBlau), least-squares approximation (\linedottedgreen), Monte Carlo reference solution (\linedashedred).}
	%     \label{fig:SBL_res_frame_PDFs_imag}
	% 	%
	% \end{figure}
%
Overall, it can be seen that the sparse rational approximation is able to accurately represent the original model. 
It results in much lower errors in comparison to the least-squares approach and successfully selects the relevant terms in the provided basis set.
\subsection{Finite Element Model: Frequency Response of an Orthotropic Plate}
\label{sec:res_fe}
\begin{figure}[!hbt]
	\centering
%	\includesvg{figures/plaate_with_acc}
	\begingroup%
	\makeatletter%
	\providecommand\color[2][]{%
		\errmessage{(Inkscape) Color is used for the text in Inkscape, but the package 'color.sty' is not loaded}%
		\renewcommand\color[2][]{}%
	}%
	\providecommand\transparent[1]{%
		\errmessage{(Inkscape) Transparency is used (non-zero) for the text in Inkscape, but the package 'transparent.sty' is not loaded}%
		\renewcommand\transparent[1]{}%
	}%
	\providecommand\rotatebox[2]{#2}%
	\newcommand*\fsize{\dimexpr\f@size pt\relax}%
	\newcommand*\lineheight[1]{\fontsize{\fsize}{#1\fsize}\selectfont}%
	\ifx\svgwidth\undefined%
	\setlength{\unitlength}{315bp}%
	\ifx\svgscale\undefined%
	\relax%
	\else%
	\setlength{\unitlength}{\unitlength * \real{\svgscale}}%
	\fi%
	\else%
	\setlength{\unitlength}{\svgwidth}%
	\fi%
	\global\let\svgwidth\undefined%
	\global\let\svgscale\undefined%
	\makeatother%
	\begin{picture}(1,0.73809524)%
		\lineheight{1}%
		\setlength\tabcolsep{0pt}%
		\put(0,0){\includegraphics[width=\unitlength,page=1]{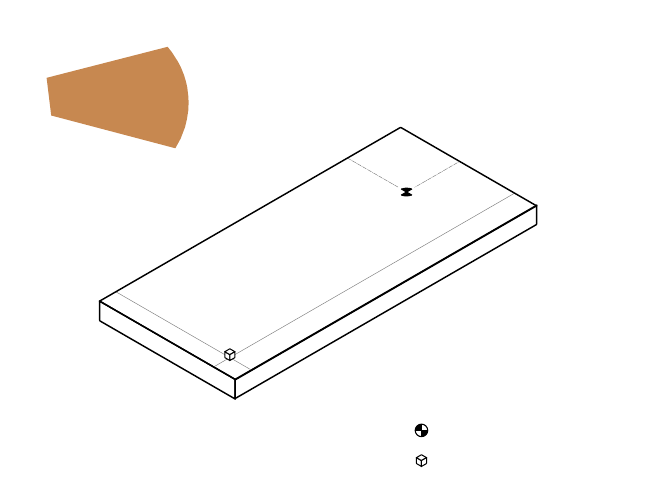}}%
		\put(0.68703671,0.02754124){\color[rgb]{0,0,0}\makebox(0,0)[lt]{\lineheight{0}\smash{\begin{tabular}[t]{l}Response location\end{tabular}}}}%
		\put(0.68703671,0.07233639){\color[rgb]{0,0,0}\makebox(0,0)[lt]{\lineheight{0}\smash{\begin{tabular}[t]{l}Excitation point\end{tabular}}}}%
		\put(0,0){\includegraphics[width=\unitlength,page=2]{plaate_with_acc_svg-tex.pdf}}%
		\put(0.76004943,0.17501507){\color[rgb]{0,0,0}\makebox(0,0)[lt]{\lineheight{0}\smash{\begin{tabular}[t]{l}$2.5$\end{tabular}}}}%
		\put(0.4473389,0.06948223){\color[rgb]{0,0,0}\makebox(0,0)[lt]{\lineheight{0}\smash{\begin{tabular}[t]{l}$0.1$\end{tabular}}}}%
		\put(0.74046405,0.25393062){\color[rgb]{0,0,0}\makebox(0,0)[lt]{\lineheight{0}\smash{\begin{tabular}[t]{l}$2.4$\end{tabular}}}}%
		\put(0.09761905,0.10858837){\color[rgb]{0,0,0}\makebox(0,0)[lt]{\lineheight{0}\smash{\begin{tabular}[t]{l}$1.1$\end{tabular}}}}%
		\put(0.1467687,0.16896245){\color[rgb]{0,0,0}\makebox(0,0)[lt]{\lineheight{0}\smash{\begin{tabular}[t]{l}$1.0$\end{tabular}}}}%
		\put(0.46868421,0.56331714){\color[rgb]{0,0,0}\makebox(0,0)[lt]{\lineheight{0}\smash{\begin{tabular}[t]{l}$0.33$\end{tabular}}}}%
		\put(0.71599513,0.55720054){\color[rgb]{0,0,0}\makebox(0,0)[lt]{\lineheight{0}\smash{\begin{tabular}[t]{l}$0.475$\end{tabular}}}}%
		\put(0,0){\includegraphics[width=\unitlength,page=3]{plaate_with_acc_svg-tex.pdf}}%
		\put(0.5554294,0.49015517){\color[rgb]{0,0,0}\makebox(0,0)[lt]{\lineheight{0}\smash{\begin{tabular}[t]{l}$x$\end{tabular}}}}%
		\put(0.66678082,0.52245396){\color[rgb]{0,0,0}\makebox(0,0)[lt]{\lineheight{0}\smash{\begin{tabular}[t]{l}$y$\end{tabular}}}}%
		\put(0.26227203,0.08312247){\color[rgb]{0,0,0}\makebox(0,0)[lt]{\lineheight{0}\smash{\begin{tabular}[t]{l}$0.1$\end{tabular}}}}%
	\end{picture}%
	\endgroup%
	\caption{Sketch of the plate model. It is parameterized by the nine orthotropic material constants, it's density $\rho$ and the loss factor $\eta$.}
	\label{fig:fig_LENO}
\end{figure}

In this section, the frequency response function of a cross-laminated timber plate is considered. 
The plate consists of three layers of crosswise glued timber and has the dimensions $l \times b \times t = 2.5 \, \mathrm{m} \times 1.1 \, \mathrm{m} \times 0.081 \, \mathrm{m}$. 
Each layer is $2.7 \, \mathrm{cm}$ thick.
The plate structure is depicted in Fig.~\ref{fig:fig_LENO}.

It is modelled as a three-dimensional orthotropic solid, where the orientation of each layer is considered.
Each layer is assigned the same stiffness value, i.e., all material parameters are constant throughout the plate domain. 
The cross-wise layering is accounted for by considering the local fiber directions in each layer.

In general, for a discrete finite element system in linear dynamics, the frequency response function $\Tilde{h}$, describing the acceleration at degree of freedom (DOF) $i$ due to a unit-force at DOF $j$, in radial frequency space $\omega$ is defined by
\begin{align}
	\Tilde{h}_{ij}  : \mathbb{R} \times \mathbb{R}^d & \longrightarrow \mathbb{C}  \nonumber \\
	\omega \times \mathbf{x} & \longmapsto - \omega^2 \mathbf{e}_{i}^T \left( \mathbf{K} (\mathbf{x})  + \mathrm{i} \omega \mathbf{C} (\mathbf{x}) - \omega^2 \mathbf{M} (\mathbf{x}) \right)^{-1} \mathbf{e}_{j}  \, ,
	\label{eq:LENO_transf_func}
\end{align}
where $\mathbf{K}$, $\mathbf{C}$ and $\mathbf{M}$ denote the stiffness, damping and mass matrix, respectively, and $\mathbf{e}_k$ denotes the single-entry unit vector that is one at entry $k$ and zero elsewhere.
The system matrices are obtained through a finite element approximation in the spatial domain. 
% Thereby, the coupled differential equation describing the motion of the medium is transformed to an ordinary differential equation in time that can be solved in the frequency domain for periodic loading.

The commercial finite element software ANSYS\textsuperscript{\textregistered} is used to solve the dynamic problem defined in Eq.~\eqref{eq:LENO_transf_func}.
We further choose a linear hysteretic damping model, as it supports a frequency-independent energy loss for steady state motion per cycle, which is a realistic assumption for many materials, including cross-laminated timber \cite{labonnote2012damping}. Under this model, the damping matrix can be expressed through
\begin{equation}
	\mathbf{C} (\mathbf{x}) = \frac{\eta}{\abs{\omega}} \mathbf{K} (\mathbf{x})
\end{equation}
The random vector $\mathbf{X}$ collects all parameters of the model, i.e., 
\begin{equation}
	\mathbf{X} = \left[ E_x, E_y, E_z, G_{xy}, G_{xz}, G_{yz}, \nu_{yx}, \nu_{zx}, \nu_{zy}, \rho, \eta \right]
\end{equation}
The parameters are assumed to be independent and lognormally distributed.
It should be noted that the assumption of independence does not necessarily hold in general \cite{schneider2022bayesian}, however, we consider it sufficient for the following numerical investigations.  
The individual mean values and coefficients of variation are given in Tab.~\ref{tab:table_LENO_nom_para}.
\begin{center}
	{
		%    	\small
		\captionof{table}{Distribution parameters for frame structure.}
		\centering
		\begin{tabular}{lccc}
			\toprule
			Parameter & & Mean value & Coefficient of variation \\ \midrule
			\multirow{3}{*}{Young's moduli} & $E_x$ & $1.1 \cdot 10^{10} \, \mathrm{\frac{N}{m^2}}$ & $0.1$ \\
			& $E_y$ & $0.85 \cdot 3.667 \cdot 10^{8} \, \mathrm{\frac{N}{m^2}}$ & $0.1$ \\
			& $E_z$ & $3.667 \cdot 10^{8} \, \mathrm{\frac{N}{m^2}}$ & $0.1$ \\ \midrule
			\multirow{3}{*}{Shear moduli} & $G_{xy}$ & $0.7 \cdot 6.9 \cdot 10^{8} \, \mathrm{\frac{N}{m^2}}$ & $0.1$ \\
			& $G_{xz}$ & $6.9 \cdot 10^{8} \, \mathrm{\frac{N}{m^2}}$ & $0.1$ \\
			& $G_{yz}$ & $6.9 \cdot 10^{7} \, \mathrm{\frac{N}{m^2}}$ & $0.1$ \\
			\midrule
			\multirow{3}{*}{Minor Poisson's ratios} & $\nu_{yx}$ & $0.014$ & $0.1$ \\
			& $\nu_{zx}$ & $0.014$ & $0.1$ \\
			& $\nu_{zy}$ & $0.3$ & $0.1$ \\
			\midrule
			Density & $\rho$ & $450 \, \mathrm{\frac{kg}{m^3}}$ & $0.1$ \\ \midrule
			Damping constant & $\eta$ & $0.04$ & 0.3 \\
			\bottomrule
		\end{tabular}
		\label{tab:table_LENO_nom_para}
	}
\end{center}

The model is investigated for the frequency $f = 90 \, \mathrm{Hz}$. 
Again, we use a pruning threshold of $\alpha_{\max} = 10^6$ and investigate three different basis sets:
\begin{itemize}
	\item Case 1: Maximum polynomial degrees $m_p = m_q = 10$, and truncation degrees $q_p = q_q = 0.5$
	\item Case 2: Maximum polynomial degrees $m_p = m_q = 5$ and truncation degrees $q_p = q_q = 0.7$
	\item Case 3: Maximum polynomial degrees $m_p = m_q = 3$ and truncation degrees $q_p = q_q = 1$
\end{itemize}
The total numbers of polynomial terms are $N_{\mathrm{pol}} = 632$, $N_{\mathrm{pol}} = 394$ and $N_{\mathrm{pol}} = 240$ for the first, second and third case, respectively. 
Again, we investigate the performance of the method for different experimental design sizes and therefore choose $N$ according to Tab.~\ref{tab:par_surr_LENO}. 

\begin{samepage}
	\begin{center}
		{
			%    	\footnotesize
			\captionof{table}{Surrogate parameters in plate finite element model}
			\begin{tabular}{lcccc}
				\toprule
				Parameter & & Case 1 & Case 2 & Case 3 \\ \midrule
				Maximum polynomial degrees & $m_p = m_q$ & $10$ & $5$ & $3$ \\[1mm]
				Hyperbolic truncation parameter & $q_p = q_q$ & $0.5$ & $0.7$ & $1$ \\[1mm]
				Number of polynomial terms & $N_p = N_q$ & $826$ & $551$ & $364$ \\[1mm]
				Number of samples & $N$ & $\{320 , 160, \ldots, 1600 \}$ & $\{240 , 120, \ldots, 1200 \}$ & $\{140 , 70, \ldots, 770 \}$ \\
				\bottomrule
			\end{tabular}
			\label{tab:par_surr_LENO}
		}
	\end{center}
\end{samepage}

\begin{figure}[!hbt]
	\centering
	\setlength\fheight{5cm}
	\setlength\fwidth{14cm}
	\centering
	\input{LENO_rel_emp_error.tikz}
	\caption{Survey on the relative empirical error of the sparse rational approximation applied to the plate finite element model for case 1 ($m_p = m_q = 10$) for varying numbers of samples in the experimental design. The boxplot depicts the interquartile range (IQR), defined by the 25\%- and 75\%-quantiles $q_1$ and $q_3$ as blue boxes (\BlockBlau). together with the median as black marker ($\bullet$) as well as the full blue line (\linefullblue). Blue circles (\CircBlau) denote outliers in the errors. Small horizontal shifts in the outliers are only for better readability. The blue vertical lines show the range of all data points except the outliers. Outliers are defined as all samples that are larger than $q_3 + 1.5 IQR$ or lower than $q_1 - 1.5 IQR$. Furthermore, the median relative empirical error of the least-squares solution is given (\linedashedgreen). The analysis is based on 50 different experimental designs.}
	\label{fig:SBL_res_LENO_box_MSE}
\end{figure}
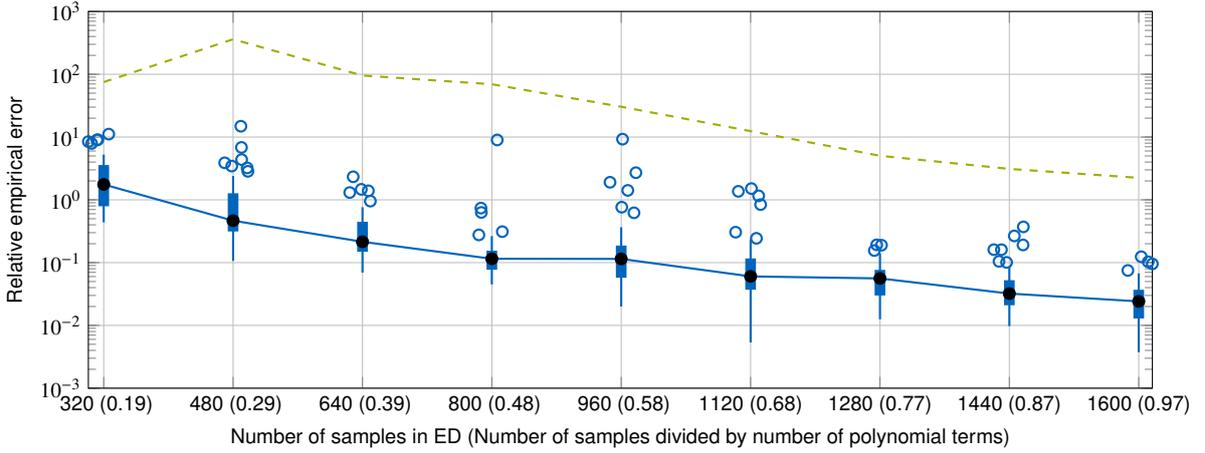

In Fig.~\ref{fig:SBL_res_LENO_box_MSE} we depict the relative empirical error in terms of the number of samples in the experimental design for case 1. 
The analysis is repeated 50 times and the results are summarized in form of a box plot.
Furthermore, the median relative empirical error of the least-squares solution, as presented in Section~\ref{sec:lsq}, is plotted.
We observe that the relative empirical error of the SBRA solution is significantly lower than the error based on the least squares solution.
In comparison to the previous example the errors in the approximation are higher and range between $10^0$ and $10^{-2}$ in median.
For increasing $N$ the median error decreases.

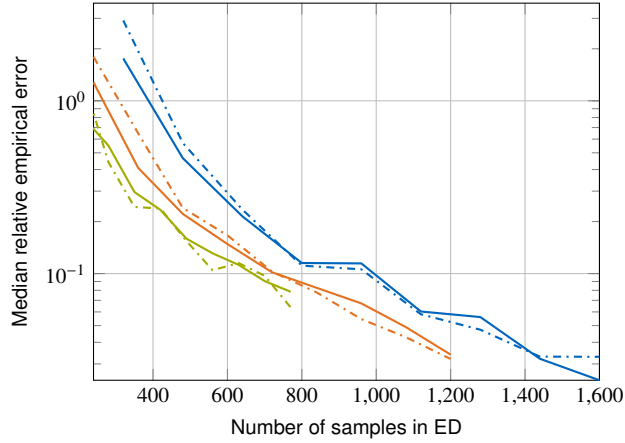
\begin{figure}[!hbt]
	\centering
	\setlength\fheight{5cm}
	\setlength\fwidth{7cm}
	\centering
	\input{LENO81EKL_comp_emp_error_median_lsq_randn.tikz}
	\caption{
		Comparison of the median relative empirical errors for the three basis sets and different initial points. 
		Blue lines (\linefullblue) represent case 1 ($m_p = m_q = 10$), orange lines (\linefullorange) represent case 2 ($m_p = m_q = 5$), and green lines (\linefullgreen) represent case 3 ($m_p = m_q = 3$). 
		Solid lines (---) represent the solutions based on the initial points stemming from the least squares solution, dash-dotted lines (--$\cdot$--) are based on random initial points sampled from a proper standard complex normal distribution. 
		It should be noted that the three experimental designs for the three cases are independent of each other, while the same experimental design is used for the two different initial points.
	}
	\label{fig:SBL_res_LENO_box_MSE_basis_sets_lsq_rand_comp}
\end{figure}

In Fig.~\ref{fig:SBL_res_LENO_box_MSE_basis_sets_lsq_rand_comp}, we compare the median relative empirical errors for the three basis sets. 
Furthermore, we evaluate the errors using random initial points for the RA coefficients.
We observe that the errors based on the least-squares and the random initial point are very similar. 
% Furthermore the lowest errors are achieved using the largest basis set, i.e., case 1, however, we do not have information on the errors for the smaller basis sets for larger sample sizes.
Overall, a strong decrease of the relative empirical error with increasing sample size can be observed for all three cases.

\begin{figure}[!hbt]
	\centering
	\setlength\fheight{4cm}
	\setlength\fwidth{7cm}
	\centering
	\input{LENO_DOS_mean.tikz}
	\caption{Mean value of the degree of sparsity for case 1 ($m_p = m_q = 10$); total model (\linefullblue), the numerator polynomial (\linedashedred) and the denominator polynomial (\linedashdotteddarkgreen). The analysis is based on 50 different experimental designs.}
	\label{fig:SBL_res_LENO_mean_DOS}
\end{figure}
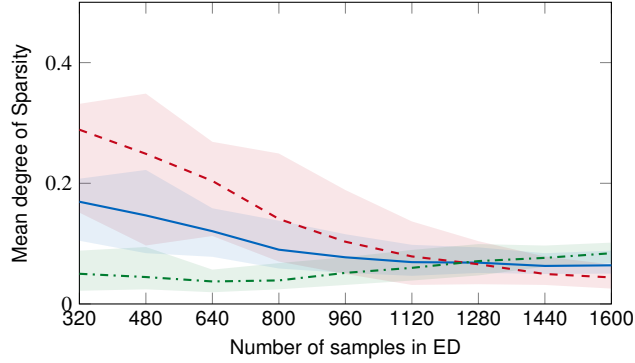

In Fig.~\ref{fig:SBL_res_LENO_mean_DOS} we depict the degree of sparsity in terms of the number of samples in the experimental design.
The median total degree is decreasing from around $0.17$ at $N=320$ to $0.06$ at $N=1600$.
For low sample sizes, the degree of sparsity is higher in the denominator than in the numerator, i.e., more terms are retained in the denominator basis set.
For large sample sizes, the degree of sparsity is slightly higher in the numerator polynomial.
Overall, Fig.~\ref{fig:SBL_res_LENO_mean_DOS} shows that the method is able to identify highly sparse rational representations of the investigated model.

\begin{figure}[!hbt]
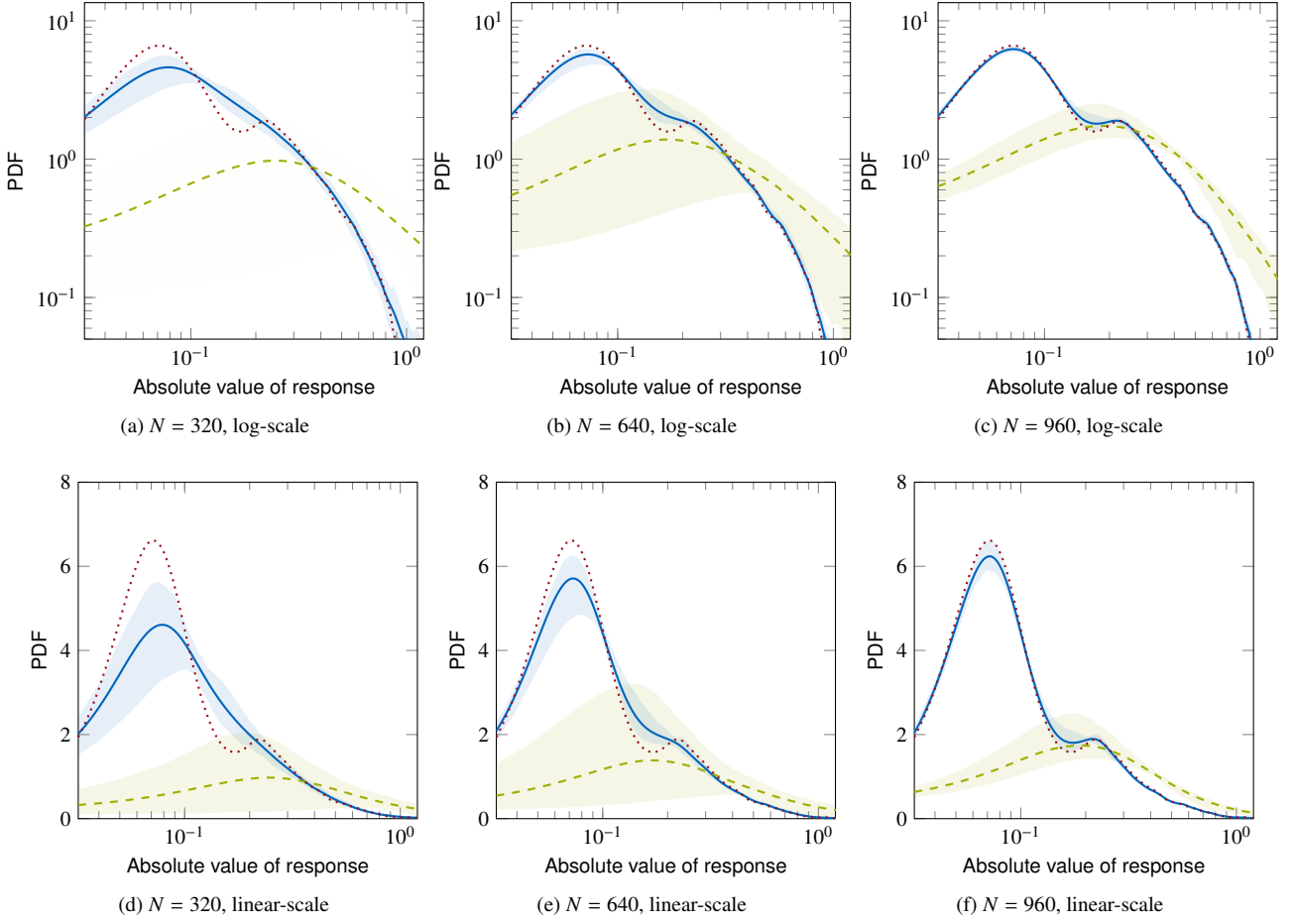

	\setlength\fheight{4.5cm}
	\setlength\fwidth{4.5cm}
	\centering
	\makebox[\textwidth][c]{
		\subfloat[$N = 320$, log-scale]{\input{LENO81EKL_pdf_y_abs_N_320.tikz}} \hfill
		\subfloat[$N = 640$, log-scale]{\input{LENO81EKL_pdf_y_abs_N_640.tikz}} \hfill
		\subfloat[$N = 960$, log-scale]{\input{LENO81EKL_pdf_y_abs_N_960.tikz}}
	} \\
	\makebox[\textwidth][c]{
		\subfloat[$N = 320$, linear-scale]{\input{LENO81EKL_pdf_y_abs_N_320_lin.tikz}} \hfill
		\subfloat[$N = 640$, linear-scale]{\input{LENO81EKL_pdf_y_abs_N_640_lin.tikz}} \hfill
		\subfloat[$N = 960$, linear-scale]{\input{LENO81EKL_pdf_y_abs_N_960_lin.tikz}} 
	}
	\caption{
		Probability density function for absolute value of response $\abs{\Tilde{h}}$ for the orthotropic plate model for case 1 ($m_p = m_q = 10)$). 
		Sparse approximation (\linefullblue) and corresponding 90 \% credible interval (\BlockBlau), least-squares approximation (\linedottedgreen) and corresponding 90 \% credible interval (\BlockHellGreen), Monte Carlo reference solution (\linedashedred).
		%    	The analysis is based on 50 different experimental designs.
	}
	\label{fig:SBL_res_LENO_PDFs}
\end{figure}

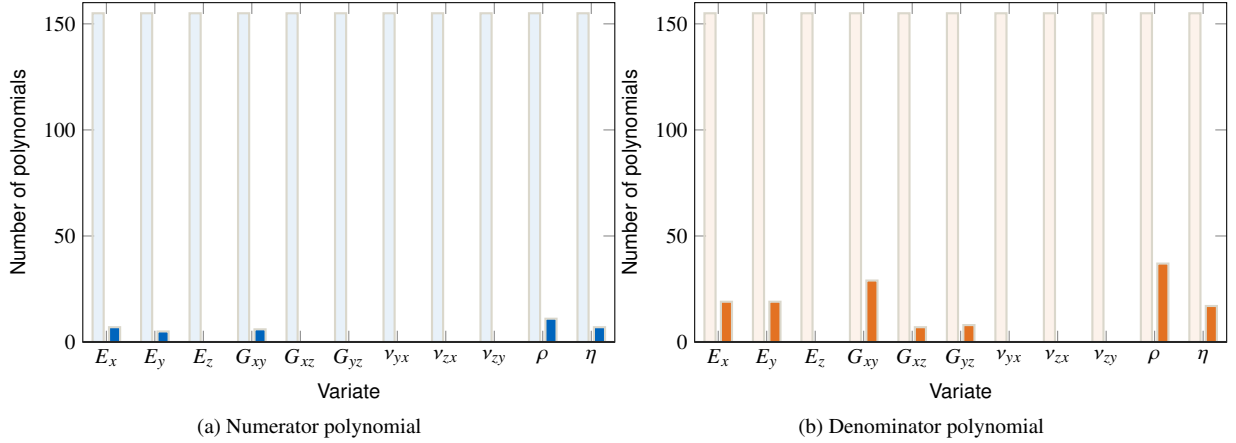
\begin{figure}[!hbt]
	\setlength\fheight{4.5cm}
	\setlength\fwidth{7cm}
	\centering
	\makebox[\textwidth][c]{
		\subfloat[Numerator polynomial]{\input{LENO81EKL_inv_variate_p.tikz}}
		\subfloat[Denominator polynomial]{\input{LENO81EKL_inv_variate_q.tikz}} 
	}
	\caption{
		Involved variates in the relevant basis sets. 
		The left, lightly coloured bar gives the corresponding number of terms in the initial full basis set, while the right, fully coloured bar represents the number of basis terms including the specified variate as identified by the SBRA.
	}
	\label{fig:SBL_res_LENO_inv_variate}
\end{figure}

Furthermore, we evaluate the marginal PDFs of the absolute value of the FRF for all experimental designs based on the test set and depict the results for the sample sizes $N=\{320, 640, 960\}$ in Figs.~\ref{fig:SBL_res_LENO_PDFs}.
The depicted PDFs are averages based on the 50 repeated LHS experimental designs.
For each experimental design, the PDFs are based on $N_t = 10^4$ test set samples and obtained by kernel density estimation. 
The reference solution is obtained by evaluating the model in Eq.~\eqref{eq:LENO_transf_func} for the test set samples. 
In addition to the average PDFs, the 90\%-credible intervals are depicted. 
% We observe that the approximated reflect the resulting error measures discussed before.
For the present model, the least-squares solution is not able to capture the PDF of the model response for all the shown sample sizes. 
In contrast, the PDF obtained by the SBRA compares reasonably well with the reference solution for $N=640$ and $N=960$. 
In particular, the bi-modality is accurately captured in the approximation.
For the lower sample size $N=320$, the SBRA does not capture the bi-modality, but improves the approximation accuracy in comparison to the least squares solution.
Furthermore, the credible intervals are narrow for the SBRA results, indicating that the SBRA robustly captures the response PDF independent of the specific experimental design.
%
% \begin{figure}[!hbt]
	% 	\setlength\fheight{4.5cm}
	% 	\setlength\fwidth{7cm}
	% 	\centering
	% 	\makebox[\textwidth][c]{
		% 		\subfloat[Numerator]{\input{figures/LENO81EKL/LENO81EKL_tot_degrees_p.tikz}}
		% 		\subfloat[Denominator]{\input{figures/LENO81EKL/LENO81EKL_tot_degrees_q.tikz}} 
		% 		}
	% 	\caption{Total degrees in the relevant basis sets.}
	%     \label{fig:SBL_res_LENO_PDFs}
	% 	%
	% \end{figure}
%
In Fig.~\ref{fig:SBL_res_LENO_inv_variate}, we depict the initial number of polynomial terms and the number of retained basis terms that include each variable for $N = 1600$ and the repetition that leads to the lowest empirical error.
One can observe that in both polynomials, the retained terms in the basis set can be associated with only few of the input random variables, that is $E_x$, $E_y$, $G_{xy}$, $G_{xz}$, $G_{yz}$, $\rho$ and $\eta$. 
Those are exactly the terms that are relevant to the model response from a mechanical point of view. 
It can be assumed that the influence of the Poisson's ratios on the model response is minor, which is reflected in the fact that all basis functions containing the Poisson's ratios are pruned from the basis set.
Furthermore, the influence of the Young's modulus in thickness direction $E_z$ can also be neglected, since the mode shapes associated with oscillating behavior through the thickness will only occur at relatively high frequencies.
Finally, for the chosen frequency, the out-of plane shear deformations are rather small, thus rendering the model response insensitive to both shear moduli, $G_{xz}$ and $G_{yz}$, reflected in a lower number of polynomials including both quantities. 
Overall, this further illustrates that the SBRA is able to identify the relevant terms even in an over-parameterized model.
\section{Conclusion}
\label{sec:concl}
This work presents a novel approach to obtain a sparse rational approximation for complex-valued models that can be used for uncertainty quantification with models that exhibit a rational input-output relationship, such as frequency domain models in structural dynamics.
The rational approximation is defined through the ratio of two polynomials with complex-valued coefficients.
The proposed method is able to to identify a sparse rational approximation through determining the coefficients with highest predictive significance using a set of evaluations of the model at a number of collocation points. 
Hence, the method is suitable for application with black box models. 
The coefficients in the surrogate model are treated as random variables and the regression problem is cast in a Bayesian setting.
We make use of the fact that the posterior distribution of the numerator coefficients conditional on the denominator coefficients can be expressed analytically. 
The posterior distribution of the denominator coefficients is approximated through a Dirac at the maximum a-posteriori estimate and the hyperparameters are finally found through maximizing the data evidence.
This transfers the problem to a two-stage optimization task in which a quasi-Newton method is applied to find the MAP estimate of the denominator coefficients and the update rules for the hyperparameters are given analytically. 
In the MAP estimation procedure, we derive the conjugate cogradient of the objective function in terms of the complex denominator coefficients analytically through application of the $\mathbb{CR}$- or Wirtinger calculus.
This two-stage procedure is carried out iteratively and basis terms are pruned from the initial set based on their precisions.

We apply the method to two models: one algebraic model that represents the transfer function model of a single-degree of freedom frame structure with seven input random variables and a finite element model of an orthotropic plate with eleven input random variables.
The resulting sparse approximation is compared to a previously introduced least-squares approach.
We specifically investigate the method for sample sizes that are approximately less or equal to the number of polynomial terms in the expansions.
For those cases, the previously introduced least-squares approach is prone to overfitting and shows poor approximation accuracy for a small sample size.
It can be shown that the proposed method improves the quality-of-fit significantly. 
This especially holds for sample sizes from around $40$ to $50 \, \%$ of the number of polynomial terms in the expansions.
Furthermore, one can observe that the method successfully identifies basis terms that involve the relevant input parameters in the case of the finite element model.
Finally, the approximation accuracy is also investigated in terms of the response PDFs for both models.
We find that the proposed method is able to capture the response PDFs accurately.

Further research could aim at extending the method to efficiently treat vector-valued model output. 
This is especially relevant for models with a high spatial or frequency resolution.
Furthermore the developed method could be applied to inverse uncertainty quantification.
Therein, adaptive strategies that improve the quality-of-fit in the posterior density regions are of great interest.
\section*{Acknowledgement}
The authors would like to thank Daniel Straub for the helpful comments on the manuscript.
The first author would like to thank Quirin Aumann for the valuable discussions on efficiently computing the gradients in the MAP-estimation problem.
\appendix
\section{Relative Empirical Error}
\label{app:rel_emp_error}
In order to assess the accuracy of the rational approximation, we investigate the relative empirical error. 
The relative empirical error is a scaled version of the empirical error, which itself is a sample approximation to the generalization error that is defined by
\begin{equation}
	err = \E \left[ \abs{\mathcal{R} (\mathbf{X}) - \mathcal{M} (\mathbf{X})}^2 \right] = \int_{\mathbb{R}^d} \abs{\mathcal{R} (\mathbf{X}) - \mathcal{M} (\mathbf{X})}^2 f_{\mathbf{X}} (\mathbf{x}) \dd \mathbf{x} \, .
\end{equation}
The empirical error is then found through evaluating
\begin{equation}
	err_{\mathrm{emp}} = \frac{1}{N_v} \sum_{i=1}^{N_v} \abs{\mathcal{R} (\mathbf{x}_i) - \mathcal{M} (\mathbf{x}_i)}^2 \, ,
\end{equation}
where $N_v$ is the number of samples in the validation set and $\left\{ \mathbf{x}_i | i = 1, \ldots, N_v \right\}$ are the validation set samples.
The relative empirical error can be defined using the sample variance.
\begin{equation}
	\varepsilon_{\mathrm{emp}} = \frac{err_{\mathrm{emp}}}{\widehat{\Var} \left[ \mathcal{M} (\mathbf{X}) \right]} \, .
\end{equation}

\section{Distributions}
\subsection{The Complex Normal Distribution}
\label{app:compl_normal}
The following definitions and derivations are based on \cite{Schreier.2010}.
The probability density function (PDF) for a complex normally distributed random variable $Z$ can be defined as the joint Normal distribution of the real and imaginary parts, $\mathbf{X} = \Re \mathbf{Z}$ and $\mathbf{Y} = \Im \mathbf{Z}$, respectively.
We define the real composite random vector $\mathbf{R} = \left[ \mathbf{X}; \mathbf{Y} \right]$ and write the joint distribution of $\mathbf{X}$ and $\mathbf{Y}$ as
\begin{equation}
	f_\mathbf{R} (\mathbf{r}) = \frac{1}{(2 \pi)^{\frac{2 n}{2}} \sqrt{\det \bm{\Sigma}_{RR}}} \exp{- \frac{1}{2} ( \mathbf{r} - \bm{\mu}_R )^T \bm{\Sigma}_{RR}^{-1} ( \mathbf{r} - \bm{\mu}_R )} \,
\end{equation}
with the joint mean vector $\bm{\mu}_R = \left[ \bm{\mu}_X ; \bm{\mu}_Y \right]$ and joint covariance matrix
\begin{equation}
	\bm{\Sigma}_{RR} = \begin{bmatrix} \bm{\Sigma}_{XX} & \bm{\Sigma}_{XY} \\ \bm{\Sigma}_{YX} & \bm{\Sigma}_{YY} \end{bmatrix} \, .
\end{equation}
Often, it is more convenient to work with a description that does not require splitting the complex quantities into real and imaginary parts.
% Using the transformation rule in Eq.~\eqref{eq:real_to_complex}, 
The complex normal PDF in terms of the complex augmented random vector $\underline{\mathbf{Z}} = \left[ \mathbf{Z};  \conj{\mathbf{Z}} \right]$ is given by
\begin{equation}
	f_{\compaug{\mathbf{Z}}} (\compaug{\mathbf{z}}) = \frac{1}{\pi^{n} \sqrt{\det \compaug{\bm{\Sigma}}_{ZZ}}} \exp{- \frac{1}{2} ( \compaug{\mathbf{z}} - \compaug{\bm{\mu}}_Z )^H \compaug{\bm{\Sigma}}_{ZZ}^{-1} ( \compaug{\mathbf{z}} - \compaug{\bm{\mu}}_Z )}
\end{equation}
In here, $\compaug{\bm{\mu}}_Z$ denotes the complex augmented mean vector, with $\compaug{\bm{\mu}}_Z = \left[ \bm{\mu}_Z ; \conj{\bm{\mu}}_Z \right] = \left[ \bm{\mu}_X + \mathrm{i} \bm{\mu}_Y ; \bm{\mu}_X - \mathrm{i} \bm{\mu}_Y \right]$ and 
\begin{equation}
	\compaug{\bm{\Sigma}}_{ZZ} = \begin{bmatrix} \bm{\Sigma}_{ZZ} & \widetilde{\bm{\Sigma}}_{ZZ} \\ \conj{\widetilde{\bm{\Sigma}}}_{ZZ} & \conj{\bm{\Sigma}}_{ZZ} \end{bmatrix}
\end{equation}
denotes the complex augmented covariance matrix, where $\bm{\Sigma}_{ZZ}$ is the (Hermitian) covariance matrix and $\widetilde{\bm{\Sigma}}_{ZZ}$ is the \textit{complementary covariance matrix}, defined by
\begin{gather}
	\bm{\Sigma}_{ZZ} = \E \left[ \left( \mathbf{z} - \bm{\mu}_Z \right) \left( \mathbf{z} - \bm{\mu}_Z \right)^H \right]  \\ %= \bm{\Sigma}_{XX} + \bm{\Sigma}_{YY} + \mathrm{i} \left( \bm{\Sigma}_{YX} - \bm{\Sigma}_{XY} \right) \, , \\
	\widetilde{\bm{\Sigma}}_{ZZ} = \E \left[ \left( \mathbf{z} - \bm{\mu}_Z \right) \left( \mathbf{z} - \bm{\mu}_Z \right)^T \right] %= \bm{\Sigma}_{XX} - \bm{\Sigma}_{YY} + \mathrm{i} \left( \bm{\Sigma}_{YX} + \bm{\Sigma}_{XY} \right) \, . 
	\label{eq:comp_cov_matrix_def}
\end{gather}
For shorthand definition, we write that $\mathbf{Z} \sim \mathcal{CN} \left( \mathbf{z}; \bm{\mu}_Z, \bm{\Sigma}_{ZZ}, \widetilde{\bm{\Sigma}}_{ZZ}\right)$, which is equivalent to $\mathbf{R} \sim \mathcal{N} \left( \bm{\mu}_R, \bm{\Sigma}_{RR} \right)$. 

For \textit{proper} random vectors it holds that $\widetilde{\bm{\Sigma}}_{ZZ} = \mathbf{0}$. 
In this case the complex augmented covariance matrix becomes block-diagonal, and we can write the complex normal PDF in terms of the complex vector $\mathbf{Z}$ as
\begin{equation}
	f_\mathbf{Z} (\mathbf{z}) = \frac{1}{\pi^{n} \det \bm{\Sigma}_{ZZ}} \exp{- ( \mathbf{z} - \bm{\mu}_Z )^H \bm{\Sigma}_{ZZ}^{-1} ( \mathbf{z} - \bm{\mu}_Z )}
\end{equation}
In the proper case, the covariance matrix $\bm{\Sigma}_{ZZ}$ fully defines the second order properties of the complex normal distribution.
In that case, we write $\mathbf{Z} \sim \mathcal{CN} \left( \mathbf{z}; \bm{\mu}_Z, \bm{\Sigma}_{ZZ}\right)$
% For the scalar case, we can write
% %
% \begin{equation}
	% 	f_Z (z) = \frac{1}{\pi^{n} \sigma_{Z}^2} \exp{- \frac{\abs{ z - \mu_Z }^2}{\sigma_{Z}^2} }
	% \end{equation}
% %
% where $\sigma_{Z} = \sqrt{\E \left[ \left( \conj{z} - \conj{\mu}_Z \right) \left( z - \mu_Z \right) \right]}$ denotes the standard deviation of $Z$.
%
\subsection{The Gamma Distribution}
\label{app:gamma}
The Gamma-distribution (in shape-rate-representation) is defined as
\begin{equation}
	\mathrm{Gam} (x | a,b) = \frac{b^a}{\Gamma (a)} x^{a-1} \mathrm{e}^{-bx}
\end{equation}
\section{Gradients with Respect to the Denominator Coefficients $\mathbf{q}$}
\label{app:grad_q}
In this section we compute the partial derivatives of the objective function in Eq.~\eqref{eq:q_MAP2} with respect to the denominator coefficients $\mathbf{q}$.
Since the coefficients $\mathbf{q}$ are complex-valued, we make use of the Wirtinger-derivatives, see, e.g., \cite{Schreier.2010,kreutz2009complex}. Due to the fact that the objective function is real-valued, we only need to consider the partial derivatives with respect to the conjugate of the denominator coefficients, $\conj{\mathbf{q}}$. 
The problem reads
\begin{equation}
	\pdv{\conj{\mathbf{q}}} \left[ \ln \det \bm{\Sigma} - \beta \mathbf{y}^H \left(  \mathbf{y} - \bm{\Psi} \bm{\mu} \right) - \mathbf{q}^H \bm{\Lambda}_{qq} \mathbf{q} \right]
	\label{eq:deriv_den_coeff}
\end{equation}
Instead of computing the vector-valued derivative directly, we compute the partial derivatives with respect to the $n_q$ individual scalar coefficients. For the first term, we obtain
\begin{equation}
	\pdv{\ln \det \bm{\Sigma}}{\conj{q}_{i}} = - \pdv{\ln \det \bm{\Sigma}^{-1}}{\conj{q}_{i}} = - \tr{\bm{\Sigma} \pdv{\bm{\Sigma}^{-1}}{\conj{q}_{i}}} \, ,
	\label{eq:F_conj_q_first_part}
\end{equation}
where $\tr(\cdot)$ denotes the trace-operator.
Furthermore, for the sake of notational convenience, we write $\mathbf{Q} = \diag(\bm{\Psi}_Q \mathbf{q})$. 
Inserting \eqref{eq:post_cov_p} and $\bm{\Psi} = \mathbf{Q}^{-1} \bm{\Psi}_p$, we obtain
\begin{equation}
	\pdv{\bm{\Sigma}^{-1}}{\conj{q}_{i}} = \beta \pdv{\bm{\Psi}^H \bm{\Psi}}{\conj{q}_{i}} = \beta \pdv{\bm{\Psi}^H}{\conj{q}_{i}} \bm{\Psi} = \beta \bm{\Psi}_p^T \pdv{\mathbf{Q}^{-H}}{\conj{q}_{i}} \bm{\Psi} \, .
	\label{eq:F_conj_q_2}
\end{equation}
%
% With $\bm{\Psi} = \mathbf{Q}^{-1} \bm{\Psi}_p$, we obtain
% %
% \begin{equation}
	% 	\pdv{\bm{\Sigma}^{-1}}{\conj{q}_{i}} = \beta \bm{\Psi}_p^T \pdv{\mathbf{Q}^{-H}}{\conj{q}_{i}} \bm{\Psi} \, ,
	% 	\label{eq:deriv_sigma_inv_conj_q}
	% \end{equation}
% %
% since
% %
% \begin{equation}
	% 	\pdv{\bm{\Psi}^H}{\conj{q}_{i}} = \bm{\Psi}_p^T \pdv{\mathbf{Q}^{-H}}{\conj{q}_{i}} \, ,
	% 	\label{eq:deriv_psi_hermit_conj_q}
	% \end{equation}
%
We define $\bm{\Xi}_i = \pdv{\mathbf{Q}^{-H}}{\conj{q}_i}$ and find
\begin{equation}
	% \conj{\mathbf{Q}^{-1}_{,i}} := 
	% \pdv{\mathbf{Q}^{-H}}{\conj{q}_i} = 
	\bm{\Xi}_i = \mathbf{Q}^{-H} \pdv{\conj{\mathbf{Q}}}{\conj{q}_i} \mathbf{Q}^{-H}
	= \diag{\left[ - \frac{\psi_i^q (\mathbf{x_1})}{\left(\bm{\psi}^q (\mathbf{x_1}) \conj{\mathbf{q}} \right)^{2}}, - \frac{\psi_i^q (\mathbf{x_2})}{\left(\bm{\psi}^q (\mathbf{x_2}) \conj{\mathbf{q}}\right)^{2}}, \ldots, - \frac{\psi_i^q (\mathbf{x_N})}{\left(\bm{\psi}^q (\mathbf{x_N}) \conj{\mathbf{q}} \right)^{2}} \right]} \, .
	% 	= \begin{bmatrix}
		% 		- \frac{\psi_i^q (\mathbf{x_1})}{\left(\bm{\psi}^q (\mathbf{x_1}) \conj{\mathbf{q}} \right)^{2}} & 0 & \hdots & 0 \\
		% 		%
		% 		0 & - \frac{\psi_i^q (\mathbf{x_2})}{\left(\bm{\psi}^q (\mathbf{x_2}) \conj{\mathbf{q}}\right)^{2}} & \hdots & 0 \\
		% 		%
		% 		\vdots & \vdots & \ddots & \vdots \\
		% 		%
		% 		0 & 0 & \hdots & - \frac{\psi_i^q (\mathbf{x_N})}{\left(\bm{\psi}^q (\mathbf{x_N}) \conj{\mathbf{q}} \right)^{2}} \\
		% 	\end{bmatrix}
	\label{eq:derivQInvHerm}
\end{equation}
Using the results in Eqs.~\eqref{eq:F_conj_q_2} and \eqref{eq:derivQInvHerm}, Eq.~\eqref{eq:F_conj_q_first_part} becomes
% Thus, (since or $- \tr{\beta \bm{\Sigma} \bm{\Psi}^H \conj{\mathbf{Q}}_{,i} \conj{\mathbf{Q}}^{-H} \bm{\Psi}} = - \tr{\beta \bm{\Sigma} \bm{\Psi}^H \mathbf{A}_i \bm{\Psi}}$)
%
\begin{equation}
	\pdv{\ln \det \bm{\Sigma}}{\conj{q}_{i}} = - \tr{\beta \bm{\Sigma} \bm{\Psi}_p^T \bm{\Xi}_i \bm{\Psi}} \, .
	\label{eq:first_result}
\end{equation}
In order to compute the trace efficiently for all $i = 1,\ldots,n_q$, we apply the cyclic property of the trace, i.e.,
% ($\tr{\mathbf{A}\mathbf{B}\mathbf{C}} = \tr{\mathbf{C}\mathbf{A}\mathbf{B}} = \tr{\mathbf{B}\mathbf{C}\mathbf{A}}$).
%
% Then,
%
\begin{equation}
	- \tr{\beta \bm{\Sigma} \bm{\Psi}_p^T \bm{\Xi}_i \bm{\Psi}} = - \tr{\beta \bm{\Psi} \bm{\Sigma} \bm{\Psi}_p^T \bm{\Xi}_i} \, .
\end{equation}
Since $\bm{\Xi}_i$ is a diagonal matrix, we can rewrite the above expression as the following sum
\begin{equation}
	- \tr{\beta \bm{\Psi} \bm{\Sigma} \bm{\Psi}_p^T \bm{\Xi}_i} = - \sum_{k=1}^{N} \left( \beta \bm{\Psi} \bm{\Sigma} \bm{\Psi}_p^T \right)_{kk} \left( \bm{\Xi}_i \right)_{kk} \, .
\end{equation}
The first matrix can be pre-computed and it's diagonal terms are extracted. 
Then, a vector-matrix product gives the result for all $i = 1,\ldots,n_q$.

For the second term, we obtain
\begin{equation}
	\pdv{\beta \mathbf{y}^H \left(  \mathbf{y} - \bm{\Psi} \bm{\mu} \right)}{\conj{q}_{i}} = - \pdv{\beta \mathbf{y}^H \bm{\Psi} \bm{\mu}}{\conj{q}_{i}} = - \beta \mathbf{y}^H \bm{\Psi} \pdv{\bm{\mu}}{\conj{q}_{i}}
\end{equation}
With
\begin{equation}
	\bm{\mu} = \beta \bm{\Sigma} \bm{\Psi}^H \mathbf{y} \, ,
\end{equation}
we find
\begin{equation}
	\pdv{\bm{\mu}}{\conj{q}_{i}} = \beta \pdv{\bm{\Sigma}}{\conj{q}_{i}} \bm{\Psi}^H \mathbf{y} + \beta \bm{\Sigma} \pdv{ \bm{\Psi}^H}{\conj{q}_{i}} \mathbf{y}
\end{equation}
Considering Eq.~\eqref{eq:F_conj_q_2}, it holds
\begin{equation}
	\pdv{\bm{\Sigma}}{\conj{q}_{i}} - \bm{\Sigma} \pdv{\bm{\Sigma}^{-1}}{\conj{q}_{i}} \bm{\Sigma}  = - \bm{\Sigma} \left( \beta \bm{\Psi}_p^T\conj{\mathbf{Q}^{-1}_{,i}} \bm{\Psi} \right) \bm{\Sigma}  \, .
\end{equation}
We define $\bm{\Gamma}_i = \beta \bm{\Sigma} \bm{\Psi}_p^T \bm{\Xi}_i$, and finally obtain
\begin{equation}
	\pdv{\beta \mathbf{y}^H \left(  \mathbf{y} - \bm{\Psi} \bm{\mu} \right)}{\conj{q}_{i}} = - \beta \mathbf{y}^H \bm{\Psi} \left( \bm{\Gamma}_i \left( \bm{\Psi} \bm{\mu} -  \mathbf{y} \right) \right) \, . \label{eq:scond_result}
\end{equation}
For the third term, we obtain
\begin{equation}
	\pdv{\mathbf{q}^H \bm{\Lambda}_{qq} \mathbf{q}}{\conj{q}_{i}} = \alpha_{q,i} q_i
	\label{eq:third_result}
\end{equation}
Thus, collecting the results in Eqs.~\eqref{eq:first_result}, \eqref{eq:scond_result} and \eqref{eq:third_result}, we find
\begin{equation}
	\pdv{\conj{q}_i} \left[ \ln \det \bm{\Sigma} - \beta \mathbf{y}^H \left(  \mathbf{y} - \bm{\Psi} \bm{\mu} \right) - \mathbf{q}^H \bm{\Lambda}_{qq} \mathbf{q} \right] = - \tr{\beta \bm{\Sigma} \bm{\Psi}_p^T \bm{\Xi}_i \bm{\Psi}} - \beta \mathbf{y}^H \bm{\Psi} \left( \bm{\Gamma}_i \left( \bm{\Psi} \bm{\mu} -  \mathbf{y} \right) \right) + \alpha_{q,i} q_i
\end{equation}
\section{Partial Derivatives with Respect to the Hyperparameters $\alpha_p$}
\label{app:grad_ap}
In this section we compute the partial derivatives of the objective function in Eq.~\eqref{eq:hyper_MAP2} with respect to $\bm{\alpha}_q$. The problem reads
\begin{equation}
	\pdv{\bm{\alpha}_p} \left[ \ln \det \bm{\Sigma} + \ln \det \bm{\Lambda_{pp}} - \beta \mathbf{y}^H \left(  \mathbf{y} - \bm{\Psi} \bm{\mu} \right) 
	% - \mathbf{q}^H \bm{\Lambda}_{qq} \mathbf{q} 
	\right]
	\label{eq:deriv_alpha_p}
\end{equation}
Instead of computing the vector-valued derivative directly, we compute the partial derivatives with respect to the individual scalar coefficients. For the first term, we obtain
\begin{equation}
	\pdv{\ln\det \bm{\Sigma}}{\alpha_{p,i}} = \pdv{\alpha_{p,i}} \ln \left( \det \bm{\Sigma}^{-1} \right)^{-1} = - \pdv{\alpha_{p,i}} \ln \det \bm{\Sigma}^{-1} = - \tr \bm{\Sigma} \pdv{\bm{\Sigma}^{-1}}{\alpha_{p,i}} = - \tr \bm{\Sigma} \mathbb{I}_{ii} = - \Sigma_{ii} \, .
	\label{eq:deriv_part_1}
\end{equation}
Here, we used:
\begin{equation}
	\pdv{\bm{\Sigma}^{-1}}{\alpha_{p,i}} = \pdv{\bm{\Lambda}_{pp} + \beta \bm{\Psi}^H \bm{\Psi}}{\alpha_{p,i}} = \mathbb{I}_{ii}
\end{equation}
where $\mathbb{I}_{jk}$ is the single-entry matrix with entry one at position $(j,k)$. 

Since $\bm{\Lambda}_{pp}$ is a diagonal matrix, we can write $\ln \det \bm{\Lambda}_{pp} = \sum_{i=1}^{n_p} \ln \alpha_{p,i}$ and the second part becomes
\begin{equation}
	\pdv{\ln\det \bm{\Lambda}_{pp}}{\alpha_{p,i}} = \tr \bm{\Lambda}_{pp}^{-1} \pdv{\bm{\Lambda}_{pp}}{\alpha_{p,i}} = \tr \bm{\Lambda}_{pp}^{-1} \mathbb{I}_{ii} = \alpha_{p,i}^{-1} \, .
	\label{eq:deriv_part_2}
\end{equation}
In order to derive the derivatives for the third part, we follow \cite{Tipping.2001} and rewrite
\begin{equation}
	\beta \mathbf{y}^H \left(  \mathbf{y} - \bm{\Psi} \bm{\mu} \right) = \beta \norm{\mathbf{y} - \bm{\Psi} \bm{\mu}}^2 + \bm{\mu}^H \bm{\Lambda_{pp}} \bm{\mu} \, ,
\end{equation}
where $\norm{\cdot}$ denotes the Euclidean vector norm.
Then, the derivative of the third part becomes
\begin{equation}
	\pdv{\beta \mathbf{y}^H \left( \mathbf{y} - \bm{\Psi} \bm{\mu} \right)}{\alpha_{p,i}} = \pdv{\bm{\mu}^H \bm{\Lambda_{pp}} \bm{\mu}}{\alpha_{p,i}} = \abs{\mu_i}^2
	\label{eq:deriv_part_3}
\end{equation}
%
% Finally, the derivative of the fourth part reads
%
Inserting Eqs.~\eqref{eq:deriv_part_1}, \eqref{eq:deriv_part_2} and \eqref{eq:deriv_part_3} into Eq.~ \eqref{eq:deriv_alpha_p}, and setting the total expression to zero, we obtain
\begin{equation}
	\Sigma_{ii} + \alpha_{p,i}^{-1} - \abs{\mu_i}^2 = 0 \, .
\end{equation}
Thus, the optimal value reads
\begin{equation}
	\alpha_{p,i} = \frac{1}{\Sigma_{ii} + \abs{\mu_i}^2}  \, .
\end{equation}
The resulting expression corresponds to the one presented in \cite{Tipping.2001} for real-valued, linear Gaussian models.
\section{Partial Derivatives with Respect to the Hyperparameters $\alpha_q$}
\label{app:grad_aq}
In this section we compute the partial derivatives of the objective function in Eq.~\eqref{eq:hyper_MAP2} with respect to $\bm{\alpha}_q$. The problem reads
\begin{equation}
	\pdv{\bm{\alpha}_q} \left[ \ln \det \bm{\Lambda}_{qq} - \mathbf{q}^H \bm{\Lambda}_{qq} \mathbf{q} \right] \, .
\end{equation}
Again, we compute the derivative with respect to each scalar hyperparameter individually. 
Since $\bm{\Lambda}_{qq}$ is a diagonal matrix, we can again write $\ln \det \bm{\Lambda}_{qq} = \sum_{i=1}^{n_q} \ln \alpha_{q,i}$.
Thus,
\begin{equation}
	\pdv{\bm{\alpha}_q} \left[ \ln \det \bm{\Lambda}_{qq} - \mathbf{q}^H \bm{\Lambda}_{qq} \mathbf{q} \right] = \alpha_{q,i}^{-1} - \abs{q_i}^2 \, .
\end{equation}
Setting the derivative to zero, we can find the optimal value analytically:
\begin{equation}
	\alpha_{q,i} = \frac{1}{\abs{q_i}^2} \, .
	\label{eq:deriv_alpha_q}
\end{equation}
\section{Partial Derivatives with Respect to the Hyperparameter $\beta$}
\label{app:grad_beta}
In this section we compute the partial derivatives of the objective function in Eq.~\eqref{eq:hyper_MAP2} with respect to $\beta$. The problem reads
\begin{equation}
	\pdv{\beta} \left[ \ln \det \bm{\Sigma}  + N \ln \beta - \beta \mathbf{y}^H \left(  \mathbf{y} - \bm{\Psi} \bm{\mu} \right) \right]
	\label{eq:deriv_beta}
\end{equation}
We treat the summands individually. 
For the first part, we find
\begin{equation}
	\pdv{\ln \det \bm{\Sigma}}{\beta} = - \tr \left( \bm{\Sigma} \pdv{\bm{\Sigma}^{-1}}{\beta} \right) = - \tr \left( \bm{\Sigma} \bm{\Psi}^H \bm{\Psi} \right) \, ,
\end{equation}
since
\begin{equation}
	\pdv{\bm{\Sigma}^{-1}}{\beta} = \pdv{\bm{\Lambda}_{pp} + \beta \bm{\Psi}^H \bm{\Psi}}{\beta} = \bm{\Psi}^H \bm{\Psi} \, .
\end{equation}
Furthermore, for the second part it holds
\begin{equation}
	\pdv{N \ln \beta}{\beta} = \frac{N}{\beta} \, .
\end{equation}
We rewrite the third part as follows
\begin{equation}
	\pdv{\left( - \beta \mathbf{y}^H \left(  \mathbf{y} - \bm{\Psi} \bm{\mu} \right) \right)}{\beta} = \pdv{\left( - \beta \mathbf{y}^H  \mathbf{y} +  \beta \mathbf{y}^H \bm{\Psi} \bm{\mu} \right)}{\beta} = - \mathbf{y}^H  \mathbf{y} + \pdv{\beta \mathbf{y}^H \bm{\Psi} \bm{\mu}}{\beta} \, 
	\label{eq:deriv_beta_last_part}
\end{equation}
and finally find the partial derivative for the last part in Eq.~\eqref{eq:deriv_beta_last_part} as
\begin{align}
	\pdv{\beta} \beta \mathbf{y}^H \bm{\Psi} \bm{\mu} & = \pdv{\beta}  \left[\beta \mathbf{y}^H \bm{\Psi} \beta \bm{\Sigma} \bm{\Psi}^H \bm{y} \right]\\
	& = \pdv{\beta}  \left[\beta^2 \mathbf{y}^H \bm{\Psi} \left( \bm{\Lambda}_{pp} + \beta \bm{\Psi}^H \bm{\Psi} \right)^{-1} \bm{\Psi}^H \bm{y} \right]\nonumber\\
	& = \underbrace{2 \beta \mathbf{y}^H \bm{\Psi} \bm{\Sigma} \bm{\Psi}^H \mathbf{y}}_{2 \mathbf{y}^H \bm{\Psi} \bm{\mu}} + \beta^2 \mathbf{y}^H \bm{\Psi} \pdv{\beta} \left( \bm{\Lambda}_{pp} + \beta \bm{\Psi}^H \bm{\Psi} \right)^{-1} \bm{\Psi}^H \mathbf{y} \nonumber
\end{align}
It holds
\begin{align}
	\pdv{\beta} \left( \bm{\Lambda}_{pp} + \beta \bm{\Psi}^H \bm{\Psi} \right)^{-1} & = - \left( \bm{\Lambda}_{pp} + \beta \bm{\Psi}^H \bm{\Psi} \right)^{-1} 	\pdv{\beta} \left( \bm{\Lambda}_{pp} + \beta \bm{\Psi}^H \bm{\Psi} \right) \left( \bm{\Lambda}_{pp} + \beta \bm{\Psi}^H \bm{\Psi} \right)^{-1} \nonumber \\
	& = - \left( \bm{\Lambda}_{pp} + \beta \bm{\Psi}^H \bm{\Psi} \right)^{-1} \left( \bm{\Psi}^H \bm{\Psi} \right) \left( \bm{\Lambda}_{pp} + \beta \bm{\Psi}^H \bm{\Psi} \right)^{-1} \nonumber \\
	& = - \bm{\Sigma} \left( \bm{\Psi}^H \bm{\Psi} \right) \bm{\Sigma} \, .
\end{align}
Then,
\begin{equation}
	\pdv{\beta} \beta \mathbf{y}^H \bm{\Psi} \bm{\mu} = 2 \mathbf{y}^H \bm{\Psi} \bm{\mu} - \beta^2 \mathbf{y}^H \bm{\Psi} \bm{\Sigma} \left( \bm{\Psi}^H \bm{\Psi} \right) \bm{\Sigma} \bm{\Psi}^H \mathbf{y} \, .
\end{equation}
In total, the partial derivative with respect to $\beta$ reads
\begin{equation}
	- \tr \left( \bm{\Sigma} \bm{\Psi}^H \bm{\Psi} \right) + \frac{N}{\beta} - \mathbf{y}^H  \mathbf{y} + 2 \mathbf{y}^H \bm{\Psi} \bm{\mu} - \beta^2 \mathbf{y}^H \bm{\Psi} \bm{\Sigma} \left( \bm{\Psi}^H \bm{\Psi} \right) \bm{\Sigma} \bm{\Psi}^H  \mathbf{y} + \frac{c}{\beta} - d
\end{equation}
We rewrite the resulting expression
\begin{equation}
	- \mathbf{y}^H  \mathbf{y} + 2 \mathbf{y}^H \bm{\Psi} \bm{\mu} - \underbrace{\beta^2 \mathbf{y}^H \bm{\Psi} \bm{\Sigma} \left( \bm{\Psi}^H \bm{\Psi} \right) \bm{\Sigma} \bm{\Psi}^H \mathbf{y}}_{\bm{\mu}^H \bm{\Psi}^H \bm{\Psi} \bm{\mu}} = - \left( \mathbf{y} - \bm{\Psi} \bm{\mu} \right)^H \left( \mathbf{y} - \bm{\Psi} \bm{\mu} \right) = - \norm{\mathbf{y} - \bm{\Psi} \bm{\mu}}^2
\end{equation}
and obtain the derivatives in Eq.~\eqref{eq:deriv_beta} as
\begin{equation}
	- \tr \left( \bm{\Sigma} \bm{\Psi}^H \bm{\Psi} \right) + \frac{N}{\beta} - \norm{\mathbf{y} - \bm{\Psi} \bm{\mu}}^2 %+ \frac{c}{\beta} - d
	\label{eq:deriv_beta_final}
\end{equation}
% %
% and
% %
% \begin{equation}
	% 	\pdv{h}{\log \beta} = - \beta \tr \left( \bm{\Sigma} \bm{\Psi}^H \bm{\Psi} \right) + N - \beta \norm{\mathbf{y} - \bm{\Psi} \bm{\mu}}^2 + c - d \beta
	% \end{equation}
%
Setting Eq.~\eqref{eq:deriv_beta_final} to zero, and solving for $\beta$, we obtain the update rule for $\beta$ as
\begin{equation}
	\beta = \frac{N + c}{\norm{\mathbf{y} - \bm{\Psi} \bm{\mu}}^2 + \tr \left( \bm{\Sigma} \bm{\Psi}^H \bm{\Psi} \right) + d}
\end{equation}
%
%% References

\clearpage
\bibliographystyle{elsarticle-num}
\bibliography{dokument_red}



	
	
	
	%\newpage
	%% \end{thebibliography}

\end{document}

%% file: frame_rel_emp_error_10_10.tikz
%auto-ignore
\definecolor{mycolor1}{rgb}{0.00000,0.39608,0.74118}%
\definecolor{mycolor2}{rgb}{0.63529,0.67843,0.00000}%
\begin{tikzpicture}

\begin{axis}[%
width=\fwidth,
height=\fheight,
scale only axis,
xmin=0.87,
xmax=10.13,
xtick={1,2,3,4,5,6,7,8,9,10},
xticklabels={{120 (0.19)},{180 (0.28)},{240 (0.38)},{300 (0.47)},{360 (0.57)},{420 (0.66)},{480 (0.76)},{540 (0.85)},{600 (0.95)},{660 (1.04)}},
xlabel style={font=\color{white!15!black}},
xlabel={Number of samples in ED (Number of samples divided by number of polynomial terms)},
ymode=log,
ymin=5e-7,%8.55836038745482e-08,
ymax=5e1,%27.1681399657779,
yminorticks=true,
ylabel style={font=\color{white!15!black}},
ylabel={Relative empirical error},
axis background/.style={fill=white},
xmajorgrids,
ymajorgrids%,
%yminorgrids
]
\addplot [color=mycolor1, forget plot]
  table[row sep=crcr]{%
1	0.000583946130731938\\
1	2.21084278789737\\
};
\addplot [color=mycolor1, forget plot]
  table[row sep=crcr]{%
2	0.000136331389437803\\
2	0.213423421531585\\
};
\addplot [color=mycolor1, forget plot]
  table[row sep=crcr]{%
3	3.96571130854734e-05\\
3	0.0459881749704564\\
};
\addplot [color=mycolor1, forget plot]
  table[row sep=crcr]{%
4	2.40695974908258e-05\\
4	0.00239078730630921\\
};
\addplot [color=mycolor1, forget plot]
  table[row sep=crcr]{%
5	1.18569624817094e-05\\
5	0.000896999612896769\\
};
\addplot [color=mycolor1, forget plot]
  table[row sep=crcr]{%
6	1.45902044912955e-05\\
6	0.00228055604873324\\
};
\addplot [color=mycolor1, forget plot]
  table[row sep=crcr]{%
7	1.03006366234452e-05\\
7	0.0014598149846159\\
};
\addplot [color=mycolor1, forget plot]
  table[row sep=crcr]{%
8	1.43312746657487e-05\\
8	0.0032955461811075\\
};
\addplot [color=mycolor1, forget plot]
  table[row sep=crcr]{%
9	1.64187050223268e-05\\
9	0.0051088162488755\\
};
\addplot [color=mycolor1, forget plot]
  table[row sep=crcr]{%
10	1.15755034210251e-05\\
10	0.000719764246156249\\
};
\addplot [color=mycolor1, line width=4.0pt, forget plot]
  table[row sep=crcr]{%
1	0.0441795067681267\\
1	0.970160710715791\\
};
\addplot [color=mycolor1, line width=4.0pt, forget plot]
  table[row sep=crcr]{%
2	0.00313556604309329\\
2	0.190471089279979\\
};
\addplot [color=mycolor1, line width=4.0pt, forget plot]
  table[row sep=crcr]{%
3	0.000296153256099209\\
3	0.0288333395437913\\
};
\addplot [color=mycolor1, line width=4.0pt, forget plot]
  table[row sep=crcr]{%
4	0.000347229633293464\\
4	0.00126619460488273\\
};
\addplot [color=mycolor1, line width=4.0pt, forget plot]
  table[row sep=crcr]{%
5	0.000207292925162507\\
5	0.000665159699059889\\
};
\addplot [color=mycolor1, line width=4.0pt, forget plot]
  table[row sep=crcr]{%
6	7.85057628985225e-05\\
6	0.00108217996803976\\
};
\addplot [color=mycolor1, line width=4.0pt, forget plot]
  table[row sep=crcr]{%
7	6.36054875482009e-05\\
7	0.000733001020793094\\
};
\addplot [color=mycolor1, line width=4.0pt, forget plot]
  table[row sep=crcr]{%
8	5.62546571706907e-05\\
8	0.00140751601271075\\
};
\addplot [color=mycolor1, line width=4.0pt, forget plot]
  table[row sep=crcr]{%
9	0.000249842248445243\\
9	0.00276937565988139\\
};
\addplot [color=mycolor1, line width=4.0pt, forget plot]
  table[row sep=crcr]{%
10	2.82982356173069e-05\\
10	0.000305826938949638\\
};
\addplot [color=black, draw=none, mark=*, mark options={solid, fill=black, mycolor1}, forget plot]
  table[row sep=crcr]{%
1	0.190719863915864\\
};
\addplot [color=black, draw=none, mark=*, mark options={solid, fill=black, mycolor1}, forget plot]
  table[row sep=crcr]{%
2	0.0254861159360678\\
};
\addplot [color=black, draw=none, mark=*, mark options={solid, fill=black, mycolor1}, forget plot]
  table[row sep=crcr]{%
3	0.000622394447679252\\
};
\addplot [color=black, draw=none, mark=*, mark options={solid, fill=black, mycolor1}, forget plot]
  table[row sep=crcr]{%
4	0.000663912242903521\\
};
\addplot [color=black, draw=none, mark=*, mark options={solid, fill=black, mycolor1}, forget plot]
  table[row sep=crcr]{%
5	0.000380521329458401\\
};
\addplot [color=black, draw=none, mark=*, mark options={solid, fill=black, mycolor1}, forget plot]
  table[row sep=crcr]{%
6	0.000285366517285173\\
};
\addplot [color=black, draw=none, mark=*, mark options={solid, fill=black, mycolor1}, forget plot]
  table[row sep=crcr]{%
7	0.000173904392170292\\
};
\addplot [color=black, draw=none, mark=*, mark options={solid, fill=black, mycolor1}, forget plot]
  table[row sep=crcr]{%
8	0.000227715189494391\\
};
\addplot [color=black, draw=none, mark=*, mark options={solid, fill=black, mycolor1}, forget plot]
  table[row sep=crcr]{%
9	0.000875365285212379\\
};
\addplot [color=black, draw=none, mark=*, mark options={solid, fill=black, mycolor1}, forget plot]
  table[row sep=crcr]{%
10	5.92073503813418e-05\\
};
\addplot [color=black, draw=none, mark=*, mark options={solid, black}, forget plot]
  table[row sep=crcr]{%
1	0.190719863915864\\
};
\addplot [color=black, draw=none, mark=*, mark options={solid, black}, forget plot]
  table[row sep=crcr]{%
2	0.0254861159360678\\
};
\addplot [color=black, draw=none, mark=*, mark options={solid, black}, forget plot]
  table[row sep=crcr]{%
3	0.000622394447679252\\
};
\addplot [color=black, draw=none, mark=*, mark options={solid, black}, forget plot]
  table[row sep=crcr]{%
4	0.000663912242903521\\
};
\addplot [color=black, draw=none, mark=*, mark options={solid, black}, forget plot]
  table[row sep=crcr]{%
5	0.000380521329458401\\
};
\addplot [color=black, draw=none, mark=*, mark options={solid, black}, forget plot]
  table[row sep=crcr]{%
6	0.000285366517285173\\
};
\addplot [color=black, draw=none, mark=*, mark options={solid, black}, forget plot]
  table[row sep=crcr]{%
7	0.000173904392170292\\
};
\addplot [color=black, draw=none, mark=*, mark options={solid, black}, forget plot]
  table[row sep=crcr]{%
8	0.000227715189494391\\
};
\addplot [color=black, draw=none, mark=*, mark options={solid, black}, forget plot]
  table[row sep=crcr]{%
9	0.000875365285212379\\
};
\addplot [color=black, draw=none, mark=*, mark options={solid, black}, forget plot]
  table[row sep=crcr]{%
10	5.92073503813418e-05\\
};
\addplot [color=black, draw=none, mark size=2.0pt, mark=o, mark options={solid, mycolor1}, forget plot]
  table[row sep=crcr]{%
1.07868092159829	2.86890491852453\\
1.1014479842689	3.66782344412663\\
};
\addplot [color=black, draw=none, mark size=2.0pt, mark=o, mark options={solid, mycolor1}, forget plot]
  table[row sep=crcr]{%
1.90674670407338	0.594192959048883\\
2.10334396403475	0.757283459700208\\
2.03308981155635	0.875666524578453\\
1.89938510124985	0.875826520427512\\
1.94462455471676	0.902302888884174\\
2.01172037980125	0.979257565848051\\
2.11437670885857	1.15174251160646\\
2.11622213379982	1.2569909271438\\
1.91440327041939	1.84144692584847\\
2.11764819544015	2.45347451468025\\
2.11429173706074	3.04522475171358\\
};
\addplot [color=black, draw=none, mark size=2.0pt, mark=o, mark options={solid, mycolor1}, forget plot]
  table[row sep=crcr]{%
2.99634391218071	0.0731770439252512\\
3.0750701172222	0.0960519258656217\\
2.9104715846568	0.106898973300901\\
2.98044032065657	0.110699973396187\\
3.10393388129727	0.155532145967958\\
3.07305183238989	0.156168920683855\\
3.11487310659823	0.241666230927173\\
3.03893517478915	0.298898580076232\\
2.88392791964355	0.319413140450841\\
3.08728232646719	1.63658389392532\\
};
\addplot [color=black, draw=none, mark size=2.0pt, mark=o, mark options={solid, mycolor1}, forget plot]
  table[row sep=crcr]{%
4.10849831193939	0.0187288965107173\\
4.04468378871444	0.0192626162093675\\
4.06443503264458	0.02008321794511\\
4.06078311703123	0.0496160031948751\\
3.97305675488354	0.0623458507998989\\
4.03886947254439	0.113677720202006\\
3.91779667195289	0.116613956654917\\
4.0515115220049	0.1185374599981\\
3.88295821159436	0.177938089724701\\
};
\addplot [color=black, draw=none, mark size=2.0pt, mark=o, mark options={solid, mycolor1}, forget plot]
  table[row sep=crcr]{%
4.94423074624022	0.00300075816802946\\
4.88654284765779	0.00616764643967608\\
4.89928294530896	0.0150891362553507\\
5.08086445708182	0.0693815065989069\\
5.04870715574395	0.0853144955016745\\
4.95427487001522	0.113605591616546\\
5.11255551220959	0.290425105802648\\
};
\addplot [color=black, draw=none, mark size=2.0pt, mark=o, mark options={solid, mycolor1}, forget plot]
  table[row sep=crcr]{%
5.88361152012573	0.0144657745382027\\
5.9846860899141	0.0186197134651214\\
5.97038961427325	0.0210265560049384\\
6.06637919703725	0.0231767373979874\\
6.07379997528427	0.026489374891659\\
5.92171815113859	0.0288412163780302\\
5.99744109894706	0.0288897953154904\\
5.98639655017772	0.0327176726941176\\
6.03657825252782	0.0501019928765262\\
};
\addplot [color=black, draw=none, mark size=2.0pt, mark=o, mark options={solid, mycolor1}, forget plot]
  table[row sep=crcr]{%
7.05234120771452	0.00183144774413757\\
7.06367167049559	0.00778975147224498\\
6.94400626924964	0.00841356782946643\\
7.04492566921342	0.0107792410184336\\
7.03877450099346	0.0147146728439641\\
6.91565293379866	0.0244949164908618\\
6.90474942038959	0.0981410921611207\\
};
\addplot [color=black, draw=none, mark size=2.0pt, mark=o, mark options={solid, mycolor1}, forget plot]
  table[row sep=crcr]{%
7.99959101299554	0.00414335443356862\\
8.11493598962902	0.00986714719944593\\
7.96009643166653	0.0208017700700109\\
};
\addplot [color=black, draw=none, mark size=2.0pt, mark=o, mark options={solid, mycolor1}, forget plot]
  table[row sep=crcr]{%
9.02131693774494	0.0076629679787556\\
8.93095298487278	0.00859198882396283\\
9.06281676482641	0.0268615232684844\\
};
\addplot [color=black, draw=none, mark size=2.0pt, mark=o, mark options={solid, mycolor1}, forget plot]
  table[row sep=crcr]{%
9.93877377886482	0.00107126649938825\\
10.0014892629163	0.00146327882325031\\
10.0497691806642	0.00173634437457703\\
10.097725813134	0.00173691732911067\\
10.1148228563014	0.00177801974663813\\
};
% Median line from other file
\addplot [color=mycolor1, forget plot]
  table[row sep=crcr]{%
1	0.190719863915864\\
2	0.0254861159360678\\
3	0.000622394447679252\\
4	0.000663912242903521\\
5	0.000380521329458401\\
6	0.000285366517285173\\
7	0.000173904392170292\\
8	0.000227715189494391\\
9	0.000875365285212379\\
10	5.92073503813418e-05\\
};
\addplot [color=mycolor2, dashed, forget plot]
  table[row sep=crcr]{%
1	1.46976951024547\\
2	1.65863486357646\\
3	1.74628638382242\\
4	1.67790803747152\\
5	0.811199445046182\\
6	1.79269543590732\\
7	1.65510005381659\\
8	0.793600405491329\\
9	0.0590404482315904\\
10	8.10116313160001e-07\\
};
\end{axis}
\end{tikzpicture}%

%% file: frame_rel_emp_error_5_5.tikz
%auto-ignore
\definecolor{mycolor1}{rgb}{0.00000,0.39608,0.74118}%
\definecolor{mycolor2}{rgb}{0.63529,0.67843,0.00000}%
\begin{tikzpicture}

\begin{axis}[%
width=\fwidth,
height=\fheight,
scale only axis,
xmin=0.885,
xmax=9.115,
xtick={1,2,3,4,5,6,7,8,9,10},
xticklabels={{ 80 (0.20)},{120 (0.30)},{160 (0.41)},{200 (0.51)},{240 (0.61)},{280 (0.71)},{320 (0.81)},{360 (0.91)},{400 (1.02)},{440 (1.12)}},
xlabel style={font=\color{white!15!black}},
xlabel={Number of samples in ED (Number of samples divided by number of polynomial terms)},
ymode=log,
ymin=5e-7,
ymax=5e1,%4122.12113284103,
yminorticks=true,
ylabel style={font=\color{white!15!black}},
ylabel={Relative empirical error},
axis background/.style={fill=white},
xmajorgrids,
ymajorgrids%,
%yminorgrids
]
\addplot [color=mycolor1, forget plot]
  table[row sep=crcr]{%
1	0.0087968579270561\\
1	4.06446990813868\\
};
\addplot [color=mycolor1, forget plot]
  table[row sep=crcr]{%
2	0.000809877255342745\\
2	1.99155187036857\\
};
\addplot [color=mycolor1, forget plot]
  table[row sep=crcr]{%
3	0.000158943955662541\\
3	0.178415896497315\\
};
\addplot [color=mycolor1, forget plot]
  table[row sep=crcr]{%
4	4.19038829545879e-05\\
4	0.00190085204165764\\
};
\addplot [color=mycolor1, forget plot]
  table[row sep=crcr]{%
5	3.46586615485648e-05\\
5	0.00216198573447614\\
};
\addplot [color=mycolor1, forget plot]
  table[row sep=crcr]{%
6	1.38198526478137e-05\\
6	0.00961266674737716\\
};
\addplot [color=mycolor1, forget plot]
  table[row sep=crcr]{%
7	2.57405422822769e-05\\
7	0.0155505811116465\\
};
\addplot [color=mycolor1, forget plot]
  table[row sep=crcr]{%
8	1.80135704274608e-05\\
8	0.00768416068891468\\
};
\addplot [color=mycolor1, forget plot]
  table[row sep=crcr]{%
9	9.67294803495149e-06\\
9	0.000199873188502366\\
};
\addplot [color=mycolor1, forget plot]
  table[row sep=crcr]{%
10	1.11400730577696e-05\\
10	0.000132779954520137\\
};
\addplot [color=mycolor1, line width=4.0pt, forget plot]
  table[row sep=crcr]{%
1	0.0969943413241678\\
1	2.41267214594958\\
};
\addplot [color=mycolor1, line width=4.0pt, forget plot]
  table[row sep=crcr]{%
2	0.0410020594800459\\
2	0.826301801895641\\
};
\addplot [color=mycolor1, line width=4.0pt, forget plot]
  table[row sep=crcr]{%
3	0.000988216920295036\\
3	0.0910748902456524\\
};
\addplot [color=mycolor1, line width=4.0pt, forget plot]
  table[row sep=crcr]{%
4	0.000355321051728882\\
4	0.00165955001785774\\
};
\addplot [color=mycolor1, line width=4.0pt, forget plot]
  table[row sep=crcr]{%
5	0.000193652577167676\\
5	0.00107568240189682\\
};
\addplot [color=mycolor1, line width=4.0pt, forget plot]
  table[row sep=crcr]{%
6	0.000136120197038654\\
6	0.00469194564132005\\
};
\addplot [color=mycolor1, line width=4.0pt, forget plot]
  table[row sep=crcr]{%
7	0.00019972048082634\\
7	0.00710488243125348\\
};
\addplot [color=mycolor1, line width=4.0pt, forget plot]
  table[row sep=crcr]{%
8	0.000173779940461693\\
8	0.0032656740687583\\
};
\addplot [color=mycolor1, line width=4.0pt, forget plot]
  table[row sep=crcr]{%
9	2.7898384183589e-05\\
9	0.000107880436102535\\
};
\addplot [color=mycolor1, line width=4.0pt, forget plot]
  table[row sep=crcr]{%
10	2.61838402255812e-05\\
10	8.88885175383166e-05\\
};
\addplot [color=black, draw=none, mark=*, mark options={solid, fill=black, mycolor1}, forget plot]
  table[row sep=crcr]{%
1	0.466238245907102\\
};
\addplot [color=black, draw=none, mark=*, mark options={solid, fill=black, mycolor1}, forget plot]
  table[row sep=crcr]{%
2	0.0749520109576602\\
};
\addplot [color=black, draw=none, mark=*, mark options={solid, fill=black, mycolor1}, forget plot]
  table[row sep=crcr]{%
3	0.0150380231357767\\
};
\addplot [color=black, draw=none, mark=*, mark options={solid, fill=black, mycolor1}, forget plot]
  table[row sep=crcr]{%
4	0.000862360763338974\\
};
\addplot [color=black, draw=none, mark=*, mark options={solid, fill=black, mycolor1}, forget plot]
  table[row sep=crcr]{%
5	0.000359773717176647\\
};
\addplot [color=black, draw=none, mark=*, mark options={solid, fill=black, mycolor1}, forget plot]
  table[row sep=crcr]{%
6	0.000386561655774691\\
};
\addplot [color=black, draw=none, mark=*, mark options={solid, fill=black, mycolor1}, forget plot]
  table[row sep=crcr]{%
7	0.000487409414138383\\
};
\addplot [color=black, draw=none, mark=*, mark options={solid, fill=black, mycolor1}, forget plot]
  table[row sep=crcr]{%
8	0.000704633319556755\\
};
\addplot [color=black, draw=none, mark=*, mark options={solid, fill=black, mycolor1}, forget plot]
  table[row sep=crcr]{%
9	4.39064256324928e-05\\
};
\addplot [color=black, draw=none, mark=*, mark options={solid, fill=black, mycolor1}, forget plot]
  table[row sep=crcr]{%
10	4.53231797687043e-05\\
};
\addplot [color=black, draw=none, mark=*, mark options={solid, black}, forget plot]
  table[row sep=crcr]{%
1	0.466238245907102\\
};
\addplot [color=black, draw=none, mark=*, mark options={solid, black}, forget plot]
  table[row sep=crcr]{%
2	0.0749520109576602\\
};
\addplot [color=black, draw=none, mark=*, mark options={solid, black}, forget plot]
  table[row sep=crcr]{%
3	0.0150380231357767\\
};
\addplot [color=black, draw=none, mark=*, mark options={solid, black}, forget plot]
  table[row sep=crcr]{%
4	0.000862360763338974\\
};
\addplot [color=black, draw=none, mark=*, mark options={solid, black}, forget plot]
  table[row sep=crcr]{%
5	0.000359773717176647\\
};
\addplot [color=black, draw=none, mark=*, mark options={solid, black}, forget plot]
  table[row sep=crcr]{%
6	0.000386561655774691\\
};
\addplot [color=black, draw=none, mark=*, mark options={solid, black}, forget plot]
  table[row sep=crcr]{%
7	0.000487409414138383\\
};
\addplot [color=black, draw=none, mark=*, mark options={solid, black}, forget plot]
  table[row sep=crcr]{%
8	0.000704633319556755\\
};
\addplot [color=black, draw=none, mark=*, mark options={solid, black}, forget plot]
  table[row sep=crcr]{%
9	4.39064256324928e-05\\
};
\addplot [color=black, draw=none, mark=*, mark options={solid, black}, forget plot]
  table[row sep=crcr]{%
10	4.53231797687043e-05\\
};
\addplot [color=black, draw=none, mark size=2.0pt, mark=o, mark options={solid, mycolor1}, forget plot]
  table[row sep=crcr]{%
1.00778163170383	16.0518547826379\\
};
\addplot [color=black, draw=none, mark size=2.0pt, mark=o, mark options={solid, mycolor1}, forget plot]
  table[row sep=crcr]{%
1.87536579379311	2.22813421162849\\
2.09594720235057	2.86133506719605\\
1.97609596432694	3.18896618607256\\
};
\addplot [color=black, draw=none, mark size=2.0pt, mark=o, mark options={solid, mycolor1}, forget plot]
  table[row sep=crcr]{%
2.95030139662837	0.291812818955499\\
3.11264539785211	0.346802735323752\\
2.99016204830155	0.407559129557699\\
2.94691187776043	0.877662840665428\\
2.8961566596441	1.1663997212038\\
3.02054533379213	2.10690538727887\\
};
\addplot [color=black, draw=none, mark size=2.0pt, mark=o, mark options={solid, mycolor1}, forget plot]
  table[row sep=crcr]{%
3.91326717436077	0.00645558690754511\\
3.89327357654237	0.00862666520553147\\
4.02014119420771	0.0142749437842784\\
3.94675368809928	0.0155250025025648\\
3.96548002110029	0.0214556859742186\\
4.05620617345504	0.0255165538545068\\
4.08957791953146	0.0285029538632562\\
3.96197892166676	0.0313731476561521\\
4.11543640133147	0.0710384256216312\\
4.11339226289168	0.454281637398032\\
3.92650934420759	0.974301453035047\\
};
\addplot [color=black, draw=none, mark size=2.0pt, mark=o, mark options={solid, mycolor1}, forget plot]
  table[row sep=crcr]{%
5.06706121476769	0.00327164131496029\\
5.02888296996013	0.00661020154658381\\
5.10473062186308	0.00696296160915927\\
5.02563431530237	0.00725136724810694\\
5.05053455707615	0.0102225378958672\\
5.06091873880102	0.0216483287400229\\
4.97127556919602	0.0220745419460066\\
4.93787594729374	0.0222585293459142\\
4.88418993435646	0.150148477308322\\
4.99303071856184	0.511815055803232\\
};
\addplot [color=black, draw=none, mark size=2.0pt, mark=o, mark options={solid, mycolor1}, forget plot]
  table[row sep=crcr]{%
6.03626808210885	0.0168918121411313\\
5.94474248510051	0.0201349690570494\\
6.00446522746806	0.0283304800139673\\
5.93641665964996	0.0298416948003903\\
5.94937704360614	0.0395812768518445\\
6.03761852331047	0.0822839019557746\\
6.09784832638023	0.105923435501604\\
6.09027661667734	0.338078574390946\\
5.92747876919057	0.450400701569132\\
};
\addplot [color=black, draw=none, mark size=2.0pt, mark=o, mark options={solid, mycolor1}, forget plot]
  table[row sep=crcr]{%
6.97477347790107	0.0178879046995591\\
7.09697075585111	0.0195274181379774\\
6.93913200297596	0.0229332622979231\\
7.11670040109186	0.0296033233973041\\
7.02978933824483	0.0400344481041678\\
6.91633644163274	0.0444688656547867\\
7.08154975720668	0.0452809260339303\\
};
\addplot [color=black, draw=none, mark size=2.0pt, mark=o, mark options={solid, mycolor1}, forget plot]
  table[row sep=crcr]{%
8.03892327530215	0.0111765329620959\\
8.01161314601468	0.0117099648936569\\
7.93783232085723	0.0117620560206249\\
7.88503909613881	0.0120960992603912\\
7.93334385314343	0.016658970022809\\
7.96527823406246	0.0273011978627668\\
8.03336375875309	0.0302650307403257\\
};
\addplot [color=black, draw=none, mark size=2.0pt, mark=o, mark options={solid, mycolor1}, forget plot]
  table[row sep=crcr]{%
9.12152452405551	0.0002375396386844\\
8.92678916153624	0.000259698006487812\\
9.06427098538894	0.000286719307483399\\
9.09658202740761	0.000331078284697873\\
8.99305746487424	0.000658203425956306\\
8.91472838151818	0.000695946237615363\\
9.07773019641218	0.000984647445005595\\
};
\addplot [color=black, draw=none, mark size=2.0pt, mark=o, mark options={solid, mycolor1}, forget plot]
  table[row sep=crcr]{%
9.99412693817924	0.000265707843438072\\
9.90407188306155	0.000340834688657314\\
10.0939301531962	0.000350621425243716\\
10.0337965948658	0.000778641444986376\\
};
% from other file
\addplot [color=mycolor1, forget plot]
  table[row sep=crcr]{%
1	0.466238245907102\\
2	0.0749520109576602\\
3	0.0150380231357767\\
4	0.000862360763338974\\
5	0.000359773717176647\\
6	0.000386561655774691\\
7	0.000487409414138383\\
8	0.000704633319556755\\
9	4.39064256324928e-05\\
% 10	4.53231797687043e-05\\
};
\addplot [color=mycolor2, dashed, forget plot]
  table[row sep=crcr]{%
1	1.39423272452608\\
2	1.5306023195118\\
3	1.70377681786993\\
4	1.59899515624534\\
5	0.925762769551495\\
6	1.93726787330502\\
7	1.90208317382883\\
8	0.434356551176084\\
9	2.77638396224268e-07\\
% 10	5.65669824845163e-08\\
};
\end{axis}
\end{tikzpicture}%

%% file: frame_rel_emp_error_3_3.tikz
%auto-ignore
\definecolor{mycolor1}{rgb}{0.00000,0.39608,0.74118}%
\definecolor{mycolor2}{rgb}{0.63529,0.67843,0.00000}%
\begin{tikzpicture}

\begin{axis}[%
width=\fwidth,
height=\fheight,
scale only axis,
xmin=0.9,
xmax=8.1,
xtick={1,2,3,4,5,6,7,8},
xticklabels={{ 30 (0.13)},{ 60 (0.25)},{ 90 (0.38)},{120 (0.50)},{150 (0.63)},{180 (0.75)},{210 (0.88)},{240 (1.00)}},
xlabel style={font=\color{white!15!black}},
xlabel={Number of samples in ED (Number of samples divided by number of polynomial terms)},
ymode=log,
ymin=5e-7,%1.68890629536695e-06,
ymax=5e1,%54.3998514723058,
yminorticks=true,
ylabel style={font=\color{white!15!black}},
ylabel={Relative empirical error},
axis background/.style={fill=white},
xmajorgrids,
ymajorgrids%,
%yminorgrids
]
\addplot [color=mycolor1, forget plot]
  table[row sep=crcr]{%
1	1.52624227004451\\
1	7.78041538640132\\
};
\addplot [color=mycolor1, forget plot]
  table[row sep=crcr]{%
2	0.019263099585588\\
2	3.14910249136287\\
};
\addplot [color=mycolor1, forget plot]
  table[row sep=crcr]{%
3	0.00106430866688693\\
3	0.164363955736626\\
};
\addplot [color=mycolor1, forget plot]
  table[row sep=crcr]{%
4	0.000132162119178273\\
4	0.040459114375018\\
};
\addplot [color=mycolor1, forget plot]
  table[row sep=crcr]{%
5	8.74895386034962e-05\\
5	0.0171108445926939\\
};
\addplot [color=mycolor1, forget plot]
  table[row sep=crcr]{%
6	2.51819711628665e-05\\
6	0.00157868665349607\\
};
\addplot [color=mycolor1, forget plot]
  table[row sep=crcr]{%
7	1.58068595190129e-05\\
7	0.00130010689017038\\
};
\addplot [color=mycolor1, forget plot]
  table[row sep=crcr]{%
8	1.43219458048252e-05\\
8	0.000443102097169803\\
};
\addplot [color=mycolor1, forget plot]
  table[row sep=crcr]{%
9	1.1685333393393e-05\\
9	0.000225405989781677\\
};
\addplot [color=mycolor1, forget plot]
  table[row sep=crcr]{%
10	1.25061727882997e-05\\
10	0.000109041017223629\\
};
\addplot [color=mycolor1, line width=4.0pt, forget plot]
  table[row sep=crcr]{%
1	3.18844132578787\\
1	5.06651515916956\\
};
\addplot [color=mycolor1, line width=4.0pt, forget plot]
  table[row sep=crcr]{%
2	0.197792518371741\\
2	1.60407804454336\\
};
\addplot [color=mycolor1, line width=4.0pt, forget plot]
  table[row sep=crcr]{%
3	0.0130730091825402\\
3	0.0745682155523825\\
};
\addplot [color=mycolor1, line width=4.0pt, forget plot]
  table[row sep=crcr]{%
4	0.000882536614410286\\
4	0.0213642253313119\\
};
\addplot [color=mycolor1, line width=4.0pt, forget plot]
  table[row sep=crcr]{%
5	0.000374206610030368\\
5	0.0101933589341834\\
};
\addplot [color=mycolor1, line width=4.0pt, forget plot]
  table[row sep=crcr]{%
6	0.000176634337949532\\
6	0.000833411614887579\\
};
\addplot [color=mycolor1, line width=4.0pt, forget plot]
  table[row sep=crcr]{%
7	7.41845201146112e-05\\
7	0.00085166715689018\\
};
\addplot [color=mycolor1, line width=4.0pt, forget plot]
  table[row sep=crcr]{%
8	3.85860596715362e-05\\
8	0.00024091199249172\\
};
\addplot [color=mycolor1, line width=4.0pt, forget plot]
  table[row sep=crcr]{%
9	2.92712083289784e-05\\
9	0.000112784016303216\\
};
\addplot [color=mycolor1, line width=4.0pt, forget plot]
  table[row sep=crcr]{%
10	2.56494658223189e-05\\
10	7.4953087609416e-05\\
};
\addplot [color=black, draw=none, mark=*, mark options={solid, fill=black, mycolor1}, forget plot]
  table[row sep=crcr]{%
1	3.828565836086\\
};
\addplot [color=black, draw=none, mark=*, mark options={solid, fill=black, mycolor1}, forget plot]
  table[row sep=crcr]{%
2	0.424039788191091\\
};
\addplot [color=black, draw=none, mark=*, mark options={solid, fill=black, mycolor1}, forget plot]
  table[row sep=crcr]{%
3	0.0255202435118665\\
};
\addplot [color=black, draw=none, mark=*, mark options={solid, fill=black, mycolor1}, forget plot]
  table[row sep=crcr]{%
4	0.00298136029093617\\
};
\addplot [color=black, draw=none, mark=*, mark options={solid, fill=black, mycolor1}, forget plot]
  table[row sep=crcr]{%
5	0.000735263830488361\\
};
\addplot [color=black, draw=none, mark=*, mark options={solid, fill=black, mycolor1}, forget plot]
  table[row sep=crcr]{%
6	0.000382746793762798\\
};
\addplot [color=black, draw=none, mark=*, mark options={solid, fill=black, mycolor1}, forget plot]
  table[row sep=crcr]{%
7	0.000359508170668483\\
};
\addplot [color=black, draw=none, mark=*, mark options={solid, fill=black, mycolor1}, forget plot]
  table[row sep=crcr]{%
8	0.000102470610813202\\
};
\addplot [color=black, draw=none, mark=*, mark options={solid, fill=black, mycolor1}, forget plot]
  table[row sep=crcr]{%
9	5.08591493771734e-05\\
};
\addplot [color=black, draw=none, mark=*, mark options={solid, fill=black, mycolor1}, forget plot]
  table[row sep=crcr]{%
10	3.92429943804015e-05\\
};
\addplot [color=black, draw=none, mark=*, mark options={solid, black}, forget plot]
  table[row sep=crcr]{%
1	3.828565836086\\
};
\addplot [color=black, draw=none, mark=*, mark options={solid, black}, forget plot]
  table[row sep=crcr]{%
2	0.424039788191091\\
};
\addplot [color=black, draw=none, mark=*, mark options={solid, black}, forget plot]
  table[row sep=crcr]{%
3	0.0255202435118665\\
};
\addplot [color=black, draw=none, mark=*, mark options={solid, black}, forget plot]
  table[row sep=crcr]{%
4	0.00298136029093617\\
};
\addplot [color=black, draw=none, mark=*, mark options={solid, black}, forget plot]
  table[row sep=crcr]{%
5	0.000735263830488361\\
};
\addplot [color=black, draw=none, mark=*, mark options={solid, black}, forget plot]
  table[row sep=crcr]{%
6	0.000382746793762798\\
};
\addplot [color=black, draw=none, mark=*, mark options={solid, black}, forget plot]
  table[row sep=crcr]{%
7	0.000359508170668483\\
};
\addplot [color=black, draw=none, mark=*, mark options={solid, black}, forget plot]
  table[row sep=crcr]{%
8	0.000102470610813202\\
};
\addplot [color=black, draw=none, mark=*, mark options={solid, black}, forget plot]
  table[row sep=crcr]{%
9	5.08591493771734e-05\\
};
\addplot [color=black, draw=none, mark=*, mark options={solid, black}, forget plot]
  table[row sep=crcr]{%
10	3.92429943804015e-05\\
};
\addplot [color=black, draw=none, mark size=2.0pt, mark=o, mark options={solid, mycolor1}, forget plot]
  table[row sep=crcr]{%
1.07724756679976	0.342292641519749\\
0.964127233370262	8.2866592569767\\
0.893310858622452	8.69201933812167\\
1.02274786381871	8.90359402626327\\
1.10254695768204	10.5167105646037\\
0.923441484041653	11.5351756231352\\
0.983091947883599	34.0830549932951\\
};
\addplot [color=black, draw=none, mark size=2.0pt, mark=o, mark options={solid, mycolor1}, forget plot]
  table[row sep=crcr]{%
2.062289932267	16.1692569130652\\
1.8847961216619	36.5428887395084\\
2.11158124745137	36.8217817224327\\
};
\addplot [color=black, draw=none, mark size=2.0pt, mark=o, mark options={solid, mycolor1}, forget plot]
  table[row sep=crcr]{%
3.06591833091594	0.309793011538088\\
3.01470513762739	0.378424128511245\\
2.92096073611644	0.381523814682503\\
2.99948720375474	0.416177758588169\\
3.00446140005851	0.571624927342991\\
3.12356075266101	1.508287977796\\
};
\addplot [color=black, draw=none, mark size=2.0pt, mark=o, mark options={solid, mycolor1}, forget plot]
  table[row sep=crcr]{%
4.08871292077267	0.0687607086523855\\
4.11560098492799	0.073884119044586\\
4.04473525224426	0.150713962271163\\
3.97587534720109	0.355436902982436\\
4.10874477154451	0.45629566839506\\
};
\addplot [color=black, draw=none, mark size=2.0pt, mark=o, mark options={solid, mycolor1}, forget plot]
  table[row sep=crcr]{%
4.99487113640812	0.0302964845053281\\
4.93294790291578	0.0365640651730238\\
4.97407256212744	0.044741206379739\\
5.0512693687782	0.054411747458044\\
5.01463975864014	0.0558654972168275\\
5.06415767522359	0.141559789326174\\
5.12387026463881	0.163740561129014\\
5.11560785109166	0.265211244046769\\
5.00876677631404	0.615069426082978\\
};
\addplot [color=black, draw=none, mark size=2.0pt, mark=o, mark options={solid, mycolor1}, forget plot]
  table[row sep=crcr]{%
6.11596753249293	0.00283678826298021\\
5.90390646979531	0.00301592616621337\\
5.88786207330755	0.00625255975627291\\
5.95108723640914	0.0169346509937974\\
6.02004795828568	0.0172455423533703\\
6.00774111308457	0.0191763270415056\\
6.10030202316329	0.022452704267472\\
6.01013760629256	0.0336230859336207\\
};
\addplot [color=black, draw=none, mark size=2.0pt, mark=o, mark options={solid, mycolor1}, forget plot]
  table[row sep=crcr]{%
6.98299515271417	0.00376504791617631\\
7.01066674685909	0.0050625698048286\\
7.05310370144738	0.00555042449589808\\
6.87916867823506	0.00760545740569673\\
7.07523022051446	0.00802909278926952\\
6.91062733123095	0.00870947560453011\\
6.99461861822572	0.013153028739532\\
6.93920883854491	0.0177229620505413\\
6.96727292220621	0.0200824747751264\\
7.04044122834135	0.0306090676237068\\
};
\addplot [color=black, draw=none, mark size=2.0pt, mark=o, mark options={solid, mycolor1}, forget plot]
  table[row sep=crcr]{%
7.91740220336358	0.000634895957209932\\
7.94469600510975	0.00279409877873277\\
};
\addplot [color=black, draw=none, mark size=2.0pt, mark=o, mark options={solid, mycolor1}, forget plot]
  table[row sep=crcr]{%
8.92455544862846	0.000282314184504681\\
8.92376788332107	0.000296077222677853\\
8.95670991208749	0.000326327861796626\\
9.0950844650948	0.000351411966620416\\
8.99277546625394	0.000397515212538844\\
8.97599234304274	0.000456749649581173\\
8.91980786910586	0.000639578471780303\\
};
\addplot [color=black, draw=none, mark size=2.0pt, mark=o, mark options={solid, mycolor1}, forget plot]
  table[row sep=crcr]{%
10.1172312490368	0.000174805213309803\\
9.97686393445066	0.000180881767011172\\
10.0861218491209	0.000198038530683433\\
10.0288312742827	0.000211737086229819\\
9.96915276950818	0.000250873725326411\\
10.0942954373343	0.000283600600588281\\
10.0712131068208	0.000312254850232836\\
9.99123857082863	0.000443045315324641\\
};
% from other file
\addplot [color=mycolor1, forget plot]
  table[row sep=crcr]{%
1	3.828565836086\\
2	0.424039788191091\\
3	0.0255202435118665\\
4	0.00298136029093617\\
5	0.000735263830488361\\
6	0.000382746793762798\\
7	0.000359508170668483\\
8	0.000102470610813202\\
% 9	5.08591493771734e-05\\
% 10	3.92429943804015e-05\\
};
\addplot [color=mycolor2, dashed, forget plot]
  table[row sep=crcr]{%
1	1.38825792053767\\
2	1.45917685525225\\
3	1.83398096445436\\
4	1.59943878162314\\
5	2.04122622799695\\
6	2.06201804233032\\
7	0.312865458784004\\
8	4.6292500375264e-05\\
% 9	4.55114552255378e-06\\
% 10	3.0977836019629e-06\\
};
\end{axis}
\end{tikzpicture}%

%% file: comp_emp_error_median_lsq_randn.tikz
%auto-ignore
\definecolor{mycolor1}{rgb}{0.00000,0.44700,0.74100}%
\definecolor{mycolor2}{rgb}{0.85000,0.32500,0.09800}%
\definecolor{mycolor3}{rgb}{0.92900,0.69400,0.12500}%
\definecolor{mycolor4}{rgb}{0.49400,0.18400,0.55600}%
\definecolor{mycolor5}{rgb}{0.46600,0.67400,0.18800}%
\definecolor{mycolor6}{rgb}{0.30100,0.74500,0.93300}%
\begin{tikzpicture}

\begin{axis}[%
width=\fwidth,
height=\fheight,
scale only axis,
%xmode=log,
xmin=30,
xmax=660,
xlabel style={font=\color{white!15!black}},
xlabel={Number of samples in ED},
ymode=log,
ymin=4e-05,
ymax=5,
yminorticks=true,
ylabel style={font=\color{white!15!black}},
ylabel={Median relative empirical error},
axis background/.style={fill=white},
xmajorgrids,
ymajorgrids,
xminorgrids
]
\addplot [color=TUMBlau, forget plot]
  table[row sep=crcr]{%
120	0.190719863915864\\
180	0.0254861159360678\\
240	0.000622394447679252\\
300	0.000663912242903521\\
360	0.000380521329458401\\
420	0.000285366517285173\\
480	0.000173904392170292\\
540	0.000227715189494391\\
600	0.000875365285212379\\
660	5.92073503813418e-05\\
};
\addplot [color=TUMBlau, dash dot, forget plot]
  table[row sep=crcr]{%
120	0.785145305037239\\
180	0.0995889679663972\\
240	0.0036780195845604\\
300	0.00076170619735376\\
360	0.000616006899226788\\
420	0.000379712065928453\\
480	0.000475260531319216\\
540	0.000273752894886166\\
600	0.000359439837295244\\
660	0.000227867339516298\\
};
\addplot [color=TUMGruen, forget plot]
  table[row sep=crcr]{%
30	3.828565836086\\
60	0.424039788191091\\
90	0.0255202435118665\\
120	0.00298136029093617\\
150	0.000735263830488361\\
180	0.000382746793762798\\
210	0.000359508170668483\\
240	0.000102470610813202\\
% 270	5.08591493771734e-05\\
% 300	3.92429943804015e-05\\
};
\addplot [color=TUMGruen, dash dot, forget plot]
  table[row sep=crcr]{%
30	3.52259344920699\\
60	0.559933963034664\\
90	0.0419069704804287\\
120	0.00719642637820356\\
150	0.000648597285344959\\
180	0.000489955992459293\\
210	0.000283390111407165\\
240	0.000217786273198999\\
% 270	0.000146933082919068\\
% 300	0.000111653621140222\\
};
\addplot [color=TUMOrange, forget plot]
  table[row sep=crcr]{%
80	0.466238245907102\\
120	0.0749520109576602\\
160	0.0150380231357767\\
200	0.000862360763338974\\
240	0.000359773717176647\\
280	0.000386561655774691\\
320	0.000487409414138383\\
360	0.000704633319556755\\
400	4.39064256324928e-05\\
% 440	4.53231797687043e-05\\
};
\addplot [color=TUMOrange, dash dot, forget plot]
  table[row sep=crcr]{%
80	0.756964072283003\\
120	0.0662854324766237\\
160	0.0222884714465487\\
200	0.00462913051761551\\
240	0.000786290443085516\\
280	0.000528959694945239\\
320	0.000386373814877944\\
360	0.000276871330965519\\
400	0.000143497932017839\\
% 440	0.000197351383275895\\
};
\end{axis}
\end{tikzpicture}%

%% file: frame_DOS_mean_5_5.tikz
%auto-ignore
\definecolor{mycolor1}{rgb}{0.76863,0.02745,0.10588}%
\definecolor{mycolor2}{rgb}{0.00000,0.48627,0.18824}%
\definecolor{mycolor3}{rgb}{0.00000,0.39608,0.74118}%
\begin{tikzpicture}

\begin{axis}[%
width=\fwidth,
height=\fheight,
scale only axis,
xmin=80,
xmax=400,
xtick={80,120,160,200,240,280,320,360,400,440},
xticklabels={{80},{120},{160},{200},{240},{280},{320},{360},{400},{440}},
xlabel style={font=\color{white!15!black}},
xlabel={Number of samples in ED},
ymin=0,
ymax=1,
ylabel style={font=\color{white!15!black}},
ylabel={Mean degree of Sparsity},
axis background/.style={fill=white}
]

\addplot[area legend, draw=none, fill=mycolor1, fill opacity=0.1, forget plot]
table[row sep=crcr] {%
x	y\\
80	0.248730964467005\\
120	0.192893401015228\\
160	0.152284263959391\\
200	0.131979695431472\\
240	0.131979695431472\\
280	0.121827411167513\\
320	0.16243654822335\\
360	0.147208121827411\\
400	0.131979695431472\\
% 440	0.142131979695431\\
% 440	0.33502538071066\\
400	0.33502538071066\\
360	0.609137055837564\\
320	0.65989847715736\\
280	0.624365482233503\\
240	0.593908629441624\\
200	0.558375634517767\\
160	0.538071065989848\\
120	0.477157360406091\\
80	0.350253807106599\\
}--cycle;

\addplot[area legend, draw=none, fill=mycolor2, fill opacity=0.1, forget plot]
table[row sep=crcr] {%
x	y\\
80	0.182741116751269\\
120	0.152284263959391\\
160	0.0761421319796954\\
200	0.065989847715736\\
240	0.065989847715736\\
280	0.0710659898477157\\
320	0.0710659898477157\\
360	0.0761421319796954\\
400	0.0710659898477157\\
% 440	0.0710659898477157\\
% 440	0.299492385786802\\
400	0.355329949238579\\
360	0.695431472081218\\
320	0.741116751269036\\
280	0.725888324873096\\
240	0.629441624365482\\
200	0.619289340101523\\
160	0.598984771573604\\
120	0.487309644670051\\
80	0.304568527918782\\
}--cycle;

\addplot[area legend, draw=none, fill=mycolor3, fill opacity=0.1, forget plot]
table[row sep=crcr] {%
x	y\\
80	0.233502538071066\\
120	0.185279187817259\\
160	0.126903553299492\\
200	0.101522842639594\\
240	0.106598984771574\\
280	0.101522842639594\\
320	0.129441624365482\\
360	0.109137055837563\\
400	0.104060913705584\\
% 440	0.106598984771574\\
% 440	0.307106598984772\\
400	0.3248730964467\\
360	0.639593908629442\\
320	0.692893401015228\\
280	0.67258883248731\\
240	0.609137055837564\\
200	0.565989847715736\\
160	0.558375634517767\\
120	0.482233502538071\\
80	0.317258883248731\\
}--cycle;
\addplot [color=mycolor3, forget plot]
  table[row sep=crcr]{%
80	0.275279187817259\\
120	0.3651269035533\\
160	0.311218274111675\\
200	0.24756345177665\\
240	0.227005076142132\\
280	0.29741116751269\\
320	0.334670050761421\\
360	0.372893401015228\\
400	0.185431472081218\\
% 440	0.195685279187817\\
};
\addplot [color=mycolor1, dashed, forget plot]
  table[row sep=crcr]{%
80	0.309035532994924\\
120	0.384670050761421\\
160	0.327309644670051\\
200	0.277868020304569\\
240	0.259390862944162\\
280	0.311776649746193\\
320	0.341522842639594\\
360	0.362131979695431\\
400	0.211370558375635\\
% 440	0.217258883248731\\
};
\addplot [color=mycolor2, dashdotted, forget plot]
  table[row sep=crcr]{%
80	0.241522842639594\\
120	0.345583756345178\\
160	0.295126903553299\\
200	0.217258883248731\\
240	0.194619289340102\\
280	0.283045685279188\\
320	0.327817258883249\\
360	0.383654822335025\\
400	0.159492385786802\\
% 440	0.174111675126904\\
};
\end{axis}
\end{tikzpicture}%

%% file: frame_DOS_mean_5_5_rand.tikz
%auto-ignore
\definecolor{mycolor1}{rgb}{0.76863,0.02745,0.10588}%
\definecolor{mycolor2}{rgb}{0.00000,0.48627,0.18824}%
\definecolor{mycolor3}{rgb}{0.00000,0.39608,0.74118}%
\begin{tikzpicture}

\begin{axis}[%
width=\fwidth,
height=\fheight,
scale only axis,
xmin=80,
xmax=400,
xtick={80,120,160,200,240,280,320,360,400,440},
xticklabels={{80},{120},{160},{200},{240},{280},{320},{360},{400},{440}},
xlabel style={font=\color{white!15!black}},
xlabel={Number of samples in ED},
ymin=0,
ymax=1,
ylabel style={font=\color{white!15!black}},
ylabel={Mean degree of Sparsity},
axis background/.style={fill=white}
]

\addplot[area legend, draw=none, fill=mycolor1, fill opacity=0.1, forget plot]
table[row sep=crcr] {%
x	y\\
80	0.263959390862944\\
120	0.192893401015228\\
160	0.17258883248731\\
200	0.152284263959391\\
240	0.137055837563452\\
280	0.137055837563452\\
320	0.131979695431472\\
360	0.121827411167513\\
400	0.142131979695431\\
% 440	0.142131979695431\\
% 440	0.538071065989848\\
400	0.522842639593909\\
360	0.477157360406091\\
320	0.609137055837564\\
280	0.639593908629442\\
240	0.609137055837564\\
200	0.573604060913706\\
160	0.558375634517767\\
120	0.472081218274112\\
80	0.350253807106599\\
}--cycle;

\addplot[area legend, draw=none, fill=mycolor2, fill opacity=0.1, forget plot]
table[row sep=crcr] {%
x	y\\
80	0.157360406091371\\
120	0.111675126903553\\
160	0.0812182741116751\\
200	0.065989847715736\\
240	0.0710659898477157\\
280	0.0710659898477157\\
320	0.0710659898477157\\
360	0.0710659898477157\\
400	0.101522842639594\\
% 440	0.116751269035533\\
% 440	0.583756345177665\\
400	0.619289340101523\\
360	0.573604060913706\\
320	0.65989847715736\\
280	0.751269035532995\\
240	0.6751269035533\\
200	0.700507614213198\\
160	0.604060913705584\\
120	0.487309644670051\\
80	0.299492385786802\\
}--cycle;

\addplot[area legend, draw=none, fill=mycolor3, fill opacity=0.1, forget plot]
table[row sep=crcr] {%
x	y\\
80	0.218274111675127\\
120	0.15989847715736\\
160	0.129441624365482\\
200	0.114213197969543\\
240	0.109137055837563\\
280	0.106598984771574\\
320	0.109137055837563\\
360	0.0964467005076142\\
400	0.131979695431472\\
% 440	0.142131979695431\\
% 440	0.565989847715736\\
400	0.571065989847716\\
360	0.525380710659898\\
320	0.642131979695431\\
280	0.695431472081218\\
240	0.647208121827411\\
200	0.634517766497462\\
160	0.558375634517767\\
120	0.469543147208122\\
80	0.312182741116751\\
}--cycle;
\addplot [color=mycolor3, forget plot]
  table[row sep=crcr]{%
80	0.275482233502538\\
120	0.353401015228426\\
160	0.354771573604061\\
200	0.356345177664975\\
240	0.321979695431472\\
280	0.320964467005076\\
320	0.302842639593909\\
360	0.276802030456853\\
400	0.300710659898477\\
% 440	0.315939086294416\\
};
\addplot [color=mycolor1, dashed, forget plot]
  table[row sep=crcr]{%
80	0.308934010152284\\
120	0.36497461928934\\
160	0.363654822335025\\
200	0.356852791878172\\
240	0.328629441624365\\
280	0.320710659898477\\
320	0.306294416243655\\
360	0.27492385786802\\
400	0.295228426395939\\
% 440	0.312791878172589\\
};
\addplot [color=mycolor2, dashdotted, forget plot]
  table[row sep=crcr]{%
80	0.242030456852792\\
120	0.341827411167513\\
160	0.345888324873097\\
200	0.355837563451777\\
240	0.315329949238579\\
280	0.321218274111675\\
320	0.299390862944162\\
360	0.278680203045685\\
400	0.306192893401015\\
% 440	0.319086294416244\\
};
\end{axis}
\end{tikzpicture}%

%% file: LENO_rel_emp_error.tikz
%auto-ignore
\definecolor{mycolor1}{rgb}{0.00000,0.39608,0.74118}%
\definecolor{mycolor2}{rgb}{0.63529,0.67843,0.00000}%
\begin{tikzpicture}

\begin{axis}[%
width=\fwidth,
height=0.999\fheight,
at={(0\fwidth,0\fheight)},
scale only axis,
xmin=0.881108504012593,
xmax=9.10527431397305,
xtick={1,2,3,4,5,6,7,8,9},
xticklabels={{ 320 (0.19)},{ 480 (0.29)},{ 640 (0.39)},{ 800 (0.48)},{ 960 (0.58)},{1120 (0.68)},{1280 (0.77)},{1440 (0.87)},{1600 (0.97)}},
xlabel style={font=\color{white!15!black}},
xlabel={Number of samples in ED (Number of samples divided by number of polynomial terms)},
ymode=log,
ymin=1e-3,
ymax=1e3,
yminorticks=true,
ylabel style={font=\color{white!15!black}},
ylabel={Relative empirical error},
axis background/.style={fill=white},
xmajorgrids,
ymajorgrids,
%yminorgrids
]
\addplot [color=mycolor1, forget plot]
  table[row sep=crcr]{%
1	0.437985462310451\\
1	5.25231756045371\\
};
\addplot [color=mycolor1, forget plot]
  table[row sep=crcr]{%
2	0.106492095088754\\
2	2.40323878067529\\
};
\addplot [color=mycolor1, forget plot]
  table[row sep=crcr]{%
3	0.069199197985216\\
3	0.762668725136215\\
};
\addplot [color=mycolor1, forget plot]
  table[row sep=crcr]{%
4	0.044976319248765\\
4	0.264548585903137\\
};
\addplot [color=mycolor1, forget plot]
  table[row sep=crcr]{%
5	0.0200174501403224\\
5	0.366232905775788\\
};
\addplot [color=mycolor1, forget plot]
  table[row sep=crcr]{%
6	0.00533441075916372\\
6	0.232672616664212\\
};
\addplot [color=mycolor1, forget plot]
  table[row sep=crcr]{%
7	0.0125390846849116\\
7	0.1407740858505\\
};
\addplot [color=mycolor1, forget plot]
  table[row sep=crcr]{%
8	0.00972330081973368\\
8	0.0846162077796675\\
};
\addplot [color=mycolor1, forget plot]
  table[row sep=crcr]{%
9	0.00374190565264806\\
9	0.0674832702845504\\
};
\addplot [color=mycolor1, line width=4.0pt, forget plot]
  table[row sep=crcr]{%
1	0.790618311515351\\
1	3.58000879860534\\
};
\addplot [color=mycolor1, line width=4.0pt, forget plot]
  table[row sep=crcr]{%
2	0.313537128690853\\
2	1.27180441254947\\
};
\addplot [color=mycolor1, line width=4.0pt, forget plot]
  table[row sep=crcr]{%
3	0.149042966576499\\
3	0.446600886151053\\
};
\addplot [color=mycolor1, line width=4.0pt, forget plot]
  table[row sep=crcr]{%
4	0.0769991798203361\\
4	0.154843262381836\\
};
\addplot [color=mycolor1, line width=4.0pt, forget plot]
  table[row sep=crcr]{%
5	0.0574123992138163\\
5	0.186819513547352\\
};
\addplot [color=mycolor1, line width=4.0pt, forget plot]
  table[row sep=crcr]{%
6	0.0369635428897863\\
6	0.116619348210516\\
};
\addplot [color=mycolor1, line width=4.0pt, forget plot]
  table[row sep=crcr]{%
7	0.0299721464285365\\
7	0.0765783895819851\\
};
\addplot [color=mycolor1, line width=4.0pt, forget plot]
  table[row sep=crcr]{%
8	0.0209322161667542\\
8	0.0524491443177013\\
};
\addplot [color=mycolor1, line width=4.0pt, forget plot]
  table[row sep=crcr]{%
9	0.0129062275629277\\
9	0.0369482894495178\\
};
\addplot [color=black, draw=none, mark=*, mark options={solid, fill=black, mycolor1}, forget plot]
  table[row sep=crcr]{%
1	1.75800489194225\\
};
\addplot [color=black, draw=none, mark=*, mark options={solid, fill=black, mycolor1}, forget plot]
  table[row sep=crcr]{%
2	0.465801632008407\\
};
\addplot [color=black, draw=none, mark=*, mark options={solid, fill=black, mycolor1}, forget plot]
  table[row sep=crcr]{%
3	0.214642482770347\\
};
\addplot [color=black, draw=none, mark=*, mark options={solid, fill=black, mycolor1}, forget plot]
  table[row sep=crcr]{%
4	0.115133762067399\\
};
\addplot [color=black, draw=none, mark=*, mark options={solid, fill=black, mycolor1}, forget plot]
  table[row sep=crcr]{%
5	0.114520583579578\\
};
\addplot [color=black, draw=none, mark=*, mark options={solid, fill=black, mycolor1}, forget plot]
  table[row sep=crcr]{%
6	0.0603370851745556\\
};
\addplot [color=black, draw=none, mark=*, mark options={solid, fill=black, mycolor1}, forget plot]
  table[row sep=crcr]{%
7	0.056011677510366\\
};
\addplot [color=black, draw=none, mark=*, mark options={solid, fill=black, mycolor1}, forget plot]
  table[row sep=crcr]{%
8	0.0321401105861375\\
};
\addplot [color=black, draw=none, mark=*, mark options={solid, fill=black, mycolor1}, forget plot]
  table[row sep=crcr]{%
9	0.0241315862678075\\
};
\addplot [color=black, draw=none, mark=*, mark options={solid, black}, forget plot]
  table[row sep=crcr]{%
1	1.75800489194225\\
};
\addplot [color=black, draw=none, mark=*, mark options={solid, black}, forget plot]
  table[row sep=crcr]{%
2	0.465801632008407\\
};
\addplot [color=black, draw=none, mark=*, mark options={solid, black}, forget plot]
  table[row sep=crcr]{%
3	0.214642482770347\\
};
\addplot [color=black, draw=none, mark=*, mark options={solid, black}, forget plot]
  table[row sep=crcr]{%
4	0.115133762067399\\
};
\addplot [color=black, draw=none, mark=*, mark options={solid, black}, forget plot]
  table[row sep=crcr]{%
5	0.114520583579578\\
};
\addplot [color=black, draw=none, mark=*, mark options={solid, black}, forget plot]
  table[row sep=crcr]{%
6	0.0603370851745556\\
};
\addplot [color=black, draw=none, mark=*, mark options={solid, black}, forget plot]
  table[row sep=crcr]{%
7	0.056011677510366\\
};
\addplot [color=black, draw=none, mark=*, mark options={solid, black}, forget plot]
  table[row sep=crcr]{%
8	0.0321401105861375\\
};
\addplot [color=black, draw=none, mark=*, mark options={solid, black}, forget plot]
  table[row sep=crcr]{%
9	0.0241315862678075\\
};
\addplot [color=black, draw=none, mark size=2.0pt, mark=o, mark options={solid, mycolor1}, forget plot]
  table[row sep=crcr]{%
0.906218189679953	7.90808343240264\\
0.881108504012593	8.46700980309428\\
0.947546316282682	9.01081126054526\\
0.954380145724807	9.07022375481809\\
1.03842253349162	11.1354715666591\\
};
\addplot [color=black, draw=none, mark size=2.0pt, mark=o, mark options={solid, mycolor1}, forget plot]
  table[row sep=crcr]{%
2.11423398101767	2.84363016525182\\
2.10893271819622	3.22293804560976\\
1.98947158346359	3.44999326425553\\
1.93511959920802	3.88740370136068\\
2.06597448607162	4.37528219561754\\
2.06483184578277	6.83056205247673\\
2.06016201624465	14.8650441111337\\
};
\addplot [color=black, draw=none, mark size=2.0pt, mark=o, mark options={solid, mycolor1}, forget plot]
  table[row sep=crcr]{%
3.06092208537183	0.953736425483839\\
2.90148010418319	1.30704407587158\\
3.04539010761758	1.39194406351407\\
2.99081514464843	1.46819385573409\\
2.92804080131373	2.32821807303974\\
};
\addplot [color=black, draw=none, mark size=2.0pt, mark=o, mark options={solid, mycolor1}, forget plot]
  table[row sep=crcr]{%
3.89962968442203	0.276280236774262\\
4.08089361848196	0.311414639907713\\
3.91875243434552	0.629166899007664\\
3.91589247744625	0.737919259101824\\
4.04149680410278	8.98695119384274\\
};
\addplot [color=black, draw=none, mark size=2.0pt, mark=o, mark options={solid, mycolor1}, forget plot]
  table[row sep=crcr]{%
5.09859734383856	0.621030503944471\\
5.00413955208782	0.763044947199845\\
5.05067557673762	1.4141588200971\\
4.91339759415485	1.90459941339035\\
5.11336426747156	2.70560427786321\\
5.01022102031037	9.25335404779979\\
};
\addplot [color=black, draw=none, mark size=2.0pt, mark=o, mark options={solid, mycolor1}, forget plot]
  table[row sep=crcr]{%
6.04493347455262	0.243786900540528\\
5.88414075451211	0.30513933185369\\
6.07730096282345	0.840106689370348\\
6.06215471794405	1.15778579348562\\
5.90504675449677	1.36504370969585\\
6.00626129119065	1.51068110876281\\
};
\addplot [color=black, draw=none, mark size=2.0pt, mark=o, mark options={solid, mycolor1}, forget plot]
  table[row sep=crcr]{%
6.95645840719081	0.155697507698958\\
7.01161235997577	0.189370883763988\\
6.9747201880958	0.192056773987148\\
};
\addplot [color=black, draw=none, mark size=2.0pt, mark=o, mark options={solid, mycolor1}, forget plot]
  table[row sep=crcr]{%
7.97877334665326	0.100846531392\\
7.9201844400637	0.104717386968616\\
7.93884668512201	0.160584071504629\\
7.88013394366455	0.160901512102266\\
8.1059189031551	0.191160238997112\\
8.03842497225206	0.266165418154808\\
8.10815339301214	0.371007786751592\\
};
\addplot [color=black, draw=none, mark size=2.0pt, mark=o, mark options={solid, mycolor1}, forget plot]
  table[row sep=crcr]{%
8.91587809213188	0.0750455409117163\\
9.10527431397305	0.0954149237698597\\
9.07366447134719	0.102988814309924\\
9.01934854917666	0.124046617080461\\
};
% from other file
\addplot [color=mycolor1, forget plot]
table[row sep=crcr]{%
	1	1.75800489194225\\
	2	0.465801632008407\\
	3	0.214642482770347\\
	4	0.115133762067399\\
	5	0.114520583579578\\
	6	0.0603370851745556\\
	7	0.056011677510366\\
	8	0.0321401105861375\\
	9	0.0241315862678075\\
};
\addplot [color=mycolor2, dashed, forget plot]
table[row sep=crcr]{%
	1	75.1087972859534\\
	2	360.722143621583\\
	3	95.4838155560067\\
	4	69.367775005667\\
	5	30.3728590863996\\
	6	12.4344847894915\\
	7	5.02874738171946\\
	8	3.1101003714543\\
	9	2.24079897629384\\
};
\end{axis}
\end{tikzpicture}%

%% file: LENO81EKL_comp_emp_error_median_lsq_randn.tikz
%auto-ignore
\definecolor{mycolor1}{rgb}{0.00000,0.44700,0.74100}%
\definecolor{mycolor2}{rgb}{0.85000,0.32500,0.09800}%
\definecolor{mycolor3}{rgb}{0.92900,0.69400,0.12500}%
\definecolor{mycolor4}{rgb}{0.49400,0.18400,0.55600}%
\definecolor{mycolor5}{rgb}{0.46600,0.67400,0.18800}%
\definecolor{mycolor6}{rgb}{0.30100,0.74500,0.93300}%
\begin{tikzpicture}

\begin{axis}[%
width=0.951\fwidth,
height=\fheight,
at={(0\fwidth,0\fheight)},
scale only axis,
%xmode=log,
xmin=240,
xmax=1600,
xminorticks=true,
xlabel style={font=\color{white!15!black}},
xlabel={Number of samples in ED},
ymode=log,
ymin=0.0241315862678075,
ymax=3.67443666113593,
yminorticks=true,
ylabel style={font=\color{white!15!black}},
ylabel={Median relative empirical error},
axis background/.style={fill=white},
xmajorgrids,
% xminorgrids,
ymajorgrids,
% yminorgrids
]
\addplot [color=TUMBlau, forget plot]
  table[row sep=crcr]{%
320	1.75800489194225\\
480	0.465801632008407\\
640	0.214642482770347\\
800	0.115133762067399\\
960	0.114520583579578\\
1120	0.0603370851745556\\
1280	0.056011677510366\\
1440	0.0321401105861375\\
1600	0.0241315862678075\\
};
\addplot [color=TUMBlau, dash dot, forget plot]
  table[row sep=crcr]{%
320	2.91776221673691\\
480	0.570262277748283\\
640	0.234959966461722\\
800	0.111168511244131\\
960	0.105840106123564\\
1120	0.0581430550259761\\
1280	0.0473178066738021\\
1440	0.0330635887666296\\
1600	0.0329945192901853\\
};
\addplot [color=TUMGruen, forget plot]
  table[row sep=crcr]{%
140	1.6560784686188\\
210	0.810820722639782\\
280	0.550835118537643\\
350	0.296610332283697\\
420	0.233564304254944\\
490	0.159150387986676\\
560	0.131097070229839\\
630	0.112303960421851\\
700	0.0905119744782028\\
770	0.0784784220617408\\
};
\addplot [color=TUMGruen, dash dot, forget plot]
  table[row sep=crcr]{%
140	3.67443666113593\\
210	1.38680868220292\\
280	0.441798899480908\\
350	0.243044805700687\\
420	0.238037351365303\\
490	0.154239774940606\\
560	0.104696888569409\\
630	0.115046882254376\\
700	0.0972529933675313\\
770	0.0636592883996149\\
};
\addplot [color=TUMOrange, forget plot]
  table[row sep=crcr]{%
240	1.27728424941456\\
360	0.409293378029159\\
480	0.221566593310997\\
600	0.148309137796198\\
720	0.101739796513522\\
840	0.0826990689862767\\
960	0.0671923038623922\\
1080	0.0489624420673373\\
1200	0.0339614228121863\\
};
\addplot [color=TUMOrange, dash dot, forget plot]
  table[row sep=crcr]{%
240	1.7980144807933\\
360	0.649139295589132\\
480	0.239886850007369\\
600	0.166525254600517\\
720	0.102673431151686\\
840	0.0777595567398942\\
960	0.0545117801538415\\
1080	0.0427964848192935\\
1200	0.0320531486950332\\
};
\end{axis}
\end{tikzpicture}%

%% file: LENO_DOS_mean.tikz
%auto-ignore
\definecolor{mycolor1}{rgb}{0.76863,0.02745,0.10588}%
\definecolor{mycolor2}{rgb}{0.00000,0.48627,0.18824}%
\definecolor{mycolor3}{rgb}{0.00000,0.39608,0.74118}%
\begin{tikzpicture}

\begin{axis}[%
width=\fwidth,
height=\fheight,
at={(0\fwidth,0\fheight)},
scale only axis,
xmin=320,
xmax=1600,
xtick={320,480,640,800,960,1120,1280,1440,1600},
xticklabels={{320},{480},{640},{800},{960},{1120},{1280},{1440},{1600}},
xlabel style={font=\color{white!15!black}},
xlabel={Number of samples in ED},
ymin=0,
ymax=0.5,
ylabel style={font=\color{white!15!black}},
ylabel={Mean degree of Sparsity},
axis background/.style={fill=white}
]

\addplot[area legend, draw=none, fill=mycolor1, fill opacity=0.1, forget plot]
table[row sep=crcr] {%
x	y\\
320	0.151331719128329\\
480	0.0968523002421307\\
640	0.112590799031477\\
800	0.0702179176755448\\
960	0.049636803874092\\
1120	0.0314769975786925\\
1280	0.0326876513317191\\
1440	0.0314769975786925\\
1600	0.0254237288135593\\
1600	0.0653753026634383\\
1440	0.0786924939467312\\
1280	0.104116222760291\\
1120	0.13680387409201\\
960	0.188861985472155\\
800	0.249394673123487\\
640	0.268765133171913\\
480	0.348668280871671\\
320	0.331719128329298\\
}--cycle;

\addplot[area legend, draw=none, fill=mycolor2, fill opacity=0.1, forget plot]
table[row sep=crcr] {%
x	y\\
320	0.0217917675544794\\
480	0.0242130750605327\\
640	0.0193704600484262\\
800	0.0242130750605327\\
960	0.0314769975786925\\
1120	0.0387409200968523\\
1280	0.0472154963680387\\
1440	0.0617433414043584\\
1600	0.0677966101694915\\
1600	0.101694915254237\\
1440	0.0968523002421307\\
1280	0.099273607748184\\
1120	0.0895883777239709\\
960	0.0774818401937046\\
800	0.0677966101694915\\
640	0.0569007263922518\\
480	0.0944309927360775\\
320	0.0883777239709443\\
}--cycle;

\addplot[area legend, draw=none, fill=mycolor3, fill opacity=0.1, forget plot]
table[row sep=crcr] {%
x	y\\
320	0.104721549636804\\
480	0.0841404358353511\\
640	0.0780871670702179\\
800	0.0587167070217918\\
960	0.051452784503632\\
1120	0.0466101694915254\\
1280	0.051452784503632\\
1440	0.049636803874092\\
1600	0.0502421307506053\\
1600	0.0883777239709443\\
1440	0.0841404358353511\\
1280	0.0938256658595642\\
1120	0.0980629539951574\\
960	0.115617433414044\\
800	0.138014527845036\\
640	0.158595641646489\\
480	0.222154963680387\\
320	0.207627118644068\\
}--cycle;
\addplot [color=mycolor3, forget plot]
  table[row sep=crcr]{%
320	0.169430992736077\\
480	0.146731234866828\\
640	0.120605326876513\\
800	0.0899394673123487\\
960	0.0773486682808717\\
1120	0.0693099273607748\\
1280	0.0684503631961259\\
1440	0.0630266343825666\\
1600	0.0640556900726392\\
};
\addplot [color=mycolor1, dashed, forget plot]
  table[row sep=crcr]{%
320	0.288837772397095\\
480	0.248861985472155\\
640	0.203898305084746\\
800	0.140871670702179\\
960	0.10317191283293\\
1120	0.0787651331719128\\
1280	0.0658353510895884\\
1440	0.0496610169491525\\
1600	0.0439951573849879\\
};
\addplot [color=mycolor2, dashdotted, forget plot]
  table[row sep=crcr]{%
320	0.0500242130750605\\
480	0.0446004842615012\\
640	0.0373123486682809\\
800	0.0390072639225182\\
960	0.0515254237288136\\
1120	0.0598547215496368\\
1280	0.0710653753026634\\
1440	0.0763922518159806\\
1600	0.0841162227602906\\
};
\end{axis}
\end{tikzpicture}%

%% file: LENO81EKL_inv_variate_p.tikz
%auto-ignore
\definecolor{mycolor1}{rgb}{0.00000,0.39608,0.74118}%
\definecolor{mycolor2}{rgb}{0.85490,0.84314,0.79608}%
\begin{tikzpicture}

\begin{axis}[%
width=\fwidth,
height=\fheight,
scale only axis,
bar shift auto,
xmin=0.514285714285714,
xmax=11.4857142857143,
xtick={1,2,3,4,5,6,7,8,9,10,11},
xticklabels={{$E_x$},{$E_y$},{$E_z$},{$G_{xy}$},{$G_{xz}$},{$G_{yz}$},{$\nu_{yx}$},{$\nu_{zx}$},{$\nu_{zy}$},{$\rho$},{$\eta$}},
xlabel style={font=\color{white!15!black}},
xlabel={Variate},
ymin=0,
ymax=160,
ylabel style={font=\color{white!15!black}},
ylabel={Number of polynomials},
axis background/.style={fill=white}
]
\addplot[ybar, bar width=0.229, fill=mycolor1!30, fill opacity=0.3, draw=mycolor2, area legend] table[row sep=crcr] {%
1	155\\
2	155\\
3	155\\
4	155\\
5	155\\
6	155\\
7	155\\
8	155\\
9	155\\
10	155\\
11	155\\
};
\addplot[forget plot, color=white!15!black] table[row sep=crcr] {%
0.514285714285714	0\\
11.4857142857143	0\\
};
\addplot[ybar, bar width=0.229, fill=mycolor1, draw=mycolor2, area legend] table[row sep=crcr] {%
1	7\\
2	5\\
3	0\\
4	6\\
5	0\\
6	0\\
7	0\\
8	0\\
9	0\\
10	11\\
11	7\\
};
\addplot[forget plot, color=white!15!black] table[row sep=crcr] {%
0.514285714285714	0\\
11.4857142857143	0\\
};
\end{axis}
\end{tikzpicture}%

%% file: LENO81EKL_inv_variate_q.tikz
%auto-ignore
\definecolor{mycolor1}{rgb}{0.89020,0.44706,0.13333}%
\definecolor{mycolor2}{rgb}{0.85490,0.84314,0.79608}%
\begin{tikzpicture}

\begin{axis}[%
width=\fwidth,
height=\fheight,
scale only axis,
bar shift auto,
xmin=0.514285714285714,
xmax=11.4857142857143,
xtick={1,2,3,4,5,6,7,8,9,10,11},
xticklabels={{$E_x$},{$E_y$},{$E_z$},{$G_{xy}$},{$G_{xz}$},{$G_{yz}$},{$\nu_{yx}$},{$\nu_{zx}$},{$\nu_{zy}$},{$\rho$},{$\eta$}},
xlabel style={font=\color{white!15!black}},
xlabel={Variate},
ymin=0,
ymax=160,
ylabel style={font=\color{white!15!black}},
ylabel={Number of polynomials},
axis background/.style={fill=white}
]
\addplot[ybar, bar width=0.229, fill=mycolor1!30, fill opacity=0.3, draw=mycolor2, area legend] table[row sep=crcr] {%
1	155\\
2	155\\
3	155\\
4	155\\
5	155\\
6	155\\
7	155\\
8	155\\
9	155\\
10	155\\
11	155\\
};
\addplot[forget plot, color=white!15!black] table[row sep=crcr] {%
0.514285714285714	0\\
11.4857142857143	0\\
};
\addplot[ybar, bar width=0.229, fill=mycolor1, draw=mycolor2, area legend] table[row sep=crcr] {%
1	19\\
2	19\\
3	0\\
4	29\\
5	7\\
6	8\\
7	0\\
8	0\\
9	0\\
10	37\\
11	17\\
};
\addplot[forget plot, color=white!15!black] table[row sep=crcr] {%
0.514285714285714	0\\
11.4857142857143	0\\
};
\end{axis}
\end{tikzpicture}%